%% file: main.tex
\documentclass[12pt,a4paper]{article}
\usepackage[a4paper, margin=1in]{geometry}
\usepackage[utf8]{inputenc}

\usepackage{amsmath}
\usepackage{amssymb}
\usepackage{bm}
\usepackage{bbm}
\usepackage{booktabs}
\usepackage{hyperref}
\usepackage[normalem]{ulem}
\usepackage{listings}
\usepackage{xcolor}
\usepackage{tabularx}
\usepackage{array}
\usepackage{float}

\usepackage{setspace}
\usepackage{fancyhdr}
\usepackage[acronym,nomain,nonumberlist]{glossaries}
\makeglossaries

\input{acronyms}

\usepackage[backend=biber, style=apa, sortcites=false,sorting=nyt,uniquelist=false,natbib=true,doi=true,isbn=false,eprint=false,pagetracker,ibidtracker=constrict, giveninits=true]{biblatex}
\usepackage{multirow}
\usepackage{makecell}

\usepackage{subcaption}
\usepackage{wrapfig}
\usepackage{float}

\newenvironment{DIFnomarkup}{}{}

\usepackage{tikz}
\usepackage{tikzscale}
\usepackage{pgfplots}
\pgfplotsset{compat=1.18}

\usepackage[edges]{forest}
\usetikzlibrary{shapes,decorations,decorations.text,decorations.pathmorphing,automata,arrows,calc,arrows.meta,fit,positioning,shadows,fit,patterns.meta,patterns,tikzmark}
\tikzset{
    point/.style = {circle, draw, inner sep=0.04cm,fill,node contents={}},
    bidirected/.style={Latex-Latex,dashed},
    el/.style = {inner sep=2pt, align=left, sloped},
}

\tikzset{
    state/.style ={rectangle, rounded corners, draw, minimum width = 0.7 cm, align=center, inner sep=0.5em},
    bidirected/.style={Latex-Latex,dashed},
    el/.style = {inner sep=2pt, align=left, sloped},
    latent/.style = {rectangle, draw=black, thick, fill=white},
    observable/.style ={rectangle, draw=black, thick, fill=blue!20},
}

\usepackage{xcolor}
\usepackage{colortbl}
\tikzset{
    > = stealth,
    every node/.append style = {
        text = black
    },
    every path/.append style = {
        arrows = ->,
        draw = black,
    },
   invisible/.style={opacity=0},
   visible on/.style={alt=#1{}{invisible}},
   alt/.code args={<#1>#2#3}{%
    \alt<#1>{\pgfkeysalso{#2}}{\pgfkeysalso{#3}} 
  },
}

\usepackage{pgfplots}

\usepackage{stmaryrd}

\title{\vspace{-3em}Classification errors distort findings in automated speech processing: examples and solutions from child-development research}

\date{}

\author{Lucas Gautheron\textsuperscript{1,2,3*}\and Evan Kidd\textsuperscript{4} \and Anton Malko\textsuperscript{4} \and Marvin Lavechin\textsuperscript{5} \and Alejandrina Cristia\textsuperscript{3}}

\begin{document}

\maketitle

\begin{abstract}
  With the advent of wearable recorders, scientists are increasingly turning to automated methods of analysis of audio and video data in order to measure children's experience, behavior, and outcomes, with a sizable literature employing long-form audio-recordings to study language acquisition. While numerous articles report on the accuracy and reliability of the most popular automated classifiers, less has been written on the downstream effects of classification errors on measurements and statistical inferences (e.g., the estimate of correlations and effect sizes in regressions).  This paper's main contributions are drawing attention to downstream effects of confusion errors, and providing an approach to measure and potentially recover from these errors. Specifically, we use a Bayesian approach to study the effects of algorithmic errors on key scientific questions, including the effect of siblings on children's language experience and the association between children's production and their input. By fitting a joint model of speech behavior and algorithm behavior on real and simulated data, we show that classification errors can significantly distort estimates for both the most commonly used \gls{lena}, and a slightly more accurate open-source alternative (the Voice Type Classifier from the ACLEW system). We further show that a Bayesian calibration approach for recovering unbiased estimates of effect sizes can be effective and insightful, but does not provide a fool-proof solution.  
\end{abstract}



\vfill
{\small
~\\
\footnotemark[1]Evolution, Science and Society, University of Missouri, Columbia, United States\\
\footnotemark[2]University of Wuppertal, Germany\\
\footnotemark[3]Laboratoire de Sciences Cognitives et Psycholinguistique, Département d’études cognitives, ENS, EHESS, CNRS, PSL University, Paris, France \\
\footnotemark[4]School of Literature, Languages and Linguistics, Australian National University, Canberra, Australia\\
\footnotemark[5]Laboratoire d'Informatique et Systèmes, Université Aix-Marseille, CNRS, France}

\newpage

\newpage


\tableofcontents

\glsresetall

\newpage 
\section{Introduction}

Children's behavior and their environments are increasingly described through automated analysis of data collected from wearables. Pioneers in such techniques, researchers working on language acquisition have shown the promise of using automated classifiers to analyze long-form audio recordings, thus enabling the processing of naturalistic data at unprecedented scale \citep{bergelson2023everyday}. A common application in this context is the automatic segmentation and classification of speech into voice categories for measuring speech afforded to and produced by infants. A growing number of studies document and discuss the accuracy of these automated classifiers (such as \gls{lena}), by comparing human and machine annotations of the same audio clips \citep{xu2009reliability,lavechin2025performance}. Concern has been raised about unexpected cases of low recall and precision (e.g., a precision of 27\% for the recognition of the child wearing the device in \citealt{gilkerson2015evaluating}), as well as trends for confusion across speaker types \citep{lehet2021circumspection}. To our knowledge, less attention has been paid to how classification errors propagate through subsequent analyses. This paper examines a critical question: To what extent do errors in automated speech processing systems like \gls{lena} affect downstream measurements and scientific conclusions? Specifically, we investigate how speaker tagging misclassifications (e.g. confusing child speech with adult speech) impact measurements of children's linguistic input and production, as well as statistical estimates (e.g. effect sizes) derived from these measurements.  To study this, we introduce a Bayesian approach that simultaneously models speech behavior as well as algorithm behavior. This joint model allows us to shed light on the profound, and downstream, consequences of algorithmic errors. Additionally, we take this investigation a step further by using the joint model to recover unbiased measurements that take into account algorithmic errors. The methodological insights gained here may also apply to signals captured by other wearable technologies (e.g., video; \citealt{long2024babyview}), and generally, to any case where event detection is performed in conjunction with classification using machine learning algorithms with non-zero error rates. However, both for clarity and to better inform one specific community, we focus on the case of voice type classifiers.

\subsection{\label{section:the_case_of}The case of voice type classifiers when describing early language acquisition}

Day-long audio recordings of children's language experience collected with wearable devices have become widespread thanks to their richness and ecological validity. The adoption of this methodology has been facilitated by \gls{lena}, a user-friendly commercial solution simplifying data collection and analysis. \gls{lena} provides both the recording device and the software for automatically annotating the audio. The latter is important, given that this technique produces a large amount of data (up to 16-24 hours of audio per session in the case of \gls{lena}, to be multiplied by the number of children and the number of sessions per child), which would be impossible to analyze entirely by hand. Among other things, \gls{lena} includes a diarization algorithm that detects speech (\textit{event detection}) and attributes it to one of four different types of speaker (\textit{classification}): the child wearing the recording (CHI), another child -- e.g. a sibling -- (OCH), female adult (FEM) and male adult (MAL).\footnote{\gls{lena} also returns other classes, such as TV/electronic noise. Since these have been more seldom the target of methodological work and less commonly used in scientific research by and large, we do not consider them here.} 
With additional processing steps that build on this essential diarization algorithm, \gls{lena} provides a number of metrics, including the child vocalization count (CVC; i.e., the number of speech-like segments attributed to the key child).  Researchers were quick to consider potential errors in the algorithm. Processing audio data collected directly from children in noisy real-life conditions is challenging, and the metrics returned by \gls{lena} (or any other classifier) are far from perfect. Yet, \gls{lena} is generally considered to have been sufficiently validated, with a meta-analysis \citep{cristia2020accuracy} finding that the correlation between human and \gls{lena} CVC in the same audio clips averaged Pearson $R^2=0.77$ $(N=5)$. Moreover, another meta-analysis found that CVC correlated with concurrent and/or longitudinal standardized measures of language (Pearson $R^2=0.33$, $N=10$; \citealt{wang2020meta}). 

\begin{figure}[h]
    \centering 
    \includegraphics[width=0.95\textwidth]{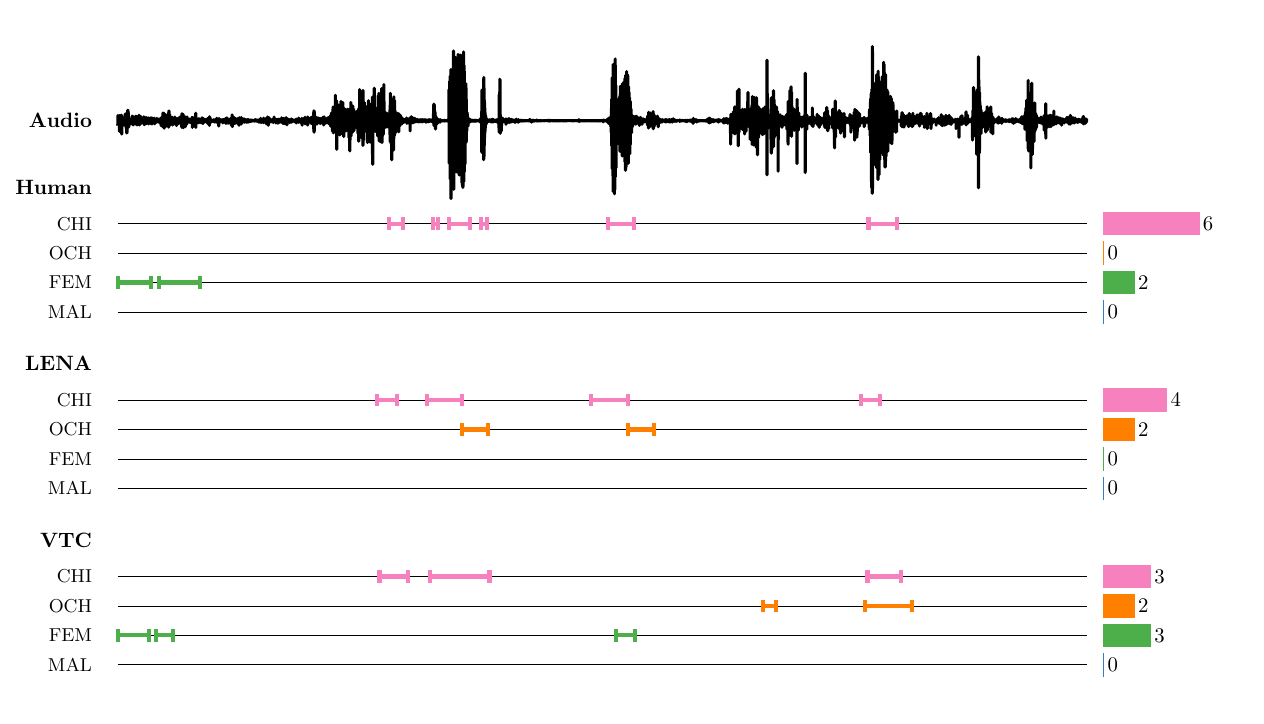}
    \caption{30-second sample of a daylong recording annotated by a human expert and two algorithms: \gls{lena}, and \gls{vtc}. CHI refers to the child wearing the recording device; OCH refers to other children; FEM and MAL refer to female and male adults. A segment of speech is referred to as a  ``vocalization'' (for instance, the expert found two female adult vocalizations in this portion of audio, but \gls{lena} found none). Vocalization counts are shown to the right.}
    \label{fig:sample}
\end{figure}

While validation studies are undeniably important, we believe that computing performance metrics is not sufficient and that there has not been enough consideration about how algorithms' errors may impact conclusions. We illustrate this on the widely used vocalization count measure. 
Figure \ref{fig:sample} shows the segmentation into speaker categories of a 30s audio clip made by a human expert, the proprietary \gls{lena} algorithm, and its open-source alternative, \gls{vtc} \citep{gilkerson2015evaluating,lavechin2020open}. Errors in the automated annotations result in erroneous vocalization counts. For instance, in this example, both \gls{lena} and \gls{vtc} incorrectly report two vocalizations from siblings.

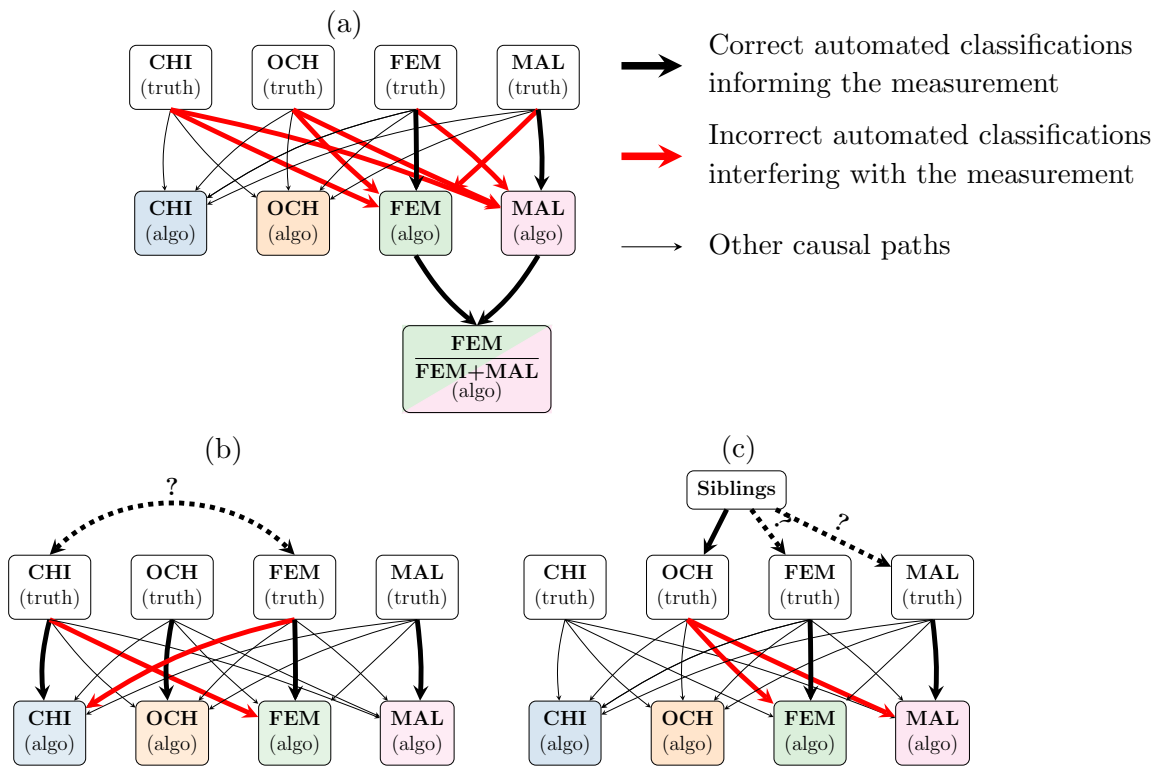
\begin{figure}[!h]
\centering
\begin{subfigure}[t]{0.4\textwidth}
    \caption{\label{fig:example_fem_prop}}
    \resizebox{\textwidth}{!}{\input{figures/Fig2d}}
\end{subfigure}\hspace{1em}\begin{subfigure}[t]{0.2\textwidth}
\begin{tikzpicture}[baseline={(0,1)}]
        \coordinate (legend-start) at (0, 0);
        
        \draw[->, line width=1mm, color=black] ($(legend-start) + (0, 0)$) -- ($(legend-start) + (0.8, 0)$);
        \node[anchor=west,align=left] at ($(legend-start) + (1, 0)$) {\small Correct automated classifications\\\small informing the measurement};
        
        \draw[->, line width=1mm, color=red] ($(legend-start) + (0, -1.2)$) -- ($(legend-start) + (0.8, -1.2)$);
        \node[anchor=west,align=left] at ($(legend-start) + (1, -1.2)$) {\small Incorrect automated classifications\\\small interfering with the measurement};
        
        \draw[->] ($(legend-start) + (0, -2.4)$) -- ($(legend-start) + (0.8, -2.4)$);
        \node[anchor=west,align=left] at ($(legend-start) + (1, -2.4)$) {\small Other causal effects of speech\\on the classifier's output};

        \draw[->,line width=1mm,color=gray] ($(legend-start) + (0, -3.6)$) -- ($(legend-start) + (0.8, -3.6)$);
        \node[anchor=west,align=left] at ($(legend-start) + (1, -3.6)$) {\small Known behavioral effects};

        \draw[->,line width=1mm,dotted,color=gray] ($(legend-start) + (0, -4.8)$) -- ($(legend-start) + (0.8, -4.8)$);
        \node[anchor=west,align=left] at ($(legend-start) + (1, -4.8)$) {\small Hypothesized behavioral effects};

\end{tikzpicture}
\end{subfigure}

\begin{subfigure}{0.4\textwidth}
    \caption{\label{fig:example_associations}}
    \resizebox{\textwidth}{!}{\input{figures/Fig2a_new}}
    
\end{subfigure}\hspace{1em}%
\begin{subfigure}{0.4\textwidth}
    \caption{\label{fig:example_siblings}}
    \resizebox{\textwidth}{!}{\input{figures/Fig2c_new}}
\end{subfigure}

\caption{\label{fig:dags}The quantity of speech attributed to each speaker (``CHI'', ``OCH'', ``FEM'', ``MAL'', i.e. the key child, other children, female adults, and male adults) in each recording by an algorithm only indirectly reflect the true quantities. In reality, speaker classification errors can distort measurements and create spurious correlations in the quantities of speech attributed to each speaker. \textbf{(\ref{fig:example_fem_prop}) Measurements of speech quantities.} The nature of the input to children may be misrepresented as a result of classification errors. For instance, the proportion of female adult speech can be distorted due to incorrect inferences about the speaker's type and gender. \textbf{(\ref{fig:example_associations}) Associations between speakers.} An increase in female adult speech may trigger an increase in detected amounts of both female adult (black arrow) \textit{and} child speech (red arrow), and vice-versa, creating the appearance of an association between the two speakers. \textbf{(\ref{fig:example_siblings}) Effect of independent variables on speech quantities.} Spurious associations can also affect inferences about the effect of independent variables on speech behavior. For example, we might draw incorrect conclusions about the existence and direction of an effect of siblings on the quantity of speech received from adults (dashed lines) if speech from siblings is incorrectly classified as adult speech (then, children with siblings might falsely appear to receive more input from adults). }
\end{figure}

In the literature, these measurement errors are generally considered as random noise and left untreated. This ignores the fact that the quantity of speech attributed to a given voice type can be systematically affected by the quantity of speech from others (Figure \ref{fig:dags}). For instance, the proportion of female adult speech can be overestimated due to children being confused with a female adult, or distorted due to male and female adults being confused with one another (\ref{fig:example_fem_prop}). Specifically, in our data, we will show that \gls{vtc} and \gls{lena} systematically distort the proportion of female adult speech. In addition, since these algorithms have different error-patterns, \gls{lena} generally reports higher proportions of female speech than \gls{vtc}\footnote{As an example, the data we introduce in the Methods section contains a day-long recording where \gls{lena} finds fewer than 200 vocalizations from male adult(s) and \gls{vtc} finds more than 2000}; this means that comparing and aggregating findings obtained with different algorithms is highly challenging.

Additionally, and perhaps more surprisingly, classification errors can also affect estimates of the association strength in the quantities of vocalizations attributed to different voice types, and even with other variables. To see why, it is useful to frame the problem in terms of causal diagrams (specifically Directed Acyclic Graphs, or DAGs), in which causal relationships between variables are represented by directed arrows. Causal paths introduce correlations between variables: for example, in the causal chain `$\text{exposure } [e]\to \text{mediator } [m]\to \text{outcome } [o]$' (describing a model where an exposure causes a mediator which itself causes an outcome), $e$, $m$ and $o$ will all appear to be correlated. In this framework, we may distinguish relevant and biasing causal paths: ``[relevant] causal paths start at the exposure [$e$], contain only arrows pointing away from the exposure [$e$], and end at the outcome [$o$]'' --- that is, ``they have the form $e \to x_1 \to \dots \to x_k \to o$.''. By contrast, ``biasing paths are all other paths from exposure to outcome, [e.g.] $e \leftarrow x_1 \to \dots \to x_k \to o$'' \citep{textor2015drawing}. In the latter case, $x_1$ is typically called a ``\textit{confounder}''. From this perspective, classification errors open ``biasing paths'' that can create completely spurious correlations between variables in sometimes unpredictable ways. This potentially leads us to over-/under-estimate the effect of an association; to believe that an association between two variables exists when it does not; or worse, to reach incorrect conclusions about the \textit{sign} of an effect. For instance, in Figure \ref{fig:example_associations}, female adult speech incorrectly labeled as child speech may lead to spurious correlations between our measurements of adult speech (input) and child speech (output).  In Figure \ref{fig:example_siblings}, the effect of siblings on adult input may be similarly distorted by classification errors, which open up a completely spurious biasing path between ``siblings'' and female/male adult speech.


\begin{figure}
\centering
\begin{subfigure}[t]{0.29\textwidth}
    \includegraphics[width=1\linewidth]{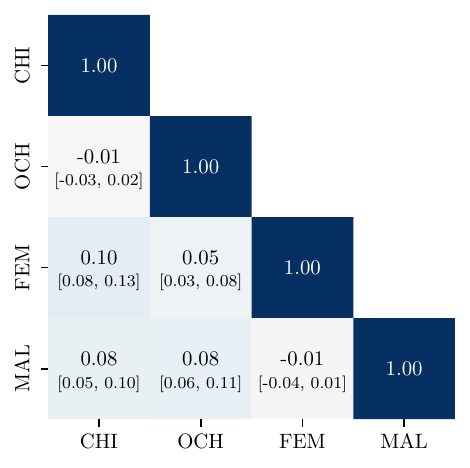}
    \caption{Human annotations}
\end{subfigure}\hspace{0.1cm}%
\begin{subfigure}[t]{0.29\textwidth}  
    \centering 
    \includegraphics[width=1\textwidth]{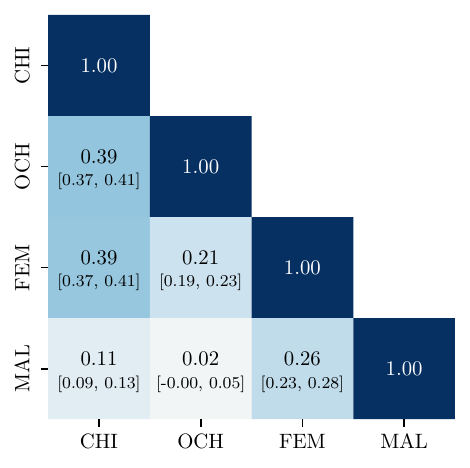}
    \caption{VTC}
\end{subfigure}\hspace{0.1cm}%
\begin{subfigure}[t]{0.29\textwidth}  
    \centering 
    \includegraphics[width=1\textwidth]{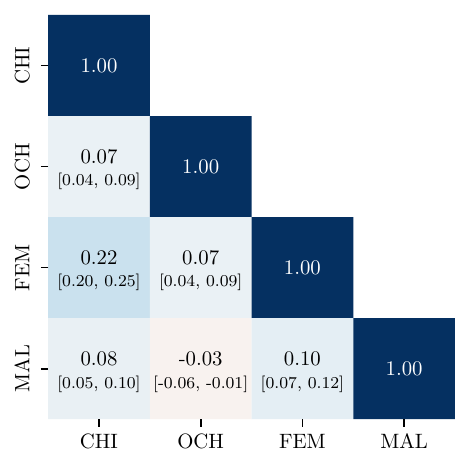
    }
    \caption{LENA}
\end{subfigure}
\caption{\label{fig:clips}Correlations between speakers' vocalization counts in $6638 \times 15$s audio clips according to human, \gls{vtc}, and \gls{lena} annotations. Estimates are generally inconsistent across the three.}
\end{figure}

As a brief illustration of the practical significance of these issues, we compare correlations between speaker vocalizations across 6638 $\times$ 15-second audio clips (from English-speaking corpora described in Section \S\ref{section:data}) based on human, \gls{vtc}, and \gls{lena} annotations (Figure \ref{fig:clips}). Manual annotations reveal statistically significant but low correlations between speakers across clips (Pearson $R \leq 0.10$). In contrast, \gls{vtc} and \gls{lena} produce much larger correlations, and are inconsistent with each other. For instance, \gls{vtc} reveals a medium correlation between CHI and OCH (Pearson $R=0.39$, $p<0.001$, $N=6638$), while \gls{lena} finds a weak correlation (Pearson $R=0.07$, $p<0.001$, $N=6638$), and manual annotations find no discernible correlation whatsoever (Pearson $R=-0.01$, $p=0.527$, $N=6638$). As we will show in \S\ref{section:confusion_matrix}, these discrepancies are consistent with biasing paths resulting from speaker misclassification.

\subsection{Previous relevant work}

Thus, while the accuracy, reliability, and validity of \gls{lena} has been well documented in previous work for many languages (for meta-analyses see \cite{cristia2021thorough,wang2020meta}; see also \cite{bastianello2023language,bruyneel2021validation,mcdonald2021evaluating,cristia2024establishing}), the downstream effect of classification errors, for example on correlational analyses, has not been properly assessed. In the context of long-form recordings for language acquisition, the issue was briefly raised in \citet{Cristia2023}, a descriptive report on 38 children's language input and output, and in which an attempt was made to discern whether correlations observed between speakers' speech quantities were entirely consistent with classification errors.

In parallel, the downstream effects of errors in automated predictions (such as classification errors) have become more widely recognized in other disciplines, due to the increasing recourse to machine learning for data processing. Most prominently, \citet{Angelopoulos2023} proposed a general approach to the issue (prediction-powered inference), which they have demonstrated on an array of datasets from different fields (biology, astrophysics, ecology, etc.). However, their strategy is currently limited to simple usages (e.g. ordinary least-squares linear regression) and lacks flexibility for complex hierarchical models such as those examined in the present paper\footnote{Their approach relies on a \textit{rectifier}, i.e, a function that compensates for measurement errors in the estimator of a quantity of interest (e.g., a linear or logistic regression coefficient) by leveraging manual annotations. 
However, deriving the rectifier is not a trivial task for an arbitrarily complex model. For very simple models, ready-to-use implementations have become available  \citep{angelopoulos2023ppi,Salerno2025}. Efforts to enhance the flexibility of this line of approach are ongoing \citep{miao2024task}.}. In addition, their approach requires human labeling for a substantial amount of observations. This makes it hard to transpose to the case of voice type classifiers, since no recording could be entirely annotated by hand (only a handful of clips at best). In a different context, \citet{teblunthuis2024misclassification} insisted that ``current practices of
`validating' [automated classifiers] by making misclassification rates transparent via metrics such as the F1 score [\dots] provide little safeguard against misclassification [\dots]''. Independently from us and building upon prior literature on measurement errors \citep[Ch.~8,15]{carroll2006measurement}, they proposed a solution (``Maximum Likelihood Adjustment'') conceptually similar to ours but limited to classification tasks\footnote{The authors present their method as an improvement over previous approaches such as Multiple Imputation (MI). In MI, multiple analyses are performed and pooled together given different imputations of missing or noisy data \citep{blackwell2017unified}.}. In a nutshell, they propose to average out statistical estimates over all possible true labels, weighted by their respective likelihood, given the labels predicted by a classifier and possibly other covariates. \citet{teblunthuis2024misclassification} set aside a Bayesian treatment for future work, acknowledging that ``[this] may provide additional strengths in flexibility and uncertainty quantification''. Therefore, the current paper makes a distinct contribution with respect to \citet{teblunthuis2024misclassification}, through a flexible Bayesian approach and by overcoming the restriction to classification tasks. Additionally, we are motivated to make these contributions accessible to a wider readership interested in behavioral research methods. 

\subsection{Present work}

The present paper seeks to make two contributions: First, highlighting the many effects of confusion errors; and second, providing a first approach to recovering from these errors. We study voice type classification errors by two popular algorithms: \gls{lena}, the historically dominant solution in the field, and \gls{vtc}, a state-of-the-art alternative gaining traction \citep{laudanska2025data}. Both were applied to the very same longitudinal audio data (1\,401 recordings of $\sim$8 hours each from 217 children across six corpora; see Section \S\ref{section:data}), allowing us to document the effects of classification errors on vocalization quantities, but also on a variety of theorized language acquisition factors. To better inform our readership, we assess the consequences of classification errors on a variety of research problems (Figure \ref{fig:dags}): (a) direct measurement of speech quantities (in our example, the proportion of female adult input); (b) measurements of associations between speech quantities (the short and long term effects of input on output); and (c) measurements of the effect of an independent variable on speech quantities (respectively, the effect of the child's age on its vocal production, and the effect of sibling number on the child's experience). This calls for a multi-level hierarchical model of speech behavior which allows us to exploit the longitudinal aspects of included data, assuming that many researchers will be interested in both developmental effects (e.g. does more adult input lead children to vocalize more over time?) as well as household effects (does the presence of siblings lead to changes in adult vocalization counts?). 

At the heart of our approach is the requirement that such a theoretically-driven model of speech behavior be complemented with a model of how the algorithm behaves. We inform our algorithm behavior model with $\sim$28h of human-annotated data, which allows us to model the relationship between the \textit{true} but unobserved vocalization counts of each different talker type (male adults, female adults, other children, and the key child) and the vocalization counts provided by the diarization algorithms in the same data. In other word, even though there is no data directly showing how counts by talker type are misestimated at the recording level, our model can learn from errors occurring in the shorter time-scales that have been human-annotated.

As a brief summary, we find that classification errors can significantly distort estimates of speech quantities and effect sizes in downstream analyses, sometimes even leading to incorrect conclusions about the existence, size, and (possibly) the direction of a correlation/effect. Furthermore, the impact of classification errors varies depending on the precise effects being estimated. For instance, while the effect of having siblings on the quantity of speech from each speaker is significantly distorted by measurement errors, distortions are smaller for developmental effects unfolding throughout child development, such as the long-term effect of input on output. 

In addition to being the first to highlight these wide-spread and complicated effects of confusion error on potential scientific inferences, this paper sought to make a second contribution, by assessing whether the proposed joint model approach may also help recover unbiased estimates. We provide a range of evidence suggesting that joint modeling does improve things. For instance, we looked at the extent to which \gls{lena} and \gls{vtc} yielded the same results: If our joint model approach works to unbias the estimates, then the two algorithms' output should come closer to the underlying truth. We indeed found that our approach improves the agreement between the two algorithms in all cases except one. While this is encouraging, we believe that the resulting Bayesian calibration model for producing unbiased estimates of correlations and effect sizes using algorithmic vocalization data may be most useful to technically proficient readers. We also provide a Python package enabling scientists to simulate the impact of classification errors on their own analysis \citep{diarization-simulation}, which may be easier to adopt while still requiring familiarity with scripting.

\subsubsection{Outline of our methodological approach}

The following is a high-level description of our methodological approach, with details being provided in the Methods section. We start by specifying a model of speech behavior that embeds our theoretical assumptions and/or the hypotheses we evaluate. Each researcher may posit a different model of speech behavior, depending on their own research questions. For the purposes of this paper, our model of speech behavior specifies that, at the recording level, children's input composition may vary as a function of how many siblings the child has; and the key child's output may vary as a function of their age, input, and random individual variation. This model of speech behavior is not a main contribution of this paper, but rather a necessary component of our approach allowing us to illustrate the downstream consequences of classification errors on reasonably motivated analyses.

Next, we embed this model of speech behavior in a larger model that also takes into account the  fact that we do not observe children's ``REAL'' number of adult vocalizations, but instead estimate them through an algorithm. To this end, we include the effect of the algorithm in the data generating process, treating the true vocalization counts as latent (i.e. unobserved) variables. Fortunately for the current research, several datasets have been partially annotated by humans, which means that we have reasonably accurate gold standard estimates of how many vocalizations were uttered by the key child, other children, as well as male and female adults for very small extracts out of the long-form recordings. For this initial foray in studying how algorithmic errors may affect scientific conclusions, we assume that the most relevant features of the algorithm relate to the model's tendency to miss vocalizations, assign them to the wrong speakers, as well as incorrectly break up or lump vocalizations.

Using a Bayesian approach, then, our task becomes estimating each of the parameters in the joint model, using audio clips for which we have both algorithmic and human annotations to inform the model of the algorithm, and using recording-level automatic annotations to inform parameters related to our model of speech behavior. We note that this does not allow us to ``correct'' the individual classification decisions made by diarization algorithms or the resulting segmentation files (e.g., the .its returned by \gls{lena}). Calibration is intended to improve, at the recording level, the estimates of aggregate quantities of vocalizations and the estimates of their relationship, within the context of a specified model of speech behavior.\footnote{That is, the posterior density of the true quantity of vocalization can never be constructed without assuming a model of speech behavior (because $P(\hat{D}|D_{\text{meas}})\propto P(D_{\text{meas}}|\hat{D})\times P(\hat{D}|\theta)P(\theta)$, where the first factor is informed by the algorithm behavior model and the second is informed by the speech behavioral model).} This approach is highly flexible, since it decouples the statistical model of speech behavior from the model of the algorithm's behavior: both can be refined in parallel as two distinct modules, regardless of their respective complexity. Finally, as an alternative to Bayesian calibration, the joint model can be used in simulations for traditional null hypothesis testing.  

The technical details are elaborated in Section (\S\ref{section:method}). Readers mostly interested in our findings and their implications may jump to the Results section directly (\S\ref{section:results}).

\section{\label{section:method}Method}

In this Section, we introduce the Bayesian calibration approach (\S\ref{section:probabilistic_calibration}), a simulation--based alternative to Bayesian calibration (\S\ref{section:simulations}), and finally, the data used in the present study (\S\ref{section:data}).

\subsection{\label{section:probabilistic_calibration}Bayesian calibration of algorithmic vocalization counts}

\subsubsection{Principle}

Traditionally, scientists directly fit a model of speech behavior against automated measurements of speech data ($D_{\text{meas}}$), producing estimates and/or posterior distributions for parameters of interest entering the model (e.g. they estimate $P(\theta|D_{\text{meas}})$, where $\theta$ may be the strength of the effect of age on children's output, or the effect of siblings on the quantity of input, etc.). However, as we explained above, when $D_{\text{meas}}$ is contaminated with classification errors (as is the case for vocalization counts derived from diarization algorithms), treating it as truth can lead to incorrect conclusions. We therefore propose a Bayesian calibration approach for learning from observations affected by classification errors. As we shall see below, a Bayesian framework offers a natural response to this challenge, by simultaneously leveraging prior theoretical knowledge, manual annotations, and automated annotations into a unified approach.

Bayesian calibration combines the assumed model of speech behavior with a model of the algorithm's behavior into a single larger model. Under the hood, this approach infers the probability distribution of the unobserved \textit{true} amounts of vocalizations of each speaker ($\hat{D}$) given the algorithm output ($D_{\text{meas}}$), and plug those estimates into the model of behavior rather than $D_{\text{meas}}$. If $\theta$ are the parameters of interest of the model, $D_{\text{meas}}$ are the data produced by the algorithm, $\hat{D}$ are the true but unobserved quantities, and $\nu$ are nuisance parameters, then the posterior distribution of $\theta$ (the parameter(s) of interest) given the observed data $D_{\text{meas}}$ is:

\begin{equation}
    \label{eq:calibration1}
    P(\theta|D_{\text{meas}}) \propto \int \textcolor{blue}{\underbrace{P(\hat{D}|\theta) P(\theta)}_{\text{speech behavior}}} \textcolor{red}{\underbrace {P(D_{\text{meas}}|\hat{D},\nu)P(\nu)}_{\text{algorithm behavior}}} d\hat{D} d\nu 
\end{equation}

The posterior distribution is factored into two terms. The first term encodes the assumed model of speech behavior. The second term encodes the algorithm's behavior, expressed as the probability that the algorithm produces quantities  $D_{\text{meas}}$ (e.g., the number of vocalizations attributed to each speaker) given the true unobserved quantities $\hat{D}$ -- if the algorithm was perfect, we would have $P(D_{\text{meas}}|\hat{D})=1$ if and only if $D_{\text{meas}}=\hat{D}$, and 0 otherwise). The nuisance parameters $\nu$ characterize the algorithm's behavior (typically, the classifier's confusion matrix), and are a priori unknown. To learn  $\nu$, one must add calibration data, for which both the algorithm's output ($D_{\text{calib}}$) and the ground truth ($\hat{D}_{\text{calib}}$) are observed. In an ideal world, every speaker in a recording would be wearing a device that detects their vocalizations -- regardless of how soft or how confusable with those of others. In our real world, however, we do not have access to this ground truth but rather to human annotations (which may miss vocalizations or confuse speakers). Incorporating a consideration of human annotations as our ``ground-truth'' allows us to decompose the algorithm behavior as follows:

\begin{equation}
    \label{eq:calibration2}
    P(\theta|D_{\text{meas}},D_{\text{calib}},\hat{D}_{\text{calib}}) \propto \int \textcolor{blue}{\underbrace{P(\hat{D}|\theta) P(\theta)}_{\text{speech behavior}}} \textcolor{red}{\underbrace {P(D_{\text{meas}}|\hat{D},\nu)P(\nu|D_{\text{calib}},\hat{D}_{\text{calib}})}_{\text{algorithm behavior}}} d\hat{D} d\nu 
\end{equation}

To the extent that the model of the algorithm's behavior is valid \textit{and} the ``ground-truth'' data is correct, the posterior distribution of $\theta$ retrieved by computing this integral will be unbiased. However, the uncertainty induced by the behavior of the algorithm will widen the posterior distribution of $\theta$, given that different values of $\hat{D}$ (the true quantities of speech) are compatible with the algorithm output $D_{\text{meas}}$; in other words, this approach trades bias for variance.

\subsubsection{\label{section:speech_model}Model of speech behavior}

In Bayesian calibration, as in traditional regression analysis, we must start by specifying a model of speech behavior, encoding the relevant underlying processes (whether they operate at the a cognitive or social level). We propose a multi-hierarchical model that implements several assumptions (see Figure \ref{fig:partial_model} for a visual representation, and Appendix \S\ref{appendix:model_assumptions} for justifications). First, the quantity of vocalizations by each speaker class (CHI, OCH, FEM, and MAL) is thought to potentially vary across children. Second, the number of siblings a child has may affect speech quantities by OCH and ADU (i.e., FEM and MAL), but not CHI directly. Third, we also assume a random child-specific effect of development on children's speech output. Specifically, our model assumes that children's speech quantities at birth are equivalent (i.e. individual newborns do not differ from each other), with random individual variation emerging as children age (using a Generalized Linear Model with a log link function). Finally, the model also assumes a long-term effect of adult input on children's output (i.e., an effect of adult speech at a child-level that interacts with the children's age). All effects of age are assumed to be log-linear, up until a threshold (24 months) after which they plateau (this threshold was validated via a change-point model). The precise model specification, including the priors on every parameter, is described in Section \S\ref{section:behavior_model_specification}. 

\begin{figure}[h]
    \centering
    \input{figures/behavior_model}
    \caption{\label{fig:partial_model}Model of speech behavior. Observed variables (vocalization counts for each speaker and recording, child age, and siblings number) are shown in blue, latent variables in white. Indices $k$ designate recordings, and $c$ designates a child.$v_k^{\text{recs}}$ is the vocalization count of each speaker class in each recording. Variables $\mu$ represent the expected speech rates per speaker at each level (population, corpus, and child). $\alpha_{c}^{\text{dev}}$ is the random effect of age on the children's output (which is assumed to be distributed around a mean value $\alpha_{\text{dev}}$). It is also assumed that the expected quantity of adult speech at the child level has a long-term effect on children's speech ($\beta^{\text{dev}}$), which interacts with age. }
\end{figure}

\subsubsection{\label{section:algorithm_model}Model of the algorithm's behavior}

The calibration approach requires a model of the algorithm in terms of the probability that it outputs certain values given the unobserved true amount of vocalizations of each speaker class $i\in\{\text{CHI},\text{OCH},\text{FEM},\text{MAL}\}$, which will be further referred to as $v_i$. We assume that each of the vocalizations from each speaker $i$ causes the algorithm to attribute a random amount (0, 1, 2, \dots) of vocalizations to each speaker $j$, resulting in a total of $n_{ij}$ vocalizations attributed to $j$ as a result of the true vocalizations from $i$. The only observable quantity is, in fact, $n_{j}=\sum_i n_{ij}$, the total amount of vocalizations attributed to each speaker $j$ by the algorithm. Different assumptions could be made about how  $n_{ij}$ is generated, given $v_i$, the unobserved true amount of vocalizations from each speaker $i$. For instance, one can assume a binomial process, such that the $v_i$ vocalizations are detected and attributed to $j$ with probability $\lambda_{ij}$ (that is, $n_{ij}\sim\text{Binomial}(v_i,\lambda_{ij})$). However, some vocalizations are detected as not one but multiple vocalizations (the algorithm breaks them down into multiple segments), which a binomial process would fail to capture. We therefore consider a generalized Poisson distribution (from \citealt{Efron1986}), such that $n_{ij}\sim \text{DPO}(\lambda_{ij} v_i, \tau)$ with mean $\lambda_{ij}v_i$ and variance $\lambda_{ij}v_i/\tau$\footnote{We assume $\tau\sim\text{Exponential}(1)$, such that the model can accomodate both under- and over-dispersion. We use the Stan implement of the Double Poisson distribution proposed in \citet{pustejovsky2024doublePoissonStan}. We find $\tau\sim 1.4$ for \gls{vtc} and $\tau\sim 1.8$ for \gls{lena}, corresponding to underdispersion.}. 

In this model, $(\lambda_{ij})$ is the confusion matrix of the algorithm; the diagonal $(\lambda_{ii})$ measures the rate of true positives, and the non-diagonal elements $\lambda_{i,j\neq i}$ measure the rate of false positives due speaker misidentification. The confusion rates $\lambda_{ij}$ are assumed to vary from one recording to another (due to unpredictable variations in recording conditions for instance), such that $(\lambda^{k}_{ij})$ (the confusion rates for a particular recording $k$) are drawn from Gamma distributions with means $\mu_{ij}$ and shapes $\alpha_{ij} \sim \mathrm{Pareto}(1,1.5)$ (truncated to values $\geq 1$\footnote{$\alpha$ is difficult to identify, especially for low values of $\mu$. Ideally, we would need more extensive human annotations within each manually annotated recording. In this work, we limited ourselves to extant annotations.}). $\lambda$, $\mu$, $\alpha$, and $\tau$ are nuisance parameters that can be partially learned from calibration data. 

Ultimately, we implement the model in Stan \citep{carpenter2017stan}, which uses a variant of Hamiltonian Monte-Carlo and therefore cannot directly sample latent discrete parameters such as $n_{kij}$\footnote{See \url{https://mc-stan.org/docs/stan-users-guide/latent-discrete.html}.}. In the case of the calibration data, for which a ``ground truth'' ($v_{ki}$) is known, the solution is to marginalize over the discrete latent parameters $n_{kij}$, which comes down to computing the sum \eqref{eq:poisson_model}:

\begin{equation}
    \label{eq:poisson_model} P(n_{kj}=n|v_{k1},\dots,v_{kC}) = \sum_{0\leq n_{kij}\leq n} \prod_{i=1}^{C} P(n_{kij}|v_{ki})\cdot \delta(n-\sum_{i=1}^{C} n_{kij})
\end{equation}

The joint knowledge of the algorithm's vocalization counts $(n_{kj})$, and the true counts $(v_{ki})$ in manually annotated clips of audio allows us to learn the distribution of  confusion rates ($\lambda$) across recordings. There is one computational caveat: the amount of combinations to be summed over becomes combinatorially large for large values of $v_{i}$. Therefore, manually annotated clips are split into windows of 15s, which keeps $v_{i}$ reasonably small with each window (originally, the duration of manually annotated clips varied between 15s and 5 minutes, always being a multiple of 15s). In fact, larger temporal windows would be less informative. However, we do not think it wise to use even shorter time windows, since it would introduce boundary effects, and we found 15s to be a good compromise between tractability, information-value, and bias.

For most of the audio, only automated annotations are available, and the true values $v_{ki}$ are unknown: they are latent parameters, and the goal is precisely to measure their distribution. 
To this end, we approximate $v_{ki}$ as a continuous parameter -- which Stan can conveniently sample from -- and assume that:

\begin{equation}
    n_{kj}\sim \mathrm{DPO}(\sum_{i=1}^{C} \lambda_{kij} v_{ki}, \tau)
\end{equation}

where DPO designates the generalized Poisson distribution from \citet{Efron1986}.

\paragraph{Limitations}
As with any model, the above simplifies reality in numerous ways. First, the effect of true vocalizations from each speaker is assumed to add linearly, which might not be a good approximation in case of overlapping speech. This could slightly bias estimates of correlations between input and output if those were driven by dense interactions over short time scales. In fact, the model ignores the fact that \gls{lena} does not handle overlap between speakers. Second, confusion rates are assumed to be random and independent of variables such as the child's age, or population level effects (e.g. differences in environment or language). Yet, if, say, an algorithm detects vocalizations from older children more accurately, this could lead to overestimating the increase in output over time. Our approach (just like that of \citealt{teblunthuis2024misclassification}) can in principle account for such biases. However, we found that we did not have access to enough human annotations to incorporate these effects directly into our full model, due to poor identification. Nevertheless, we independently assessed the effect of the child's age and the environment (urban vs rural) in Appendix \S\ref{section:confusion_covariates}, which did not reveal unambiguous and significant effects (see also \citealt{peurey2025fifteen}). We recommend that additional work reflects on the potential consequences of confusion rate depending on these factors. Finally, our approach relies on the assumption that human annotations provide a reliable ground truth, but humans make mistakes. Agreement between human annotators in typical longform recordings has been reported in terms of the average F-score\footnote{The F-score is the harmonic mean of precision ($\text{true positives}\over\text{true positives+false positives}$) and recall ($\text{true positives}\over\text{true positives+false negatives}$).} (across speaker types), yielding $F_1\sim0.70$, which is far from perfect \citep{kunze2024challenges}. In a subset of our dataset, we evaluate the agreement between raters in terms of intra-class Correlation Coefficient (ICC) (by comparing vocalization counts estimates from different raters for multiple 15-second audio clips) and find $\mathrm{ICC}\sim 0.7-0.9$ for humans versus humans on average, instead of 0.2-0.6 for humans versus \gls{vtc}/\gls{lena} (Table \ref{tab:icc_agreement}, Appendix \S\ref{appendix:icc}). Relatedly, our model does not accept overlapping annotations from multiple annotators. In principle, an approach combining our model with annotation models \citep{paun2018comparing} such as \citet{dawid1979maximum} and \citet{Albert2004} would have the simultaneous abilities of acknowledging human errors and aggregating redundant annotations. However, given the sparsity of human annotation available in the context of longform recordings, and the added computational complexity, such strategies may prove highly challenging for this task in practice\footnote{Instead, as a mere starting point, we suggest a simple, albeit naive step for accepting redundant annotations by making the following change to the model: when a unique audio clip $k$ has been annotated by $A>1$ individuals, assume that there exists $1\leq a\leq A$ such that $\bm{v_{ka}}$ are the true vocalization counts for this clip, and draw algorithmic counts accordingly. Marginalizing over all possible $a$ gives: \begin{equation}
    p(\bm{n}_k) = \sum_{a=1}^A p(\bm{n}_k|\bm{v}_{ka})p(\bm{v}_{ka})
\end{equation}. This minimal form of aggregation, however, does not dispense with the assumption that at least one annotator is correct.}. Even if these challenges were addressed, however, humans tend to make similar errors due to biases (such as attributing loud child vocalizations to the key child regardless of their true origin) which constrains the ability of annotation models to reliably infer truth from agreement patterns.

\begin{figure}[h]
    \centering
    \input{figures/full_model}
    
    \caption{\label{fig:full_model} Combined model of speech behavior and of the algorithm behavior. Compared to Figure \ref{fig:partial_model}, the actual quantity of vocalizations $(v^{\text{recs}})$ is treated as latent variables. Colored arrows represent the effect of real vocalizations from each speaker (e.g. CHI, in \textcolor{chi}{blue}) on the amount of vocalizations attributed to each speaker by the algorithm $(n^{\text{recs}})$.  The unobserved confusion rates $\lambda_{kij}$ represent the rate at which vocalizations from a speaker $i$ are detected and attributed to a speaker $j$ in recording $k$. The distribution of $\lambda_{kij}$, parameterized by $\mu_{ij}$  and $\alpha_{ij}$, is learned via calibration data (for which both the true counts $n^{\text{clips}}$ and the algorithmic counts $v^{\text{clips}}$ are known).}
\end{figure}

\subsubsection{The joint model}

Figure \ref{fig:full_model} shows the combination of the models of speech behavior and of the algorithm. It visually emphasizes how algorithmically derived vocalization counts no longer inform the model of speech behavior directly; rather, they inform the latent true quantity of speech, which in turn informs the behavioral model. The relationship between the true and the measured vocalization counts is itself informed by the calibration data. We reiterate that this approach is highly flexible, given that the model of speech behavior and the model of the algorithm can be elaborated, reused, and improved separately as independent modules (in contrast to \cite{Angelopoulos2023}, which is practically limited to simple models, and which requires deriving specific solutions for every estimator). 

\paragraph{Validation}

First, we considered the algorithm model in isolation. We evaluated its capacity to predict the output of each algorithm (\gls{vtc} and \gls{lena}), given human estimates of vocalizations for each speaker in manually annotated clips (Appendix \S\ref{section:validation}). We found that the prediction of our model did align with the actual automated measurements, which means that our model captures the relationship between human and automated vocalization counts. Additionally, fitting the model to simulated calibration data (for which, by design, the confusion matrix was known) confirmed that the model correctly identified the true values of the parameters (Appendix \S\ref{section:simulation-validation}).

In a second step, we evaluated the calibration procedure as a whole. When fitting the model to real-data, the calibration procedure generally improves the correlation between the vocalization counts estimated by \gls{lena} and \gls{vtc} on average (see Appendix \S\ref{section:vtc_lena_comparison}, Table \ref{table:correlation_vtc_lena}). Notably, calibration increases the correlation ($R^2$) between the quantity of female adult speech measured by \gls{lena} and \gls{vtc} from 0.62 to 0.73. 

Most importantly, we compared manual vocalization count estimates (obtained by extrapolating partial human annotations) with the automated estimates derived before and after calibration. Prior to calibration, automated estimates are severely biased (Figures \ref{fig:truth_human_vtc} and \ref{fig:truth_human_lena}, Appendix \S\ref{section:vtc_lena_comparison}).
Calibration increases the agreement between manual and automated vocalization counts estimates, with statistically significant improvements for CHI, FEM and MAL under both \gls{vtc} and \gls{lena}. For female adult speech, the ICC increased dramatically, from 0.15 to 0.85 in \gls{vtc} and from 0.02 to 0.81 in \gls{lena}. This is despite the fact a perfect match cannot be expected, since manual estimates based on partial annotations do not provide ground truth values for whole recordings, and because of the stochasticity of algorithmic errors.

Finally, these validation strategies were complemented by a simulation study, which enabled us to assess the ability of the calibration procedure to reliably estimate vocalization counts and parameters of interest when the true values are known. To this end, we chose plausible values for every parameter of the model of speech behavior, then simulated 1000 observations (200 children $\times$ 5 recordings each). We simulated automated vocalization counts for each recording by modeling confusion errors (using a strategy described in detail in the next section) and applied the Bayesian calibration on these synthetic corpora. This revealed that Bayesian calibration systematically improves the accuracy of automated vocalization counts (Table \ref{table:truth_vs_synthetic}, Appendix \S\ref{appendix:calibration_validation_simulation}) and generally improves the estimates of quantities and effect sizes of interest (Figure \ref{fig:synthetic}). The simulation study gives us an opportunity to assess whether the posterior distributions obtained via Bayesian calibration  correctly reflect the uncertainty in the quantities of interest -- specifically, in our example, the vocalization counts per synthetic recording -- \citep{Cook2006}. The posterior distributions are properly calibrated if their credible intervals achieve nominal coverage -- that is, an X\% credible interval contains the true value X\% of the time. Calibration curves in Figure \ref{fig:truth_human_calibration} (Appendix \S\ref{appendix:calibration_validation_simulation}) reveal that the posterior distributions are indeed very well calibrated, with perhaps the exception of male adult speech in \gls{vtc}, where the posteriors are a bit too wide.


\subsection{\label{section:simulations}An alternative to calibration: simulations for sensitivity analysis and null-hypothesis testing}

Bayesian calibration can be computationally demanding, particularly for large datasets. Moreover, it is sometimes unnecessary: certain measurements are only marginally affected by classification errors, making the added complexity unwarranted. In such cases, simulation-based approaches are straightforward alternatives that can effectively assess the sensitivity of an analysis to classification errors or evaluate whether a finding could be explained by classification errors alone (see Figure \ref{fig:simulations}). Like Bayesian calibration, however, simulations still require (1) a model of speech behavior and (2) a model of the algorithm's behavior.

In a simulation-based approach, the first step is to generate synthetic datasets reproducing key characteristics of the actual data to be analyzed (including the amount of participants and observations), by sampling from the assumed model of speech behavior. Synthetic data can be simulated by fixing the value of one or several parameters of interest, say  $\hat{\theta}$ (for instance, by considering a null-hypothesis, $\hat{\theta}=0$). This gives $\hat{D}$, the synthetic ``true'' vocalization counts. The behavior of the algorithm itself is then simulated, by drawing samples from the corresponding model, yielding $D_{\text{meas}}$ (the vocalization counts as they would be reported by the algorithm). The resulting synthetic datasets can then finally be processed within an analysis pipeline (e.g. a linear regression), producing an estimate of $\theta_{\text{meas}}$. This estimate can be compared to the known true value ($\hat{\theta}$). This procedure can be repeated for different values of $\hat{\theta}$. If the difference between $\mathbb{E}(\theta_{\text{meas}})$ and $\hat{\theta}$ is generally negligible, then algorithmic errors do not seriously bias inferences. We illustrate this approach with the measurement of the proportion of female adult speech in Section \ref{section:female_proportion}. Whenever simulations indicate significant vulnerability to classification bias, one can consider the Bayesian calibration procedure described above (Section \S\ref{section:algorithm_model}) for deriving unbiased estimates of $\theta$. Alternatively, in null-hypothesis testing settings, simulations can assess whether a finding (e.g., a positive correlation between two speakers) is compatible with the effect of classification errors alone. We provide a Python package to facilitate the implementation of such simulations for \gls{vtc} and \gls{lena}, including tutorial examples in Python and R \citep{diarization-simulation}.

\begin{figure}[h]
    \centering
    \begin{equation*}
        \hat{\theta} \ \textcolor{blue}{\xrightarrow[\text{speech behavior}]{\text{simulate}}} \ \hat{D} \ \textcolor{red}{\xrightarrow[\text{algorithm}]{\text{simulate}}} \ D_{\text{meas}} \ \xrightarrow[\text{analysis pipeline}]{\text{run}}  \ \theta_{\text{meas}} \neq \hat{\theta}?
    \end{equation*}
    \caption{Process for assessing the impact of classification errors on a particular analysis pipeline. A true value $\hat{\theta}$ for a parameter of interest is drawn at random. The model of speech behavior is simulated given $\hat{\theta}$, which yields synthetic ground truth data, $\hat{D}$. The behavior of the algorithm is simulated, thus returning synthetic \textit{algorithmic} data, $D\neq \hat{D}$. Finally, the analysis pipeline is run on $D_{\text{meas}}$. If $\theta_{\text{meas}}$ significantly departs from $\hat{\theta}$, then the process is sensitive to algorithmic bias.}
    \label{fig:simulations}
\end{figure}

\subsection{\label{section:data}Data}

To inform the current exploration, we build on six corpora: Bergelson (English monolinguals, North America, \citealt{Bergelson2015}), Cougar (English monolinguals, North America, \citealt{vandam_homebank_2018}), LucCiD (English monolinguals, United Kingdom, \citealt{rowland_language_2018}), Kidd (English,  predominantly monolinguals, Australia,  \citealt{Donnelly2021}), Warlaumont (English-Spanish bilinguals, North America, \citealt{warlaumont_warlaumont_2016}), Winnipeg (mostly English monolinguals, with some French spoken, North America, \citealt{mcdivitt_mcdivitt_2016}), Fausey (English monolinguals, North America, \citealt{mendoza2022everyday,fausey-trio}). Dataset selection was constrained by a conjunction of imperatives: the data had to be longitudinal and to simultaneously contain the raw recordings, together with LENA \textit{and} VTC annotations. Additionally, we only included audio data between 10am and 6pm, and we excluded recordings which do not cover this time range entirely. This allows for more consistent comparisons across data points. The selected audio amounts to $\sim 11\,200$ hours total. 

The two algorithms we consider differ on many aspects, from their training data to their architecture, but for the purposes of conciseness, we focus here on the aspect of their behavior most likely relevant to the question at hand: Their propensity to miss vocalizations, and to confuse or co-activate speaker classes. \gls{lena} uses a Dirichlet Process Gaussian Mixture Model to classify audio frames into categories including key child speech, other speakers, noise, and silence. The system prioritizes precision over recall, meaning it tends to be conservative in identifying vocalizations to avoid false positives. VTC is an open-source neural network-based alternative to LENA that detects the same four speaker types but can handle overlapping speech and has been trained on multilingual data. Unlike LENA, VTC balances precision and recall by maximizing F-score, resulting in higher recall but lower precision.  For detailed technical specifications and implementation details, see \citep{xu2008signal,gilkerson2017mapping} for LENA and \citep{lavechin2020open,lavechin2025performance} for VTC.  

\begin{table}[!h]
\centering
\resizebox{\textwidth}{!}{\input{tables/data}}
\caption{\label{tab:corpora}Corpora used in the present analyses. Fausey-trio is only used for purposes of calibration (full recordings from this corpus are not considered in the model).}
\end{table}

Human annotation existed for snippets of five of our six key corpora, jointly covering only 0.23\% of the total audio duration. Most of our human annotation data (totaling 20h) comes from the ACLEW collaboration, and has been documented in \citet{Soderstrom2021}. Annotators were trained until they met stringent criteria on a gold standard, and all labs used the exact same annotation manual (the ACLEW DAS template, \citealt{ACLEW-DAS}), including definitions of what constitutes a vocalization. We employed here the so-called ``random sample'': Fifteen two-minute clips were randomly sampled from one recording from each of 9-10 children in four English-spoken corpora. The use of random sampling safeguards against any bias in data selection coming from the use of an algorithm. An additional 6.5h come from the VanDam-5-minute corpus \citep{vandam,carns2015question}, which has been less overtly documented, but the description on Homebank \citep{vandam_homebank_2018} provides sufficient details by explaining that three non-consecutive five-minute sections were sampled that had the highest child-adult conversational turns according to the automated \gls{lena} analysis. Annotators had access to \gls{lena} segmentation and could correct it, but did not have to, as their priority was to produce orthographic transcriptions of what was said. To further inform our analyses, we also included in-house human annotations on Fausey-Trio \citep{fausey-trio}, a seventh corpus that is otherwise not included in our analyses. Human annotation of this corpus was done independently of the present paper, with the purposes of contributing to a dataset with greater representation of male adults and other children. To this end, audio was sampled using 15-second long snippets and a loudness-based filter (i.e., independent from both \gls{lena} and VTC). Human annotators first listened to the snippets and only annotated them if there was at least one vocalization by adult males and/or other children (i.e., snippets with only key child and/or female adult vocalizations were not annotated). The annotation followed a simplification of ACLEW DAS focused only on segmentation, without transcription. In sum, most of the human annotation was done independently of the algorithms whose behavior is studied in this paper; and by design all four speaker classes are present in the human-annotated data. In total, the 27.6h of audio annotated by a human, VTC and \gls{lena} yielded 6638 $\times$ 15-second clips to be used for calibration purposes.\footnote{While assessing the impact of the recording environment on confusion rates (Appendix \S\ref{section:confusion_covariates}, Figure \ref{fig:confusion_environment}), we consider annotations from  corpora outside this table. To avoid reader confusion, those corpora and annotations are introduced in the relevant Appendix.}

\section{\label{section:results}Results}

We now turn to our main goal, the study of the effect of classification errors on downstream statistical analyses. In Section \S\ref{section:confusion_matrix}, we start by comparing the two speech-detection algorithms in our case-study (\gls{vtc} and \gls{lena}) against manual annotations. We will see that the two algorithms make errors that can explain the discrepancies between correlation estimates derived from human, \gls{vtc}, and \gls{lena} annotations (first evoked in Figure \ref{fig:clips}, Section \S\ref{section:the_case_of}). In Section \S\ref{section:effects}, we apply the Bayesian calibration strategy to our dataset. We compare the estimates of a variety of quantities and effect sizes relevant to language acquisition, using either manual annotations, or annotations from each algorithm, before and after applying our calibration strategy. We will see that without calibration, \gls{vtc} and \gls{lena} produce generally inconsistent estimates; using our calibration procedure, the tension is often reduced; in addition, calibrated estimates can significantly deviate from the naive estimates. However, in some cases, Bayesian calibration failed to resolve the disagreement between \gls{vtc} and \gls{lena}.
Finally, in Section \S\ref{section:female_proportion}, we show how simulations can help anticipate and diagnose biases resulting from algorithmic errors. 



\subsection{\label{section:confusion_matrix}Comparing \gls{lena} and \gls{vtc} against manual coding}

This section uses our joint model to assess how \gls{lena} and \gls{vtc} align with manual coding. This analysis addresses two questions: (1) how accurately do the automated methods detect speakers' vocalizations compared to manual coding, and how consistently they behave, and (2) how do the models' classification errors explain the distortions of correlations between speakers shown in Figure \ref{fig:clips}.

\begin{figure}[p]
    \centering
    \begin{subfigure}{\textwidth} 
        \centering
        \includegraphics[width=0.67\linewidth]{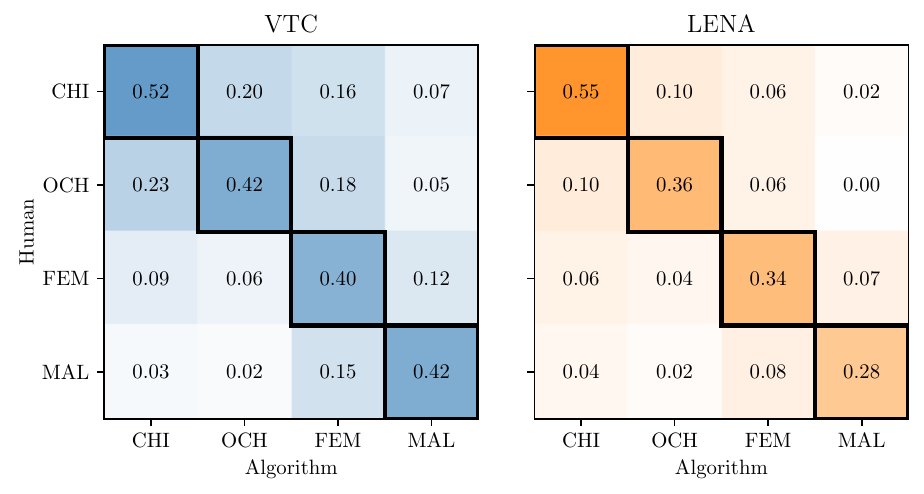}
        \caption{Average confusion rates ($\mu_{ij}$ in our algorithm model \S\ref{section:algorithm_model}) of \gls{vtc} and \gls{lena}. Rows indicate the true speaker, and columns indicate the speaker class attributed by the algorithm. Diagonal elements represent the true positive rate for each speaker. Non-diagonal elements represent the distributions of the rates of false positives.}
    \label{fig:confusion_matrix}
    \end{subfigure}

    \begin{subfigure}{\textwidth}
        \centering
        \includegraphics[width=0.8\textwidth]{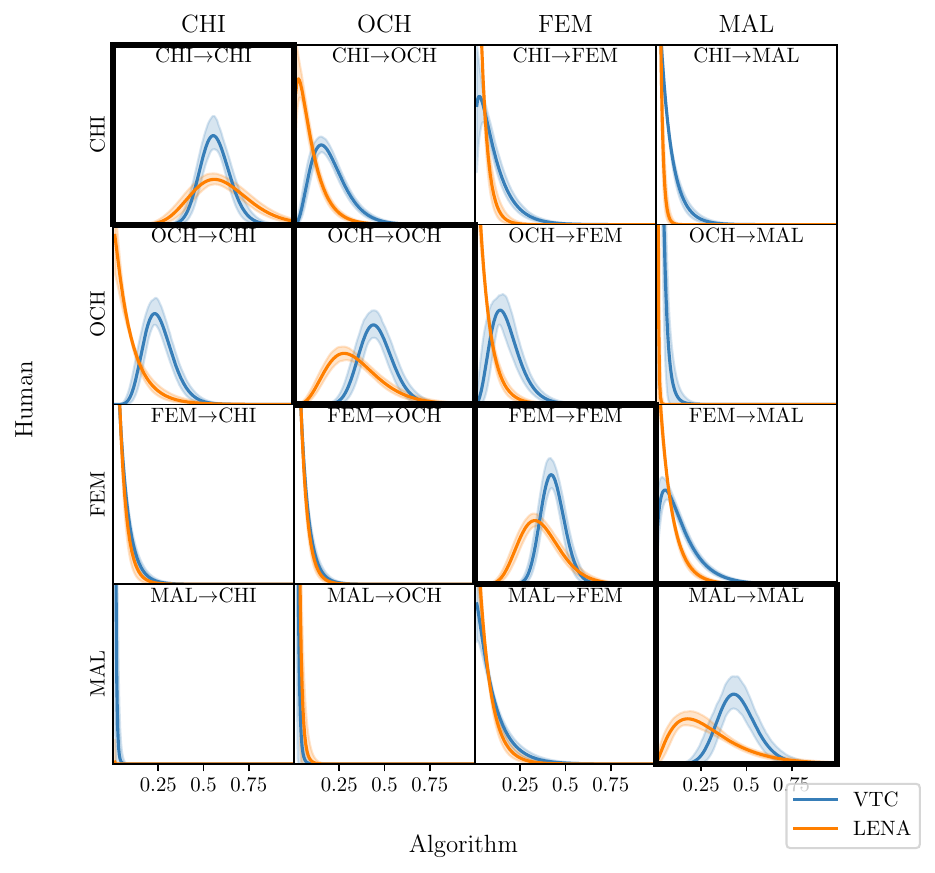}
        \caption{Distribution of confusion rates across recordings, for \gls{vtc} and \gls{lena} (i.e., the posterior distribution of $\lambda_{kij}$, in our algorithm model \S\ref{section:algorithm_model}). Rows indicate the true speaker, and columns indicate the speaker class attributed by the algorithm.  Diagonal elements represent the distributions of the rates of true positives for each speaker. Non-diagonal elements represent the distributions of the rates of false positives. Shaded areas represent the 95\% credible interval of the error rates' probability density.}
    \label{fig:confusion_matrix_density}
    \end{subfigure}
    \caption{Confusion rates of \gls{vtc} and \gls{lena}.}
\end{figure}

By fitting the calibration model introduced in Section \S\ref{section:algorithm_model} to the calibration data (i.e. clips of audio for which human annotations are available), we can evaluate the accuracy of classifications against human annotation. These can be represented in the form of confusion matrices, as shown in Figure \ref{fig:confusion_matrix}. The confusion matrices show that both algorithms confuse certain pairs of speakers more often: CHI and OCH (i.e. children) on the one hand, and FEM/MAL (i.e. adults) on the other hand. In addition, female adults are more often confused with children (and vice-versa) compared to with male adults. Comparing the two algorithms,  \gls{lena} achieves fewer false positives compared to \gls{vtc}, but also exhibits lower true positive rates, in line with previous observations that \gls{lena} has a generally higher precision but lower recall \citep{lavechin2025performance}.

Average confusion rates, however, are not sufficient metrics for comparing the merits of multiple classifiers. Another criterion is whether the performance of these algorithms is \textit{consistent} throughout recording conditions. Imagine a classifier with a very consistent detection rate of 50\%. The actual quantity of events could still be estimated precisely, by multiplying the amount of detected events by two. Now imagine a different classifier that has lower confusion rates but more variable performance, for instance  with a detection rate varying widely between 50 and 70\%; although it is more accurate on average than the first classifier (because the detection rate is higher),  its calibration would be much trickier because one cannot simply multiply detected events by two. To compare the consistency of \gls{lena} and \gls{vtc}, Figure \ref{fig:confusion_matrix_density} shows the distribution of confusion rates throughout recordings. The distributions of the rates of true positives (on the diagonal) are more peaked for \gls{vtc} than \gls{lena}, which indicates that the recall rates of the former is more consistent across recordings than the latter. The shading behind each of the curves in this Figure also gives us another important piece of information, as it represents the uncertainty about the underlying distribution of confusion rates. We note that this uncertainty is higher for under-represented speakers (other children, and especially male adults).

Having established how \gls{lena} and \gls{vtc} differ in their classification accuracy against manual coding, we now turn to question (2): how do these classification errors explain the correlation distortions observed in Figure \ref{fig:clips}, Section \S\ref{section:the_case_of}? Turning back to this Figure, we indeed observe that \gls{vtc} and \gls{lena} report higher correlations between CHI, OCH and FEM, MAL than humans. We also find that \gls{vtc} and \gls{lena} report higher correlations between CHI/OCH and FEM than between CHI/OCH and MAL, again in line with what we expect from classification errors. In addition, the confusion matrix shows that \gls{vtc} exhibits higher rates of false positives than \gls{lena} generally. \gls{vtc} reports higher correlations between speakers than \gls{lena} across the board, as expected from diarization errors; this is true even when correlations are computed at the levels of recordings and children (Figures \ref{fig:recordings} and \ref{fig:children}, Appendix \S\ref{appendix:correlations}). 
These findings are consistent with the distortions of correlations observed in the previous section. First of all, \gls{vtc} finds higher correlations across the board, which is consistent with its higher rates of false positives. In addition, regardless of the algorithm, the distortion of correlations is generally greater for pairs of speakers that are often confused with one another (e.g. CHI/OCH, OCH/FEM, or FEM/MAL). This is true for \gls{lena} as well, despite its lower rate of false positives. While this establishes the biasing effect of speaker confusion on correlations between speakers, the next section shows that this issue also affects measurements and inferences, implying quantities more obviously relevant to language acquisition.

\subsection{\label{section:effects}Effect of classification bias on downstream analyses and measurements}

\begin{figure}[!h]
    \centering
    \includegraphics[width=\linewidth]{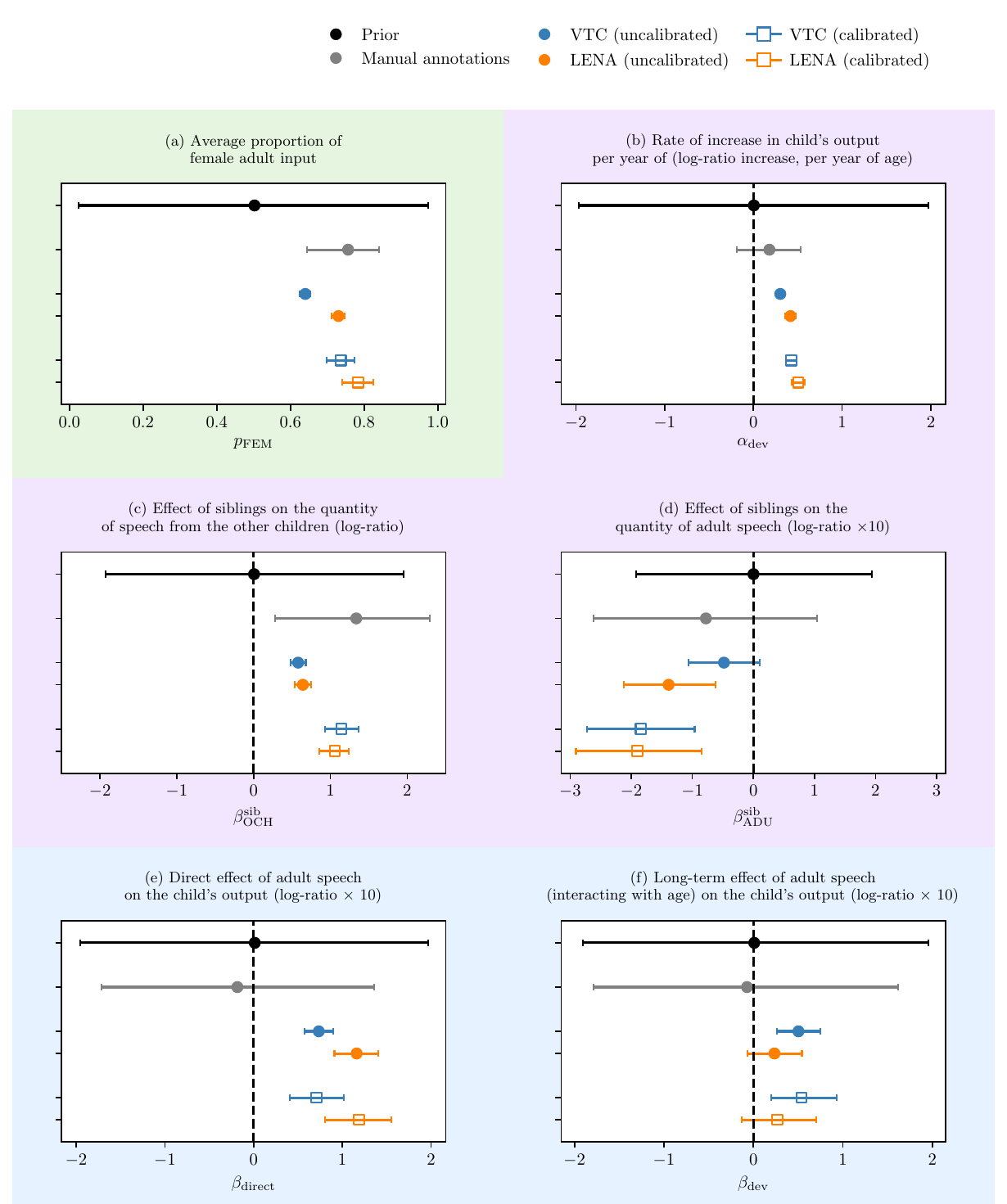}
    \caption{\label{fig:effects} Comparison of effects' sizes derived with manual annotations alone (in gray) and automated annotations (in colors), without any calibration and with calibration. The prior distribution ($\mathcal{N}(0,1)$ or $\mathcal{U}(0,1)$, depending on the variable support) is shown in black for purposes of comparison. We distinguish three types of measurements: direct measurements of speech quantities (a); measurements of the effect of an independent variable on speech quantity (b, c, d); and measurements of the effect of a quantity of speech on another quantity of speech (e, f). Numerical values are reported in Appendix \S\ref{section:regression}, Table \ref{table:regression_comparison}.}
\end{figure}

In what follows, we report the effect of classification bias on six measurements of variables and effects relevant to language acquisition, using the model detailed in Section \S\ref{section:speech_model}  (see Figure \ref{fig:effects}). These are: (a) the proportion of female adult input; (b) the effect of age on children's speech output; (c) the effect of siblings on the quantity of input from other children and (d) from adults; (e) the direct effect of adult speech on children's output; (f) the long-term effect of adult input on children's output. Each measure is estimated using manual annotations alone, automated annotations without any calibration, and automated annotations with Bayesian calibration. Results are grouped by type of measurement.

In every case, the prior distribution (in black) reveals the initial hypothesis with respect to each measurement (the Bayesian prior), before considering any kind of data. The next set of observations (in gray) pertains to manual annotations. Since manual annotations are available only for portions of a limited subset of recordings, they result in wide credible intervals largely reflecting the prior, indicating that little to nothing could be learned. Automated annotations are associated with much narrower credible intervals in all cases, demonstrating the benefits of machine learning classifiers in harnessing more data from which to estimate an effect. 
Finally, we show posteriors extracted from the joint model, which separates the contributions to vocalization quantities from the model of speech behavior from those due to confusion, captured by the model of algorithm behavior. The calibrated estimates are associated with larger credible intervals, reflecting our uncertainty in light of the stochastic nature of the algorithms' behavior.

\paragraph{What is the contribution of female adults to children's speech input?}  First, we consider the proportion of input attributed to female adult voices (\ref{fig:effects}a). The prevalence of female versus male adult speech is relevant to a range of disciplines, feeding theoretical discussions on variation in parental investment as a function of family organization (e.g., \cite{cassar2023makes}) as well as interventions geared at greater involvement of fathers (e.g., \cite{ferjan2022habla}). 
However, confusion between different speakers may distort our estimates of female input proportion (Figure \ref{fig:example_fem_prop}). For instance, vocalizations from children may be falsely attributed to adults; and vocalizations from male and female adults could be mistaken with each other, possibly at different rates. In Figure \ref{fig:effects}a, we find that \gls{lena} and \gls{vtc} yield incompatible measurements of the proportion of female adult speech, with \gls{vtc} probably underestimating it. After calibration, \gls{vtc} and \gls{lena} estimates are closer to each other.

\paragraph{Do children vocalize more with age?} Next, we consider the rate of increase in children's output with age (\ref{fig:effects}b). Age-related increases in typically developing children's vocal production in long-form recordings have been widely documented, using both \gls{lena} (e.g. \cite{bergelson2023everyday}) and \gls{vtc} (e.g., \cite{herve2024daily}). These changes could potentially be due to various reasons, including simple maturation – by which we mean processes that occur independently from exposure or learning. For instance, the emergence of canonical syllables in children’s production is independent of auditory experience because deaf and hearing infants start babbling at roughly the same ages; but canonical syllables are more frequent and diverse in hearing than deaf children, suggesting that a process in addition to simple maturation is necessary for the latter changes \citep{Oller1994}. Age-related increases in vocal production are thought to reflect not simply maturational changes, but actually improvements in children's language skills, based on correlations between vocalization quantity and standardized measures of language development (a meta-analysis in \cite{wang2020meta}) and the observation that children with atypical development show different age-related changes \citep{warlaumont2014social,hamrick2023semi}. Figure \ref{fig:effects}b shows that manual annotations are too sparse to confidently measure the increase in children's output with age. By contrast, automated annotations from \gls{lena} and \gls{vtc} produce highly confident but non-overlapping estimates. Calibration increases the developmental effect of age, and reduces the gap between the two algorithms. Most likely, when classification errors are unaccounted for, speech output is contaminated with input speech that is less sensitive to the child's age, thus damping out the estimated effect of age.

\paragraph{How do siblings affect children's language input?} Third, we consider the effect of siblings on the quantity of input children receive from other children (\ref{fig:effects}c) and adults (\ref{fig:effects}d). Among certain populations, speech from other children in the main exposure to language in infants (\citealt{Cristia2023}), and the contribution of siblings (as opposed to e.g. non-kins) may vary in rural and non-rural contexts. In our data, automated annotations without calibration show that children with no siblings receive about $\sim 40\%$ fewer vocalizations from other children than children with siblings. This difference is implausibly low in the corpora under consideration in which siblings are expected to be, by far, the primary source of exposure to speech from other children. By contrast, the calibrated estimate ($\sim 67\%$ less input from children for participants without siblings) is much more plausible. Without proper calibration, both algorithms under-estimate the effect size by a factor of two. Thus, without calibration, the contribution of non-kin children cannot be evaluated reliably, which is a problem for populations among which it is substantial. We then considered the effect of siblings on the input received from adults. Using one-hour long home-recorded videos, \citet{Laing2024} find that children with more than one sibling receive less input from their caregivers. This result is consistent with ``the resource dilution'' model, a hypothesis that was put forward to explain a pattern of lower educational attainment as a function of sibship size by arguing that, as the number of children increases in a household, the main holders of intellectual resources (the adults) have to split them among the children, resulting in fewer resources per child (e.g., \cite{kalmijn2016sibship}). We tried to replicate this finding using our comprehensive automatic annotations of child-centered longform recordings. First of all, the effect of siblings on adult input is imperceptible with manual annotations alone. Automated annotations provide strikingly different estimates; while \gls{lena} finds that siblings reduce the amount of input afforded by adults to the child, \gls{vtc} estimate is compatible with the null hypothesis (95\% credible level). After calibration, the estimate of this effect is much larger, pointing to a 20\% reduction in adult input among children with siblings, and consistent across \gls{vtc} and \gls{lena}. 

\paragraph{Does hearing more speech make children talk more, in that recording or in the long run?} The final examples  consider the measurement of the correlation between ``input'' (speech heard by children) and ``output'' (how much speech they produce) (\ref{fig:effects}e, \ref{fig:effects}f). These have been previously approached in many different ways (as discussed in the meta-analysis by \cite{anderson2021linking}), including using daylong recordings (as in the meta-analysis by \citealt{Coffey_2024}). For example, \citet{bergelson2023everyday} in the most diverse \gls{lena} study to date, reported a strong association between the quantity of adult speech afforded to children and the rate at which their speech production increases over time, a finding that is worth attempting to replicate. In that study, both measures were drawn from the same audio-recordings, analyzed with \gls{lena} software, with CVC being predicted by adult vocalization counts (AVC). Of course, an association between input and output cannot be easily interpreted as indicating a causal relationship. For instance, \citep{bergelson2023everyday} raise concerns a positive correlation may (partially) reflect shared genetics between the adults and children recorded\footnote{This explanation would only be partial, since input is required to acquire language, and that input can have effects independent of shared genetics between mother and child \citep{coffey2022effects}.}. As shown in Figure \ref{fig:example_associations}, speaker classification errors could constitute an additional cause of confound, which, to our knowledge, has not been discussed. In our case, we distinguish the recording-level effect of input on output (surfacing as higher amounts of child speech in recordings with more input from adults, \textit{ceteris paribus}), from the long-term developmental effect of input on child's speech, resulting from sustained exposure to higher input over time. No effects are found when using human annotations alone. Without calibration, \gls{vtc} and \gls{lena} find positive effects, although their estimates are non-overlapping. After calibration, the effects remain roughly unaltered (except for larger uncertainties), and \gls{vtc}/\gls{lena} continue to disagree.  Persisting discrepancies between \gls{lena} and \gls{vtc} indicate that the algorithms differ in ways unaccounted for by the calibration model. 

In our typology of measurement, calibration works best for effects of a known variable on a speech quantity: intuitively, independent variables can help the model discriminate between spurious and actual correlations. Calibration was less useful for direct correlations between quantities of speech. One potential reason is that such correlations are most directly affected by classification errors and thus harder to tell apart, especially if the size of the actual effect is comparable in magnitude to the confusion rates. 

\subsection{\label{section:female_proportion}Anticipating biases with simulations}

Classification bias can distort statistical measurements to varying extents. How to predict the level of sensitivity of a particular measurement to classification errors? This question can be answered with the simulation strategy proposed in Section \S\ref{section:simulations}. For instance, let us consider the relative contribution of female adults to adult input. We generate random datapoints representing the true amount of vocalizations from each speaker (CHI, OCH, FEM, and MAL) using the following generative process:

\begin{align*} 
\text{CHI} &= 1500\\ 
\text{OCH} &= \begin{cases}
      0 & \text{with probability } 1/2\\
      1000 & \text{with probability } 1/2
    \end{cases}\\
\text{FEM} &= 3000 \times p\\
\text{MAL} &= 3000 \times (1-p)\\
p &\sim \text{uniform}(0,1)
\end{align*}

\begin{figure}[H]
\begin{subfigure}[t]{0.48\textwidth}
    \hspace{-0.2cm}\includegraphics[width=1.1\linewidth]{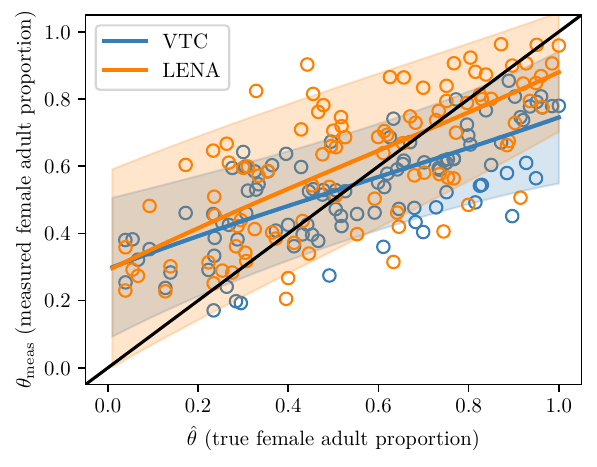}
    \caption{True values versus measured values of female adult input proportion, according to simulations of each classifier (LENA and VTC). Markers show a fraction of the samples. Ideally, all points should fall onto the black line ($\theta_{\text{meas}}=\hat{\theta}$). Colored lines show the average trends. They depart from $\theta_{\text{meas}}=\hat{\theta}$, which implies the presence of bias. The shaded area represents the 90\% probable interval.}
    \label{fig:simulation}
\end{subfigure}\hspace{0.5cm}%
\begin{subfigure}[t]{0.48\textwidth}  
    \centering 
    \includegraphics[width=1.1\textwidth]{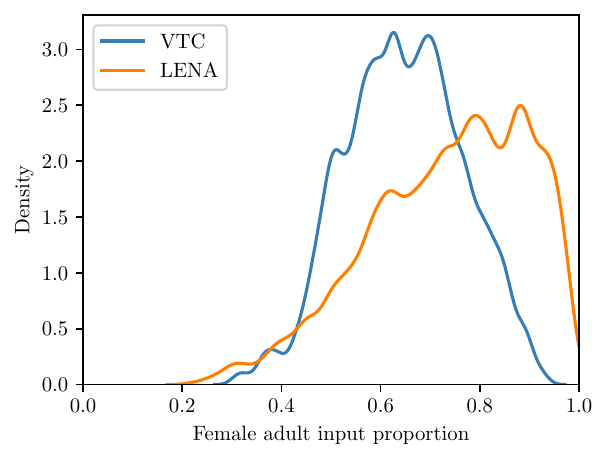}
    \caption{Distribution of the proportion of female adult input across recordings in the data, as estimated with VTC and LENA. LENA measures higher values than VTC, as predicted with simulations. The distributions, derived from the same recordings, are strikingly different.}
    \label{fig:female_proportion}
\end{subfigure}
\caption{Impact of speaker confusion on measurements of female and male adult speech.}
\end{figure}

The total quantity of adult input (FEM+MAL) is fixed. The proportion of female adult input is represented by a parameter $\hat{\theta}=p$ drawn uniformly between $0$ and $1$. We can then simulate the algorithm's output for each generated sample, and compare the measured fraction of female adult speech $\theta_{\text{meas}}$ to the true value $\hat{\theta}=p$. The results are shown in Figure \ref{fig:simulation} (using 2000 samples). They demonstrate that estimates of the proportion of female adult speech are biased, especially for extreme proportions (close to $0$ or $1$). Moreover, \gls{lena} estimates higher fractions of female adult speech than \gls{vtc}, and tends to overestimate female adult speech up to $p\sim 0.75$. This is probably because it was calibrated on training data dominated by female adult speech, together with the fact that the algorithm relies a lot on speakers' average prevalence to inform its classification boundary. Simulations also reveal the variance in measurements (as indicated by the shaded areas in Figure \ref{fig:simulation}). \gls{lena} has generally higher dispersion. The reliability of this simulation approach can be assessed by comparing these predictions to actual data. Figure \ref{fig:female_proportion} shows the distribution of the proportion of female adult input estimated by \gls{vtc} and \gls{lena} across the same corpus of audio recordings. The distribution obtained with \gls{lena} is shifted to the right (i.e., \gls{lena} reports higher proportions) and it has larger variance than the distribution given by \gls{vtc}, as predicted by the simulations. Moreover, \gls{vtc} saturates around $p=0.9$, again in accordance with the simulations.

\section{Discussion}

\subsection{Empirical findings}

We investigated the statistical biases resulting from errors in automated classification. Although we validated our approach on longform recordings in the context of language acquisition research, we note that it is conceptually applicable to any classification algorithm which detects events (in speech or other domains). 
We found evidence of spurious associations between speakers resulting from ``biasing paths'' opened up by speaker misidentification. Differences in correlation estimates across \gls{lena} and \gls{vtc} appear consistent with what is expected given differences in these algorithms' confusion rates (e.g., \gls{vtc} has higher confusion rates, and reports higher correlations between speakers). Using simulations, we found even more evidence that classification errors can account for certain discrepancies between \gls{lena} and \gls{vtc}. For instance, simulations informed by calibration data correctly predict that \gls{vtc} underestimates the proportion of speech from female adults, more than \gls{lena}, and produces less dispersed estimates across recordings.

We also studied how our Bayesian calibration approach could be used to alleviate biases. We found that our approach produces wider credible intervals, better reflecting our true uncertainty about the final measurements. Additionally, this approach reduces the disagreement in inferences across the two algorithms in almost all research questions we looked at. For instance, after calibration, \gls{lena} and \gls{vtc} find a consistent negative effect of siblings on adult input, 20–80\% stronger than measured prior to calibration. However, in the case of recording-level association between adult input and child output, Bayesian calibration failed to reduce the disagreement between \gls{lena} and \gls{vtc}. In this particular case, our approach fails to resolve the gap between \gls{vtc} and \gls{lena}. This suggests that there remain differences in these algorithms unaccounted for by our model of algorithm behavior, and invites further work to complexify it. 
This exception not withstanding, we provide ample evidence that our proposed approach can improve inferences. 

Below, we unpack the implications of our work (\S\ref{section:implications}), for statistical inference with machine learning classifiers in behavioral science (\S\ref{section:implications_stats}) and for child-development/language acquisition research in particular (\S\ref{section:implications_childev}). Finally, we review present limitations of Bayesian calibration and opportunities for future works (\S\ref{section:limitations}).

\subsection{\label{section:implications}Implications and recommendations}

\subsubsection{\label{section:implications_stats}Implications for statistical inference with machine learning classifiers}

Although our main goal in this paper is to lay out some ways in which classifier errors may affect scientific conclusions rather than provide a fool-proof, out-of-the-box solution, we feel that it would be inappropriate to end the paper without attempting to produce a set of actionable recommendations. 

First of all, standard measures in psychology such as accuracy, reliability, and validity are insufficient for addressing biases resulting from classification errors: two measures (e.g.: voc counts CHI, voc counts FEM) can be individually reasonably accurate (i.e., each measure is strongly correlated with human judgment), reliable (i.e., each measure is stable across repeated measurements), and valid (i.e., each measure is strongly correlated with independent relevant metrics), and nevertheless spuriously correlate to each other. Instead, we recommend the recourse to simulations for assessing the impact of algorithmic errors on a measurement of interest, as we have illustrated with the proportion of female adult speech. Simulations are rather computationally inexpensive and technically easy to implement.  They may be used in null-hypothesis testing contexts, for assessing whether a finding is compatible with the absence of an effect, once classification errors have been taken into account. For the specific case of \gls{lena} and \gls{vtc}, we facilitate the adoption of this approach by introducing a Python package simulating their outputs given synthetic ground-truth data \citep{diarization-simulation}. In addition, experienced users may consider Bayesian calibration, which can directly recover unbiased estimates. However, this strategy remains experimental, and comes at the expense of higher technical and computational costs.

As a robustness check, one may execute an analysis pipeline using annotations from different classifiers. This can provide some indication about the sensitivity of the analysis to classification errors. For instance, \gls{lena} and \gls{vtc} often yield non-overlapping estimates, given the differences in how they balance recall and precision (see Figure \ref{fig:confusion_matrix}). This strategy, however, is not nearly as reliable as the appeal to simulations, because multiple classifiers can misbehave in similar ways, achieving mutually compatible but nevertheless biased estimates (see, e.g., Figure \ref{fig:effects}c). 

We would like to conclude with a recommendation to producers of machine learning algorithms. As we have seen, \gls{lena} and \gls{vtc} behave differently in part due to a different approach towards the precision/recall trade-off. The optimal balance of recall and precision might depend on the task at hand, and the relative harm of false positive versus false negatives \citep{SilvaFilho2023}. We therefore suggest that computer scientists allow the users of their algorithms to re-adjust the recall/precision trade-off according to their need, based on the confidence scores of the predictions\footnote{For instance, \gls{vtc} computes an evidence score for each class (each speaker) at each audio frame, which is in a number between 0 and 1. In a second step, \gls{vtc} decides if a class is active or not depending on whether the evidence score exceeds a certain threshold (which, in the case of \gls{vtc}, is chosen to maximize the F-score, but could also be optimized differently)}.

\subsubsection{\label{section:implications_childev}Implications for child development and language acquisition research} 

\paragraph{Re-assessing \gls{lena} and \gls{vtc}}

Besides the observation that statistical measures of interest for child development are distorted by classification errors, this work offers a new opportunity to assess the relative merits of \gls{lena} and \gls{vtc}. On average, due to a stronger emphasis on precision with respect to recall, \gls{lena} may seem less subject to classification bias. However, we observed that \gls{lena}'s behavior is more variable, which increases measurement noise and renders calibration more challenging; and its inability to support overlapping speech is harder to model and account for rigorously. That said, \gls{lena}'s algorithm has been stable for over ten years, in principle allowing greater comparability across studies. In contrast, \gls{vtc} is an open-source algorithm, which benefits from improvements incorporating state-of-the-art solutions in speech processing \citep{kunze2024challenges}. This is both a strength and a limitation. Recent developments (VTC 2.0, \citealt{charlot2025babyhubert}) radically improved precision and recall, drawing close to human-human agreement for the MAL and FEM categories and closing the human-algorithm performance gap for OCH. While this improves accuracy, it entails that users will need to revisit the present calibration work, as our above-report may not generalize to the new \gls{vtc}. We hasten to indicate that two of the authors of the present paper are involved in both the original \gls{vtc} and ongoing work, which may constitute a conflict of interest biasing our perception. We thus strongly encourage readers to peruse our results carefully to make up their own minds.


\paragraph{Data collection}

Considering the amount of manual annotations required to implement the simulation or the Bayesian calibration approach, researchers may find it necessary to momentarily rely on our own estimates of the confusion matrices of \gls{lena} and \gls{vtc} (as provided by our package \citealt{diarization-simulation}), even though these estimates were derived from non-representative corpora (urban, English-speaking populations). Nevertheless, several reasons may warrant further manual annotation efforts.

First of all, there are a number of outstanding questions that will require either coordinated annotation efforts (as in \citet{ACLEW-DAS}) or sharing facilitated by standardized solutions \citep{Gautheron2022}. We observe a great deal of variation in confusion rates, but at present lack sufficient data to find the sources of this variation. Indeed, preliminary experimentation suggested our models could not converge with less human annotation than the 28h we included. We believe it would be desirable to multiply the quantity of human annotation by a factor of 4, so as to be able to subset it at least into 25\%, 50\%, and 75\% and thus assess how data quantity affects uncertainty. Since it took 4 teams 3 years to annotate the 20h in bergelson, lucid, warlaumont, and winnipeg \citep{Soderstrom2021}, it may take several years to quadruple that data amount. 

When possible, we think it may be strategic to sample data in ways that inform both sections of the joint model. Regarding the speech behavior section of the joint model, and as explained above, other researchers may propose alternative (simpler or more complex) models of speech behavior, which would have effects on the data required. Nonetheless, assuming our model of speech behavior, the largest gap is the absence of human annotation of samples extracted from multiple recordings within each child; that is, the ACLEW project drew multiple samples from a given child, but all of them came from the same recording day, which misses both random cross-recording variation and potentially systematic longitudinal variation.

Regarding strategic sampling to inform the algorithm behavior section of the joint model, most of the human annotation data we used was based on random sampling, because we wanted to avoid potential biases emerging from sampling mainly from sections where conversation was more prevalent. However, future iterations could benefit from a better understanding of how different auditory contexts affect both the algorithms and the human annotators. There is already some work looking at the agreement across humans and algorithms as a function of diverse factors ranging from background noise to the key child’s age. For instance, \citealt{peurey2025fifteen} discusses several factors that could explain variation in performance in voice type classifier algorithms, while also providing evidence that only some of them act as predicted. We extended these analyses in Appendix \ref{section:confusion_covariates}, which shows little compelling evidence that confusion rates vary as a function of child age or corpora. Nonetheless, this work still does not distinguish increased errors in the algorithm versus increased errors in human annotation. To this end, annotating data so that there is more overlap between annotators would be necessary, as well as extensions to the algorithm behavior model, to capture this variance.

\subsection{\label{section:limitations}Limitations of Bayesian calibration and opportunities for future works}

Bayesian calibration leverages both human and automated annotations to produce unbiased estimates with credible intervals reflecting our true uncertainty given the stochasticity of the underlying machine learning algorithms. Consistent with prior work, we indeed find that more robust results are obtained through the \textit{combination} of human and automated annotations \citep{Angelopoulos2023,teblunthuis2024misclassification}. In contrast to prediction-powered inference \citep{Angelopoulos2023}, Bayesian calibration is effective even with very few human annotations (equivalent to 0.2\% of the audio in our case). It is also much more flexible, since it decouples the model of the algorithm from the assumed behavioral model, allowing both to be refined in parallel. However, in contrast to \citet{Angelopoulos2023}, Bayesian calibration requires assumptions about the behavior of the algorithm itself. Since these assumptions can be too simplistic or incorrect, this provides less guarantee in general. 

Interestingly, Bayesian calibration has the potential to integrate predictions from different classifiers into a single analysis, whether these algorithms cover disjoint or overlapping portions of the data. While this requires minimal work in principle (see Appendix \S\ref{appendix:multiple-algos} for an example), we found that our approach was too slow to integrate both \gls{vtc} and \gls{lena} annotations in our large corpus. However, we believe further optimization is possible. Alternatively, approximate inference methods such as Variational Inference may provide further speedups, at the expense of accuracy.

Additionally, Bayesian calibration may achieve efficiency gains by incorporating additional predictors of the confusion rates. For example, confidence scores\footnote{Whether or not these have been calibrated themselves, in the sense of \citealt{guo2017calibration}.}, when provided by machine learning classifiers, could also be used as covariates in our calibration model. Similarly, automatically inferred estimates of audio quality (e.g. signal-to-noise ratios or reverberation levels) are also correlated with accuracy and could act as complementary covariates \citep{lavechin2023brouhaha,kunze2024challenges}.  

Finally, some researchers study properties of conversations \citep{Abney2016}, such as the probability of adults reinforcing children's speech-like vocalizations \citep{Warlaumont2014}, or the timing of inter-speaker turns \citep{Ritwika2025}. Extending our models to address the temporal nature of speech will be challenging. In our Bayesian framework, this would require estimating the probability that a true sequence of vocalizations (e.g., adult, child, adult, child, \dots) is detected as any other sequence (e.g., adult, adult, child, \dots). The space of possible sequences is combinatorially large, limiting this approach to short sequences. Time-coding precision, which we neglected, may also becomes critical. Finally, algorithms distort sequences in complex ways—through constraints; \gls{lena} imposes a minimum duration on vocalizations, and is unable to handle overlaps, while \gls{vtc} overproduces vocalizations with durations that are multiples of 250ms. Modelling such distortions is challenging but important for certain applications.

\section{Conclusions}

With this paper, we aimed to bring attention to the potential downstream consequences of classification and segmentation errors made by machine learning algorithms. As research attempts to capture behavior through denser and more ecological datasets, machine learning will become increasingly prominent. We thus do not recommend abandoning it, but rather increasing our understanding of where and how this tool's imperfection can affect our scientific conclusions, even if our data modeling needs to become more complex in order to account for those imperfections. The proposed Bayesian calibration approach is a first step in this direction.

\paragraph{Acknowledgments}

We are grateful to Richard McElreath for in-depth discussions and advice about our modeling strategy early-on in the project. We would like to thank Mark Vandam for providing additional data and information about the cougar corpus. We must also acknowledge Meg Cychosz, who suggested the idea examining the proportion of female adult speech. Finally, we are thankful to Riccardo Fusaroli, Camila Scaff, and Tarek Kunze for their feedback on this manuscript, and to DARCLE for the opportunity to present an early report of our findings.

\paragraph{Funding}

This work was granted access to the HPC resources of IDRIS under the allocation 2024-AD011012186 made by GENCI. This research was also undertaken with the assistance of resources from the National Computational Infrastructure (NCI Australia), an NCRIS enabled capability supported by the Australian Government. L.G. acknowledges funding from the DFG Research Training Group 2696. E.K. acknowledges funding from the Australian Research Council (CE140100041). M.L. acknowledges funding from The Simons Foundation International (034070-00033). A.C. acknowledges the J. S. McDonnell Foundation Understanding Human Cognition Scholar Award and European Research Council (ERC) under the European Union’s Horizon 2020 research and innovation programme (ExELang, Grant agreement No. 101001095). This work would not have been possible without the ACLEW project, for which we acknowledge Agence Nationale de la Recherche (ANR-16-DATA-0004 ACLEW).

\paragraph{Conflicts of interests}

The authors declare no competing interests.

\paragraph{Ethics approval}

Not applicable.

\paragraph{Consent to participate}

Not applicable.

\paragraph{Consent for publication}

Not applicable.

\paragraph{Availability of data and material}

The datasets analysed during the current study are not publicly available. However, the code includes synthetic data and routines to generate synthetic datasets, which allows readers to check the correctness of their implementation. In addition, the code can be run on any dataset formatted according to the ChildProject guidelines \citep{Gautheron2022}.

\paragraph{Code availability}

The code of our models and analyses is available at: \url{https://gin.g-node.org/LAAC-LSCP/speaker-confusion-model}. The Python simulation package is available at \url{https://github.com/LAAC-LSCP/diarization-simulation}.

\paragraph{Authors' contributions}

L.G.: Conceptualization, Data curation, Formal analysis, Methodology, Software, Validation, Visualization, Writing – original draft; E.K.: Resources, Writing – review \& editing; A.M.: Data curation, Writing – review \& editing; M.L.: Writing – review \& editing; A.C.: Writing – original draft, Supervision, Funding acquisition, Project administration.

\printbibliography

\appendix

\section{Supplementary materials}

\subsection{Models of speech behavior}

\subsubsection{\label{section:behavior_model_specification}Main model}

For each recording $k$ of child $c$, the vocalization counts are modeled as:
\begin{align}
v_{k,\text{CHI}}^{\text{recs}} &\sim \text{Gamma}\left(\alpha_{\text{child}}^{\text{CHI}}, \frac{\alpha_{\text{child}}^{\text{CHI}}}{\mu_{k,\text{CHI}}^{\text{rec}}}\right)\\
v_{k,s}^{\text{recs}} &\sim \text{Gamma}\left(\alpha_{\text{child}}^{s}, \frac{\alpha_{\text{child}}^{s}}{\mu_{c,s}^{\text{child}}}\right), \quad s \in \{\text{FEM}, \text{MAL}, \text{OCH}\}
\end{align}

The child's expected vocalization rate incorporates age effects and the influence of adult speech:
\begin{align}
\mu_{k,\text{CHI}}^{\text{rec}} &= \mu^{\text{pop}}_{\text{CHI}} \exp\left( 
    \alpha_c^{\text{dev}} \cdot \frac{\text{age}_k}{12} + 
    \beta^{\text{dev}} \cdot \frac{\text{age}_k}{12} \cdot \frac{\mu_{c,\text{ADU}}^{\text{child}} - \mu_{\text{ADU}}}{\sigma_{\text{ADU}}} \right.\\
    &\qquad\qquad\qquad\left.+ \beta^{\text{direct}} \cdot \frac{v_{k,\text{ADU}}^{\text{recs}} - \mu_{c,\text{ADU}}^{\text{child}}}{\sigma_{c,\text{ADU}}^{\text{child}}} \nonumber
\right)
\end{align}

At the child level, for children with known sibling status:
\begin{align}
\mu_{c,\text{OCH}}^{\text{child}} &\sim \text{Gamma}\left(\alpha_{\text{pop},d}^{\text{OCH}}, \frac{\alpha_{\text{pop},S_c}^{\text{OCH}}}{\mu^{\text{pop}}_{\text{OCH}} \exp(S_c \beta^{\text{OCH}})}\right)\\
\mu_{c,s}^{\text{child}} &\sim \text{Gamma}\left(\alpha_{\text{pop},d}^{s}, \frac{\alpha_{\text{pop},S_c}^{s}}{\mu^{\text{pop}}_{s} \exp(S_c \beta^{\text{ADU}}/10)}\right), \quad s \in \{\text{FEM}, \text{MAL}\}
\end{align}
where $S_c = 1$ if child has siblings, $S_c = 0$ otherwise.

For children with unknown sibling status, the model uses a mixture, marginalizing over the cases $S_c\in\{0,1\}$, with probability $p^{\text{sibs}}$ and $1-p^{\text{sibs}}$ respectively.


The population-level parameters have the following priors:
\begin{align}
\mu^{\text{pop}}_s &\sim \text{Gamma}(2, 8) \quad \text{(prior mean = 250 vocs/hour)}\\
\alpha_{\text{pop},d}^{s} &\sim \text{Gamma}(8, 1)\\
\alpha_{\text{child}}^{s} &\sim \text{Gamma}(4, 1)
\end{align}

Developmental effects are modeled with:
\begin{align}
\alpha_c^{\text{dev}} &\sim \text{Normal}(\alpha^{\text{dev}}, \sigma^{\text{dev}})\\
\alpha^{\text{dev}} &\sim \text{Normal}(0, 1)\\
\sigma^{\text{dev}} &\sim \text{Exponential}(1)\\
\beta^{\text{dev}} &\sim \text{Normal}(0, 1)\\
\beta^{\text{direct}} &\sim \text{Normal}(0, 1)
\end{align}

Sibling effects are captured by:
\begin{align}
S_c &\sim \text{Bernoulli}(p^{\text{sibs}})\\
p^{\text{sibs}} &\sim \text{Uniform}(0, 1)\\
\beta^{\text{OCH}} &\sim \text{Normal}(0, 1)\\
\beta^{\text{ADU}} &\sim \text{Normal}(0, 1)
\end{align}

Notation:
\begin{itemize}
\item $v_{k,s}^{\text{recs}}$: vocalization count for speaker $s$ in recording $k$
\item $\mu_{c,s}^{\text{child}}$: expected vocalization rate for speaker $s$ for child $c$  
\item $\alpha_{\text{child}}^{s}$: variance parameter for speaker $s$ at child level
\item $\alpha_{\text{pop},d}^{s}$: variance parameter for speaker $s$ at population level
\item $\mu^{\text{pop}}_s$: population-level average for speaker $s$
\item $\alpha_c^{\text{dev}}$: child-specific age effect
\item $\beta^{\text{dev}}$, $\beta^{\text{direct}}$: developmental coefficients
\item $S_c$: indicator for whether child $c$ has siblings
\end{itemize}

\subsubsection{\label{appendix:model_assumptions}Justification for the assumptions}

Our multi-hierarchical model implements several assumptions via equations that relate different variables to one another. Although the speech behavior model's shape is not per se a contribution of the present paper, we informed it on previous research as described in Table \ref{tab:assumptions}.

\begin{table}[H]
\centering
\renewcommand{\arraystretch}{1.4}
\begin{tabularx}{\textwidth}{>{\raggedright\arraybackslash}p{3cm} >{\raggedright\arraybackslash}X >{\centering\arraybackslash}p{2.5cm}}
\toprule
\textbf{Assumption} & \textbf{How the assumption is justified} & \textbf{Relevant equations} \\
\midrule
Individual variation in voc count &
Much previous research suggests there is individual variation in how voluble families are (e.g.\ Bergelson et al., 2023) &
9 and 10 \\
\addlinespace
Sibs $\rightarrow$ OCH &
Having siblings may affect the number of ``other child'' vocalizations found because there are more children around. Note that the distribution (eq.\ 21) includes the null as a possible outcome. &
9 \\
\addlinespace
Sibs $\rightarrow$ ADU &
Having siblings may affect how much adults speak around the key child (Laing \& Bergelson). Note that the distribution (eq.\ 22) includes the null as a possible outcome. &
10 \\
\addlinespace
No individual variation in how much newborns vocalize &
Although individuals may vary already at birth in their language skills, according to data like that in Bergelson et al.\ (2023) vocalization rates are incredibly low even at around 3 months relative to vocalization rates later in development, such that variation at birth is likely negligible, so we can simplify our model by not including individual variation in children's vocalization rates at age zero. &
(Implicit in eq.\ 8, since the relevant terms are multiplied by age) \\
\addlinespace
Individual variation in how much children vocalize increases with age &
Bergelson et al.\ (2023) and much other work supposes there is relevant individual variation in children's vocalization rates, with increases in this divergence with age. &
8 \\
\addlinespace
Adults' voc $\rightarrow$ how much children vocalize &
Bergelson et al.\ (2023) and much other work supposes there are long-term effects of adults' vocalization quantities on children's vocalization rates. &
8 \\
\bottomrule
\end{tabularx}
\caption{Model assumptions, justifications, and relevant equations.}
\label{tab:assumptions}
\end{table}

Our model of speech behavior allows us to look at a number of downstream effects of confusion errors, while limiting complexity. Future work relying on more data could consider increasing the complexity of the speech model by adding assumptions like the following: (1) Children’s age may affect the quantity of vocalizations of others (e.g., perhaps people vocalize more around older children). (2) Number of siblings and the siblings’ ages could affect the quantity of vocalization of other children (and perhaps that of adults).

\subsubsection{Fitting the model on human annotations alone}

Recordings for which human annotations are available are only  partially annotated (typically 30 minutes of audio is annotated, out of many hours). We therefore make the assumption that the relationship between the manual vocalization counts ($\bm{n_k^{\text{human}}} = (n_{k,CHI}^{\text{human}}, n_{k,OCH}^{\text{human}}, n_{k,FEM}^{\text{human}}, n_{k,MAL}^{\text{human}})$) and the unobserved vocalization counts for the whole recording ($\bm{v_k}$) is:

\begin{align}
    \bm{n_k^{\text{human}}} &\sim \mathrm{Poisson}\left(\dfrac{\tau_{\text{annotated}}}{\tau_{\text{rec}}} \cdot \bm{v_k} \right)
\end{align}

Where $\tau_{\text{annotated}}$ is the duration of the audio that was hand-annotated. This makes the drastic (and false) assumption that the vocalization rate is constant throughout the recordings, leading to overconfident credible intervals. Thus, in reality, human annotations alone are even \textit{less} informative than we report in Figure \ref{fig:effects}. The benefit of complementing human annotations with automated annotations is thus even larger.

\subsubsection{Observation and parameters summary}

\begin{table}[H]
    \centering
    \input{tables/dof}
    \caption{Summary of observations and parameters entering the joint model.}
    \label{table:summary}
\end{table}

\subsubsection{Stan parameters}

\begin{DIFnomarkup}
\begin{tabular}{l|l|l|l|l|l|l|l|}
\cline{2-8}
                                                                                                       & \textbf{Chains} & \textbf{\begin{tabular}[c]{@{}l@{}}Warmup\\ iter.\end{tabular}} & \textbf{\begin{tabular}[c]{@{}l@{}}Sampling\\ iter.\end{tabular}} & \textbf{\begin{tabular}[c]{@{}l@{}}Accept.\\ delta\end{tabular}} & \textbf{\begin{tabular}[c]{@{}l@{}}Max\\ tree-\\ depth\end{tabular}} & \textbf{\begin{tabular}[c]{@{}l@{}}CPU\\ cores\end{tabular}} & \textbf{\begin{tabular}[c]{@{}l@{}}Runtime\\ (h)\end{tabular}} \\ \hline
\multicolumn{1}{|l|}{\textbf{Behavioral model}}                                                        & 4               & 2,000                                                           & 2,000                                                             & 0.95                                                             & 15                                                                   & 48                                                           & 0.5-1                                                          \\ \hline
\multicolumn{1}{|l|}{\textbf{\begin{tabular}[c]{@{}l@{}}Full model\\ (with calibration)\end{tabular}}} & 1               & 1,000                                                           & 1,000                                                             & 0.95                                                             & 15                                                                   & 48                                                           & 10-12                                                           \\ \hline
\end{tabular}
\end{DIFnomarkup}

\subsection{Validation}

\subsubsection{\label{section:validation}Confusion model}

\begin{figure}[H]
    \centering
    \includegraphics{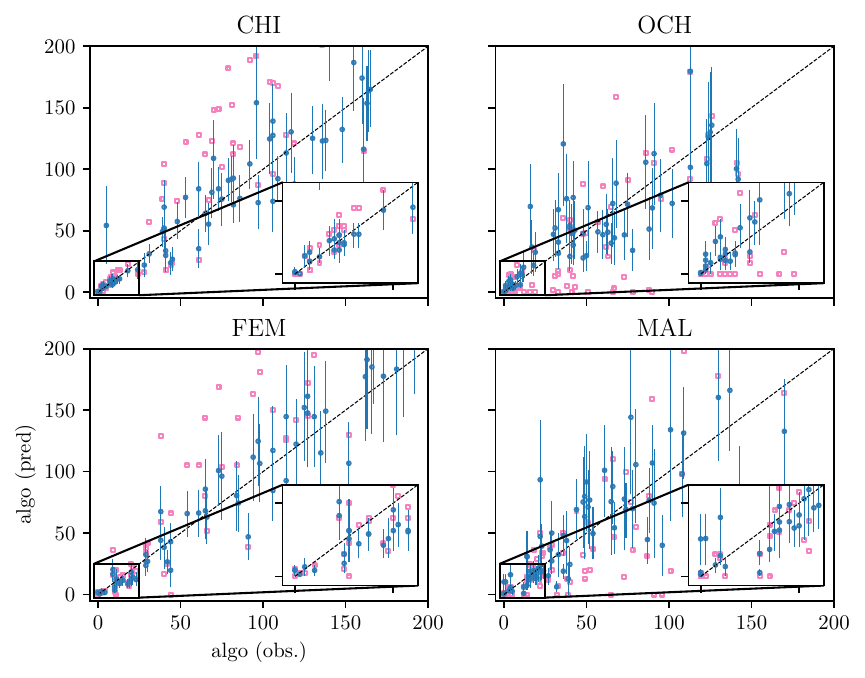}
    \caption{Relationship between the vocalization counts actually derived with \gls{vtc} and the quantities expected to be derived from \gls{vtc} given the model of the algorithm behavior and the true vocalization counts. Each blue point represents one of the recordings from the calibration data. The x-axis indicates the amount of vocalizations detected by \gls{vtc} for each speaker. The y-axis represents the amount expected based on the algorithm behavior (in blue) and the true amount of vocalizations for each speaker (in pink). The error bars indicate 68\% probable intervals; most of the uncertainty lies in the variance in the confusion rates across recordings.}
    \label{fig:validation-vtc}
\end{figure}

\begin{figure}[H]
    \centering
    \includegraphics{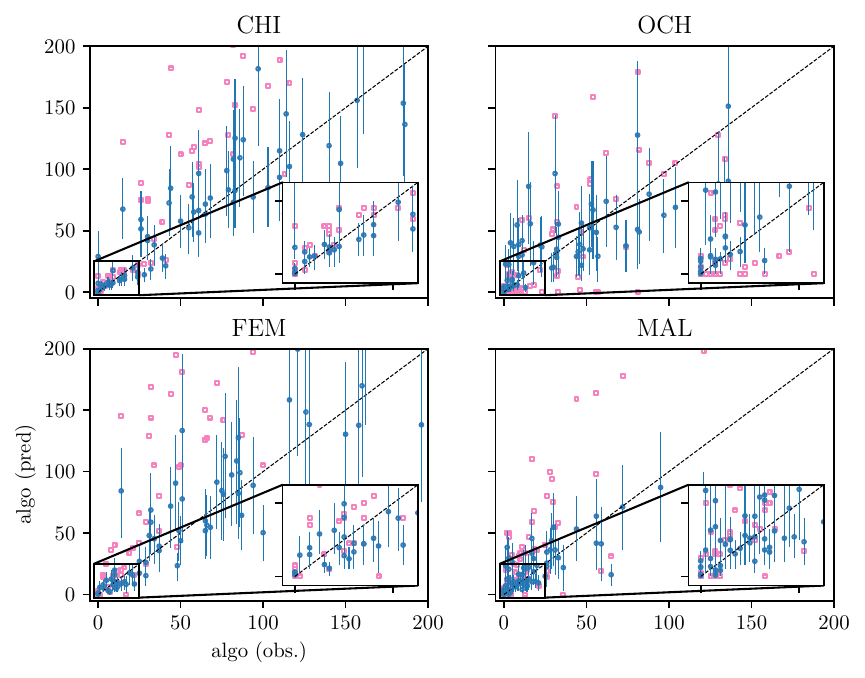}
    \caption{Relationship between the vocalization counts actually derived with \gls{lena} and the quantities expected to be derived from \gls{lena} given the model of the algorithm behavior and the true vocalization counts. Each blue point represents one of the recordings from the calibration data. The x-axis indicates the amount of vocalizations detected by \gls{lena} for each speaker. The y-axis represents the amount expected based on the algorithm behavior (in blue) and the true amount of vocalizations for each speaker (in pink). The error bars indicate 68\% probable intervals; most of the uncertainty lies in the variance in the confusion rates across recordings.}
    \label{fig:validation-lena}
\end{figure}

\subsubsection{\label{section:simulation-validation}Confusion model (validation via simulations)}

\begin{figure}[H]
    \centering
    \includegraphics[width=0.95\linewidth]{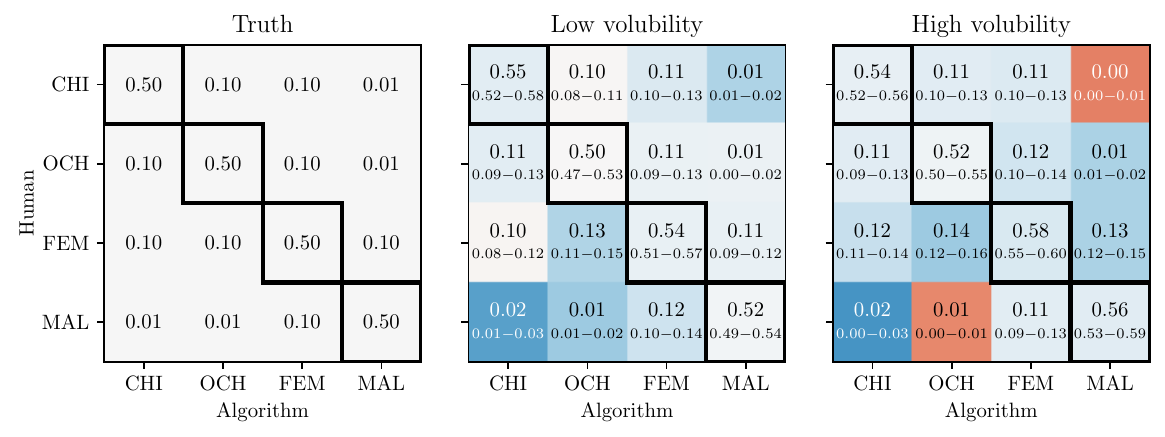}
    \caption{True confusion $(\mu_{ij})$ compared to the confusion matrices recovered by the model from simulated data, under normal and high volubility. Colors indicate deviations from the true values (blue indicates overestimates, and red indicates underestimates). High volubility can lead to overestimating $(\mu_{ij})$.}
    \label{fig:confusion_matrix_synthetic}
\end{figure}

To further validate our approach, we apply the model to simulated human and algorithmic annotations comparable in size to our calibration dataset. We draw vocalizations by assuming a Poisson process. Vocalization durations are drawn uniformly between 1s and 2s. Confusion rates across simulated recordings are drawn from a Beta distribution with mean $\mu_{ij}$ and shrinkage parameter $\eta=50$. Simulations show that our approach is able to identify the correct confusion rates as long as speech density remains not too high (Figure \ref{fig:confusion_matrix_synthetic}).

\subsubsection{\label{section:vtc_lena_comparison}Bayesian calibration}

\paragraph{VTC/LENA comparison before and after calibration}
\

\begin{figure}[H]
    \centering
    \includegraphics[width=0.75\linewidth]{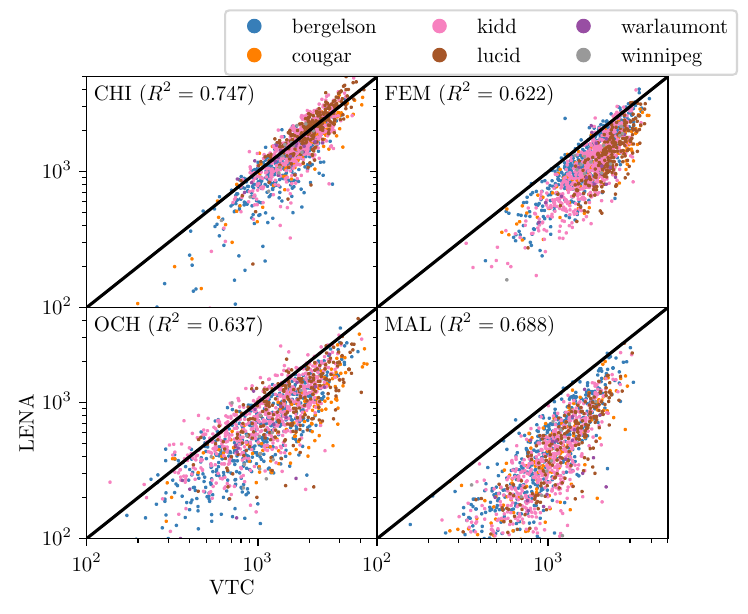}
    \caption{Comparison of vocalization counts derived with \gls{vtc} (x-axis) and \gls{lena} (y-axis) per speaker and per recording, prior to any calibration. }
    \label{fig:vocs_vtc_lena}
\end{figure}

\input{tables/correlation_vtc_lena}

\paragraph{Automated versus manual estimates}
\

\begin{figure}[H]
\centering
\includegraphics[width=0.8\textwidth]{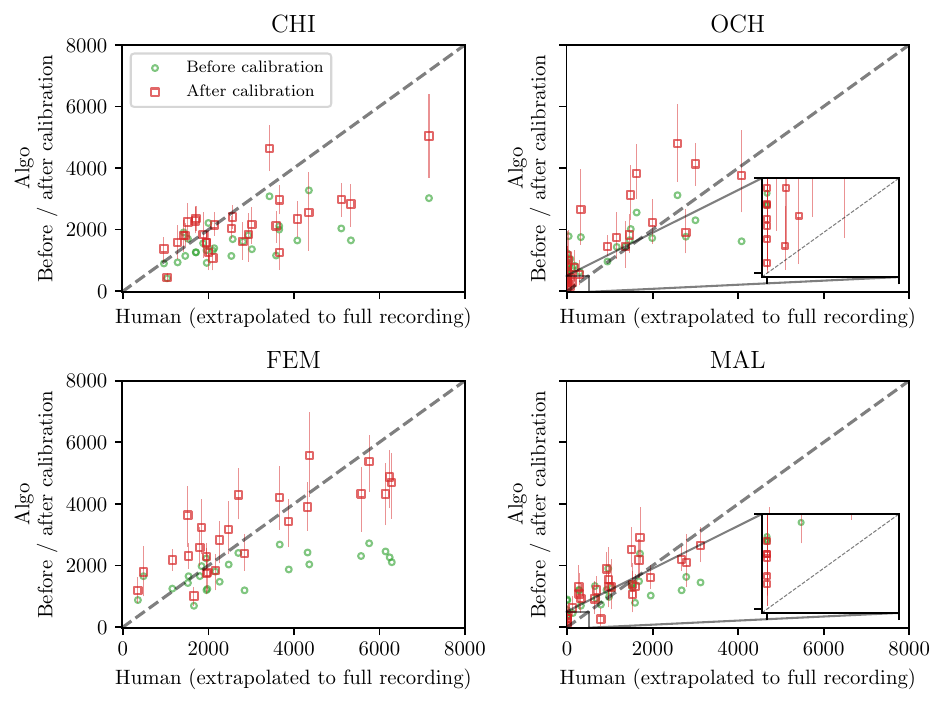}
\caption{\label{fig:truth_human_vtc}Comparison of manual estimates, uncalibrated VTC estimates, and calibrated VTC posterior estimates of total vocalization counts in recordings partially covered by human annotations. Prior to calibration, biases are manifest (CHI, FEM and MAL are almost systematically underestimated, and OCH is almost systematically overestimated).}

\end{figure}

\begin{figure}[H]
\centering
\includegraphics[width=0.8\textwidth]{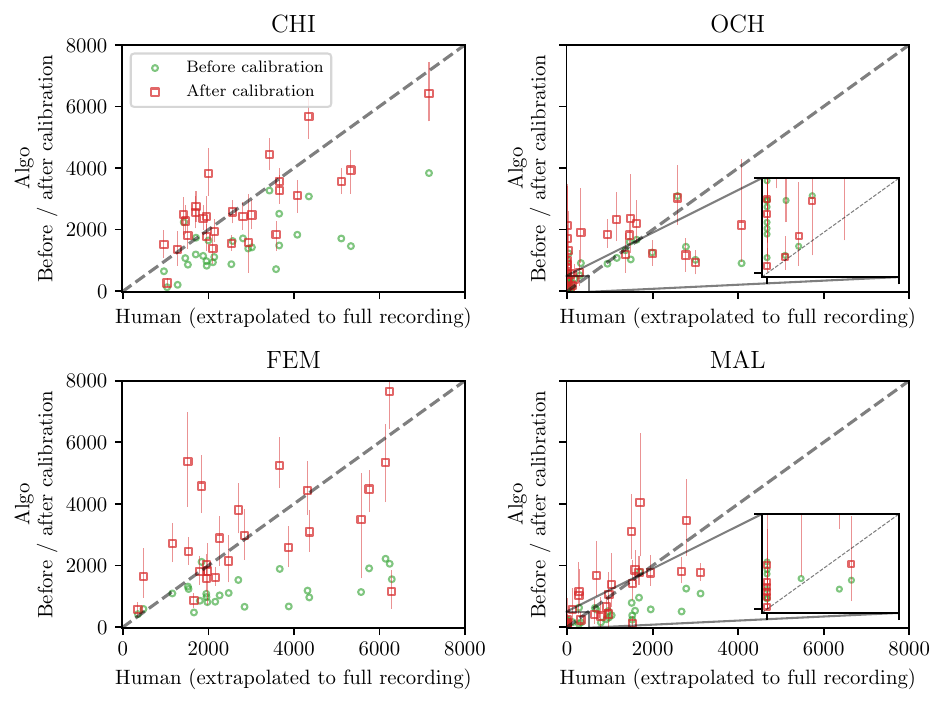}
\caption{\label{fig:truth_human_lena}Comparison of manual estimates, uncalibrated LENA estimates, and calibrated LENA posterior estimates of total vocalization counts in recordings partially covered by human annotations.}
\end{figure}

\begin{table}[H]
\centering
\begin{minipage}{0.48\textwidth}
\centering
\input{tables/calibration_vtc_data}
\caption*{VTC}
\end{minipage}%
\hfill
\begin{minipage}{0.48\textwidth}
\input{tables/calibration_lena_data}
\caption*{LENA}
\end{minipage}
\caption{\label{tab:truth_vs_data}Agreement between algorithmic vocalization counts (before/after calibration) and manual vocalization counts extrapolated to whole recordings. Brackets indicate 95\% confidence intervals. Best-values are shown in bold, when differences are statistically significant ($\ast$).}
\end{table}

\subsubsection{\label{appendix:calibration_validation_simulation}Bayesian calibration (validation via simulations)}

\begin{figure}[H]
    \centering
    \includegraphics[width=\textwidth]{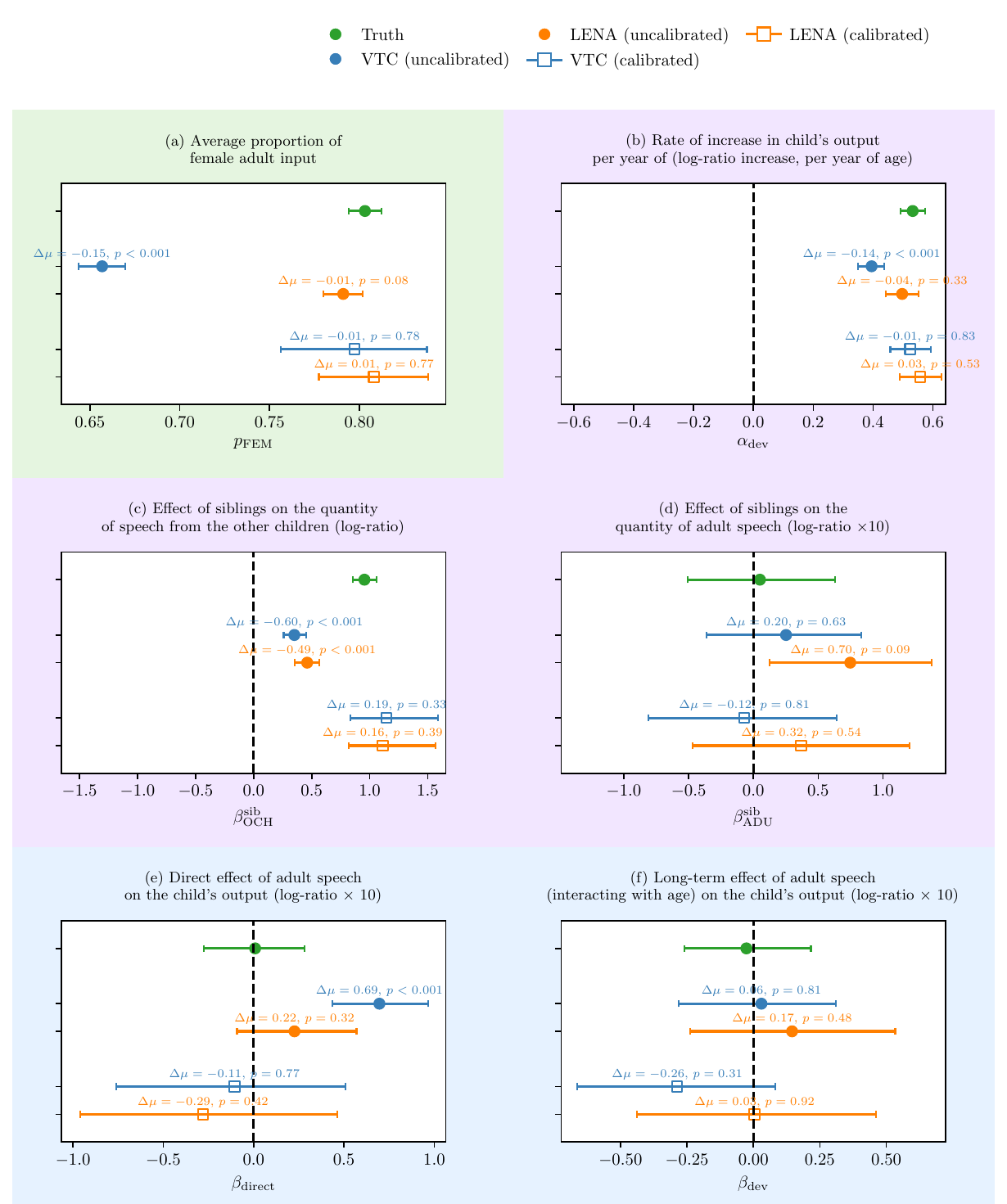}
    \caption{Performance of the calibration procedure on synthetic data. ``Truth'' values represent the posterior estimates obtained with the synthetic ground truth. $\Delta\mu$ is the difference between automated estimates and truth estimates. The p-values $p$ measure the significance of the difference using a two-tailed permutation test (higher is better). Unlike uncalibrated estimates, calibrated estimates never appear in significant tension with the truth estimates. In fact, calibration almost always reduces the tension (except for $\beta^{\text{dev}}$ with VTC).}
    \label{fig:synthetic}
\end{figure}

We provide an additional layer of validation of the calibration approach by running the procedure on synthetic vocalization data simulated under fixed values of the parameters of interest. We set plausible values for the effects that are expected to be large in real-life ($\alpha^{\text{dev}}=0.5$, $\sigma^{\text{dev}}=0.1$, $\beta^{\text{OCH}}=-1$) and zero-values for less trivial effects ($\beta^{\text{ADU}}=\beta^{\text{direct}}=\beta^{\text{dev}}=0$). We set the priors such that $\mu^\text{FEM}/(\mu^\text{FEM}+\mu^\text{MAL})=0.8$. We simulate vocalization counts for 1000 observations (200 children with 5 observations each). For each observation, we simulate automated vocalization counts using the procedure described in Section \S\ref{section:simulations}. We estimate the model parameters using three different inputs: a) the true synthetic vocalization counts; b) the simulated automated vocalization counts, without calibration; and c) the simulated automated vocalization counts, with calibration. The results are shown in Figure \ref{fig:synthetic}. This confirms that algorithmic estimates can be strongly biased, and that calibration generally reduces the tension between the ground truth and the posterior estimates.


\begin{table}[H]
\centering
\begin{minipage}{0.49\textwidth}
\centering
\input{tables/calibration_vtc_synthetic_table}
\caption*{VTC}
\end{minipage}%
\hfill
\begin{minipage}{0.49\textwidth}
\centering
\input{tables/calibration_lena_synthetic_table}

\caption*{LENA}
\end{minipage}
\caption{\label{table:truth_vs_synthetic}Accuracy of automated estimates, evaluated in terms of $R^2$ and relative error ($\frac{\mathrm{RMSE}}{\sigma_{\mathrm{truth}}}$), before calibration (using the simulated algorithm counts) and after calibration (using $\mathbb{E}(v_{ik})$, the posterior expectancy value of the true vocalization counts). Brackets indicate 95\% confidence intervals. Best values are shown in bold, when before/after differences are significant ($\ast$). Calibration significantly improves accuracy in simulated data.}
\end{table}

\begin{figure}[H]
\centering
\begin{subfigure}[b]{0.45\textwidth}
    \centering
    \includegraphics[width=\textwidth]{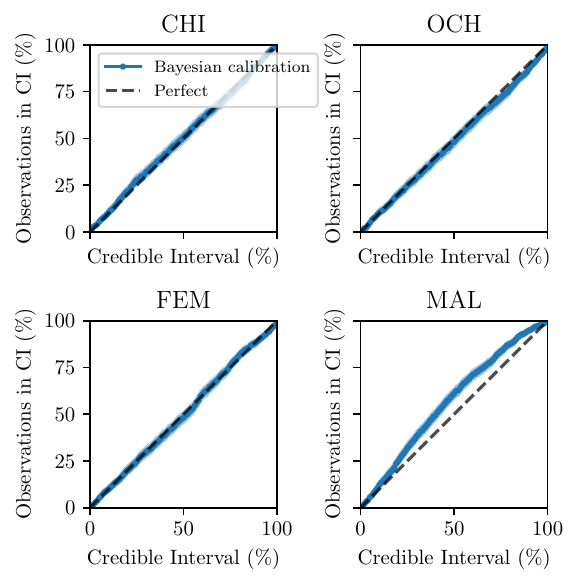}
    \caption{VTC}
    \label{fig:truth_human_vtc_calibration}
\end{subfigure}
\hfill
\begin{subfigure}[b]{0.45\textwidth}
    \centering
    \includegraphics[width=\textwidth]{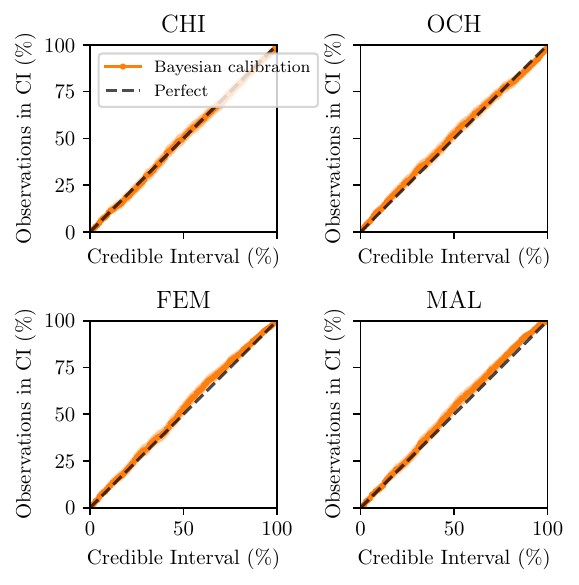}
    \caption{LENA}
    \label{fig:truth_human_lena_calibration}
\end{subfigure}
\caption{Calibration of posterior credible intervals for vocalization counts by speaker. Each point shows the proportion of observations falling within credible intervals of a given nominal level (x-axis). Perfect calibration (diagonal line) occurs when, e.g., 90\% credible intervals contain the true value 90\% of the time.}
\label{fig:truth_human_calibration}
\end{figure}

\subsubsection{\label{section:regression}Results}

{\onehalfspacing\input{tables/regression_table}}

\subsubsection{\label{section:alternative}Alternative approach (direct comparison)}

An undesirable feature of the approach based on vocalization counts in 15s clips is that reliance on an arbitrary window-size. Shorter windows introduce boundary effects; longer windows decrease the informativeness of each individual clip, challenging the ability of the model to learn the distribution of confusion rates.

\begin{figure}[H]
    \centering
        \includegraphics[width=0.67\linewidth]{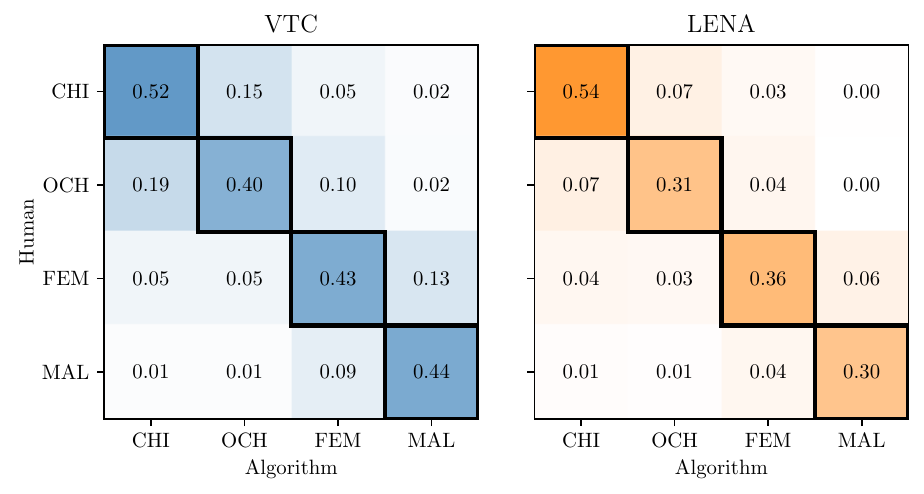}
        \caption{\label{fig:confusion_matrix_alternative}Average confusion rates of \gls{vtc} and \gls{lena} estimated within our model. Rows indicate the true speaker, and columns indicate the speaker class attributed by the algorithm. Diagonal elements represent the true positive rate for each speaker. Non-diagonal elements represent the distributions of the rates of false positives.}
\end{figure}

We considered an alternative strategy, based on an direct comparison of the vocalizations retrieved by the algorithms and the human annotators. In this strategy, an algorithmic vocalization attributed to a speaker $i$ is considered a true positive iif it has a non-zero intersection with a real vocalization from the same speaker. It is considered a false positive iif it has no intersection with actual vocalizations from the correct speaker ($i$), but intersects with vocalizations from a single other speaker $j\neq i$. For each recording $k$ with human annotations, we thus directly derive an estimate of $n_{kij}$, the amount of vocalizations attributed to $j$ as a result of vocalizations from $i$. This approach, however, necessarily underestimates confusion rates; in particular, it fails to capture vocalizations misidentified every time a detected vocalization intersects with actual vocalization from two speakers. Figure \ref{fig:confusion_matrix_alternative} confirms that this yields lower confusion rates than the 15s clips method.

\begin{figure}[H]
    \centering
    \includegraphics[width=0.95\linewidth]{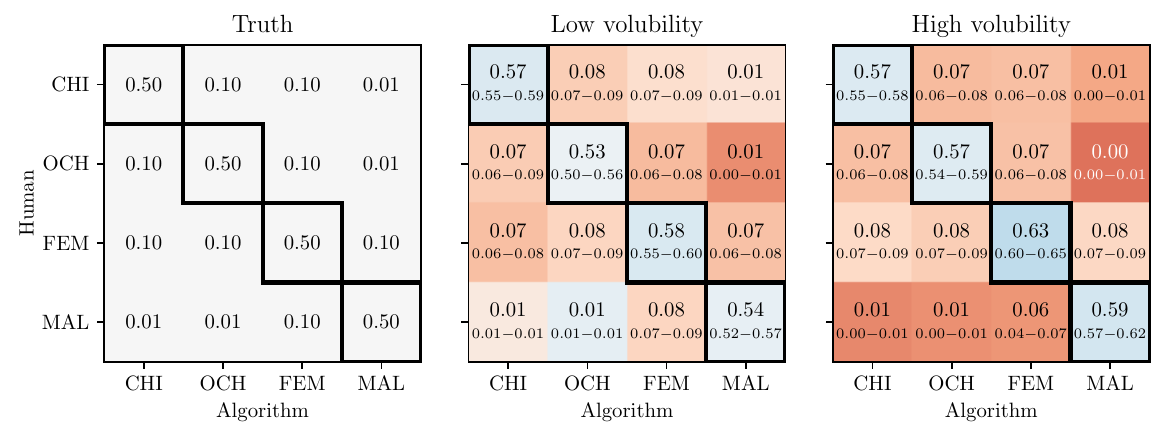}
    \caption{True confusion $(\mu_{ij})$ compared to the confusion matrices recovered by the alternative inference strategy from simulated data, under normal and high volubility. The matrix to the left represents the true values. Colors indicate deviations from the true values (blue indicates overestimates, and red indicates underestimates). The alternative strategy generally underestimates misclassification errors and overestimates true positives. }
    \label{fig:confusion_matrix_synthetic_comparison}
\end{figure}

\subsubsection{Downstream comparison of the two calibration strategies}

\begin{figure}[H]
    \centering
    \includegraphics[width=\linewidth]{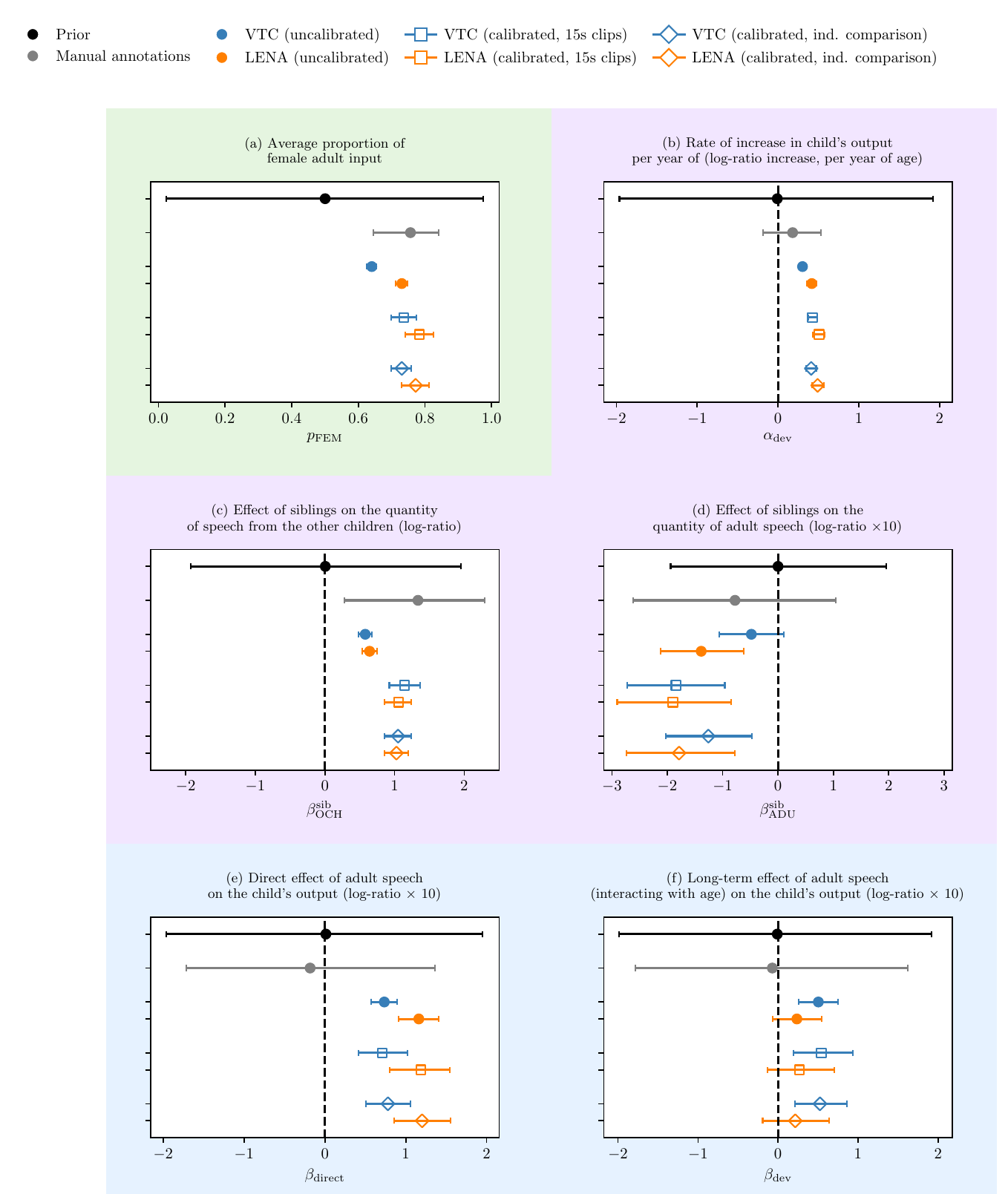}
    \caption{\label{fig:effects_full} Comparison of effects' sizes derived with manual annotations alone (in gray) and automated annotations (in colors), without any calibration and with calibration. The prior distribution ($\mathcal{N}(0,1)$ or $\mathcal{U}(0,1)$, depending on the variable support) is shown in black for purposes of comparison. We distinguish three types of measurements: direct measurements of speech quantities (a); measurements of the effect of an independent variable on speech quantity (b, c, d); and measurements of the effect of a quantity of speech on another quantity of speech (e, f). }
\end{figure}

\subsection{\label{section:confusion_covariates}Effect of the child's age and environment on confusion rates}

\begin{figure}[h]
    \centering
    \includegraphics[width=0.8\textwidth]{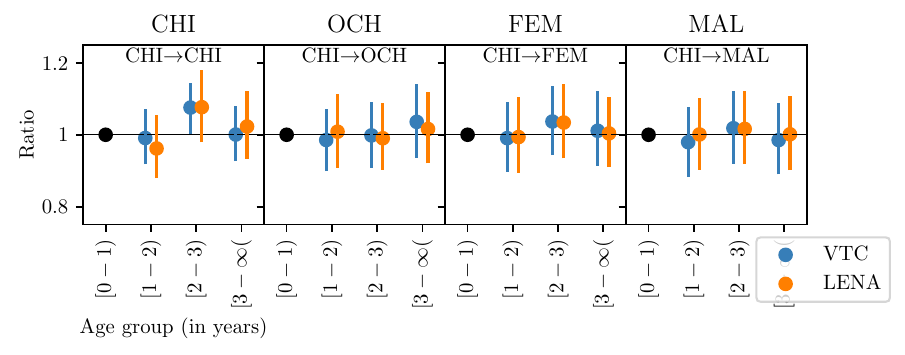}
    \caption{Effect of child age on the confusion rates $\lambda_{\text{CHI},i}$ (\text{CHI}$\to i \in\{\text{CHI,OCH,FEM,MAL}\}$). The mean confusion rate for each age bin is compared the mean confusion rates for children between zero and one year of age (a ratio of one implies no difference). For \gls{vtc}, variations larger than $\sim$10\% in the mean true positive rates are excluded.}
    \label{fig:confusion_age}
\end{figure}

Figures \ref{fig:confusion_matrix} and \ref{fig:confusion_matrix_density} collapse across all children and corpora. In reality, the confusion matrix might depend on a number of factors, in which case collapsing across them is inappropriate. One of them is the child's age, which could affect the ability of the algorithm to correctly detect and classify children's vocalization. If that were the case, this could potentially undermine the ability of inferring the effect of age on the child's speech production. The effect of age on the algorithm performance on the vocalizations of the key child is shown for different age groups in Figure \ref{fig:confusion_age}. There is some evidence that the detection rate for children increases after two years of age.

Besides child age, it is also  conceivable that confusion rates depend on environmental factors (e.g. time spend outside, exposure to noise, etc.) and vary across languages. We therefore sought to compare the confusion rates for corpora drawing from urban English-speaking populations with confusion rates estimated from recordings of rural and non-English speaking populations. To this end, we drew from corpora that sampled ``rural'' populations in Papua New Guinea \citep{cristia2020lena}, the Solomon Islands \citep{Cassar2025},  Vanuatu \citep{Cristia2023}, and Bolivia \citep{Scaff2023}. Some of these recordings were done with devices other than \gls{lena}. Moreover, all of these languages are considered under-resourced, which may mean that algorithms built on cumulative knowledge in the speech technology literature may be particularly ill-suited to them. In any case, English was seldom if ever spoken in these recordings, which makes them mismatch in training set with \gls{lena} (which was trained on North American English audio). As a result, our comparison conflates across many dimensions, all of which predict poorer algorithm performance in the ``rural'' recordings. Thus, while this comparison does not allow to isolate the effect of the environment -- urban vs rural --, language, or recording device (which we cannot do due to limited sample size), it gives an idea about whether any of these factors could alter classification errors.  

\begin{figure}[H]
    \centering
    \includegraphics[width=0.8\textwidth]{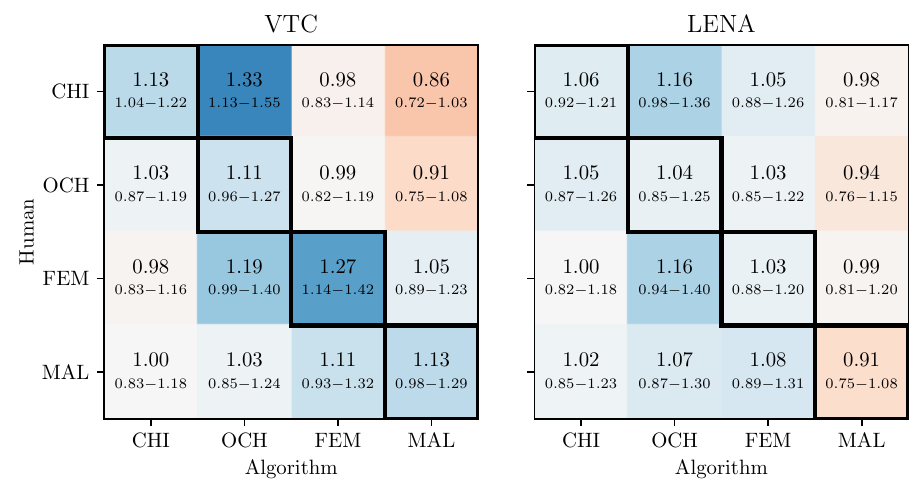}
    \caption{Rural/urban confusion rates for \gls{vtc} (left) and \gls{lena} (right). The latter include non-English speaking, non-WEIRD populations and recorders other than \gls{lena}. Values greater than one (blue cells) signal higher confusion rates among rural corpora than among urban corpora. 95\% credible intervals are indicated underneath each value. }
    \label{fig:confusion_environment}
\end{figure}

Figure \ref{fig:confusion_environment} finds modest differences, most of them being statistically non-significant. 

\subsection{\label{section:confidence-scores}The potential of classifiers' confidence scores as covariates}

\begin{figure}[H]
    \centering
    \includegraphics[width=0.95\linewidth]{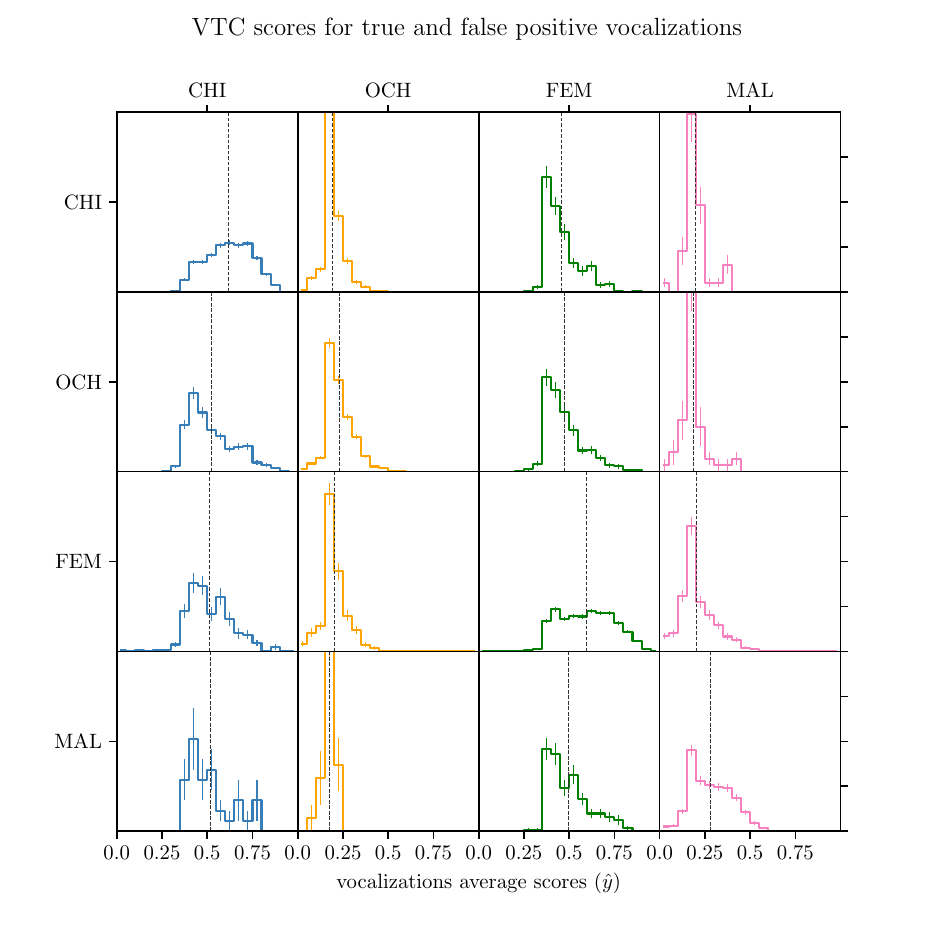}
    \caption{Distribution of the confidence score of \gls{vtc} for each vocalization given the detected speaker type (in columns) and the true speaker (in rows). Dashed vertical lines indicate the mean of the distribution. Vocalizations for which the  speaker is correctly identified exhibit higher confidence scores, which suggests these scores could be used as informative covariates in a calibration approach.}
    \label{fig:confidence-scores}
\end{figure}

\subsection{\label{appendix:icc}Inter-rater agreement}

\input{tables/icc_table}

\subsection{\label{appendix:correlations}\gls{vtc} and \gls{lena} produce inconsistent correlation estimates}

\begin{figure}[H]
\centering
\begin{subfigure}[t]{0.29\textwidth}  
    \centering 
    \includegraphics[width=1\textwidth]{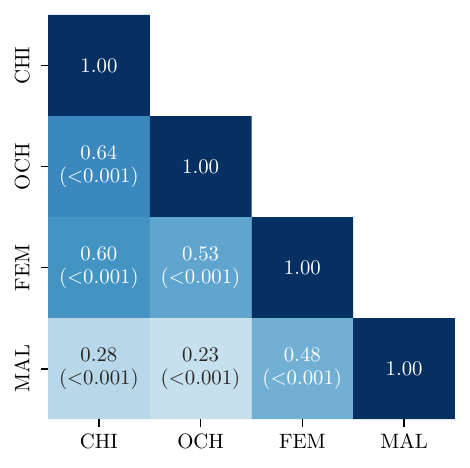}
    \caption{VTC}
\end{subfigure}\hspace{0.25cm}%
\begin{subfigure}[t]{0.29\textwidth}  
    \centering 
    \includegraphics[width=1\textwidth]{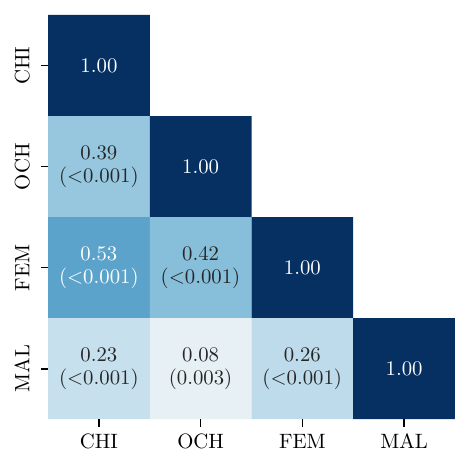
    }
    \caption{LENA}
\end{subfigure}
\caption{\label{fig:recordings}Correlation between the quantity of vocalizations attributed to each speaker across recordings. The correlation matrix is extracted using a hierarchical multivariate log-normal model described in Section \S\ref{section:behavior_surrogate_models_specification}. Estimates from manual annotations are not included due to a lack of data at the full-recording level.}
\end{figure}

\begin{figure}[H]
\centering
\begin{subfigure}[t]{0.33\textwidth}  
    \centering 
    \includegraphics[width=1\textwidth]{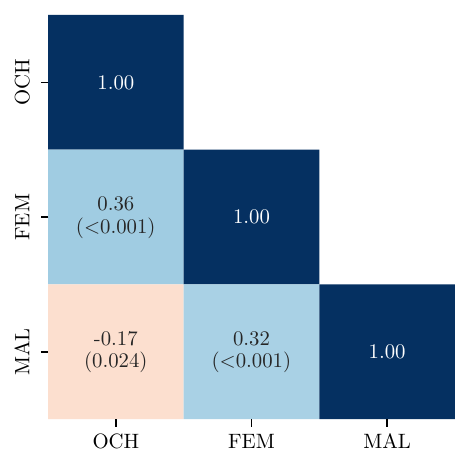}
    \caption{VTC}
\end{subfigure}\hspace{0.25cm}%
\begin{subfigure}[t]{0.33\textwidth}  
    \centering 
    \includegraphics[width=1\textwidth]{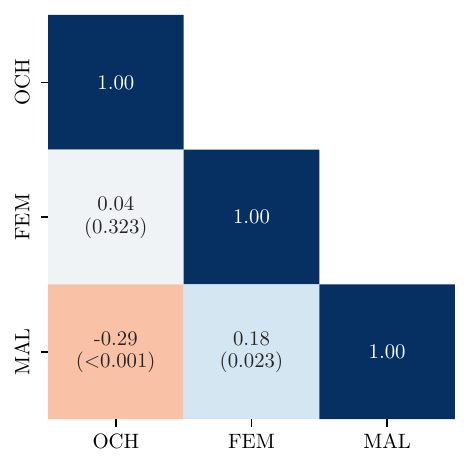}
    \caption{LENA}
\end{subfigure}
\caption{\label{fig:children}Correlation between the quantity of vocalizations attributed to each speaker across children. ``CHI'' is omitted since it varies significantly with age. The correlation matrix is extracted using a hierarchical multivariate log-normal model described in Section \S\ref{section:behavior_surrogate_models_specification}.}
\end{figure}

While revealing of classification errors, correlations between speech quantities measured in short clips bear little importance in themselves. Users are often more interested in correlations between vocalization counts aggregated at the level of whole recordings or the level of each child -- those are shown in Figures \ref{fig:recordings} and \ref{fig:children} respectively. These correlations exhibit similar issues, with \gls{vtc} and \gls{lena} reporting inconsistent correlations -- particularly for those with vocalizations attributed to other children. This further suggests that the estimation of such correlations is affected by classification errors.

\subsubsection{\label{section:behavior_surrogate_models_specification}Surrogate models for correlation estimates}

In order to evaluate recording-level and child-level correlations between speakers, we use surrogate models as alternatives to the above model.

\paragraph{Recording-level correlations between speakers}

In this model for each recording $k$ of child $c$, the vocalization counts are now modeled using a multivariate log-normal distribution:
\begin{align}
\log \mathbf{v}_k^{\text{recs}} \sim \text{MVN}\left(\boldsymbol{\mu}_k - \frac{\text{diag}(\Sigma)}{2}, \Sigma\right)
\end{align}
where $\mathbf{v}_k^{\text{recs}} = (v_{k,\text{CHI}}^{\text{recs}}, v_{k,\text{OCH}}^{\text{recs}}, v_{k,\text{FEM}}^{\text{recs}}, v_{k,\text{MAL}}^{\text{recs}})$ is the vector of vocalization counts for all speaker types.

The mean vector $\boldsymbol{\mu}_k$ is defined as:
\begin{align}
\mu_{k,\text{CHI}} &= \log(\mu^{\text{pop}}_{\text{CHI}}) + \chi_k\\
\mu_{k,s} &= \log(\mu_{c,s}^{\text{child}}), \quad s \in \{\text{OCH}, \text{FEM}, \text{MAL}\}
\end{align}

The child's developmental effect $\chi_k$ is:
\begin{align}
\chi_k &= \alpha_c^{\text{dev}} \cdot \frac{\text{age}_k}{12} + 
    \beta^{\text{dev}} \cdot \frac{\text{age}_k}{12} \cdot \frac{\mu_{c,\text{ADU}}^{\text{child}} - \mu_{\text{ADU}}}{\sigma_{\text{ADU}}}
\end{align}

Note that this formulation removes the direct effect term ($\beta^{\text{direct}}$) that was present in the original model and introduces a covariance structure $\Sigma$ between the different speaker types. The term $-\text{diag}(\Sigma)/2$ ensures the expected values after exponentiation match the desired means.

The covariance matrix $\Sigma$ is parameterized through its Cholesky decomposition $L_\Sigma$ such that $\Sigma = L_\Sigma L_\Sigma^T$.

All other aspects of the model (child-level parameters, population-level parameters, developmental effects, and sibling effects) remain unchanged from the original specification.

\paragraph{Child-level correlations between speakers}

In a third model aimed at measuring correlations between speakers at the child-level, the child-level parameters are assumed to follow a multivariate log-normal distribution:
\begin{align}
\log(\boldsymbol{\mu}_c^{\text{child}}) \sim \text{MVN}\left(\log(\boldsymbol{\mu}^{\text{pop}}_{2:n}) - \frac{\text{diag}(\Sigma_{\text{child}})}{2}, \Sigma_{\text{child}}\right)
\end{align}
where $\boldsymbol{\mu}_c^{\text{child}} = (\mu_{c,\text{OCH}}^{\text{child}}, \mu_{c,\text{FEM}}^{\text{child}}, \mu_{c,\text{MAL}}^{\text{child}})$ and $\boldsymbol{\mu}^{\text{pop}}_{2:n} = (\mu^{\text{pop}}_{\text{OCH}}, \mu^{\text{pop}}_{\text{FEM}}, \mu^{\text{pop}}_{\text{MAL}})$.


The parameters for defining the adult speech influence are now:
\begin{align}
\mu_{\text{ADU}} &= \mu^{\text{pop}}_{\text{FEM}} + \mu^{\text{pop}}_{\text{MAL}}\\
\sigma_{\text{ADU}} &= \sqrt{(\exp(\Sigma^{\text{child}}_{\text{FEM,FEM}}) - 1)(\mu^{\text{pop}}_{\text{FEM}})^2 + (\exp(\Sigma^{\text{child}}_{\text{MAL,MAL}}) - 1)(\mu^{\text{pop}}_{\text{MAL}})^2}
\end{align}

All other model components (developmental effects, population-level parameters) remain unchanged from the first model.

The key difference in this model is that it captures correlations between different speaker types at the child level through the multivariate log-normal distribution, while maintaining the original Gamma distributions for the recording-level observations.

\begin{table}[H]
    \centering
    \input{tables/dof_multivariate}
    \caption{Summary of observations and parameters entering the multivariate uncalibrated models.}
    \label{table:summary_multivariate}
\end{table}

\subsection{\label{appendix:multiple-algos}Aggregating across multiple algorithms}

Our approach could be adapted to aggregate annotations from multiple algorithms. We decided to leave this to future work for both conceptual and practical reasons. On the conceptual side, the present work draws from \gls{vtc} and \gls{lena}, which mostly differ in overall performance and recall-precision trade-off. We expect gains in such a case to be minimal, in contrast to combining two algorithms that have complementary abilities; e.g., if algorithm A was better at telling apart male from female, and algorithm B was better at telling apart key child from other children. On the practical side, the model described below failed to converge within the time limits imposed by our cluster (48h per chain), suggesting its practical applicability is limited under traditional Hamiltonian Monte Carlo.

Aggregating across two algorithms, the likelihood factorizes such that 
\[
\log P(\text{algo}(A), \text{algo}(B) \mid \text{truth}) = \log P(\text{algo}(A) \mid \text{truth}) + \log P(\text{algo}(B) \mid \text{truth}).
\]
This means that in Stan, the likelihood computation can be decomposed accordingly:

\begin{lstlisting}[language=C]
model {
    // ...
    // inverse confusion model
    target += reduce_sum(
       inverse_model_lpdf, actual_confusion_algo1, 1,
       n_recs, n_classes, recs_duration,
       vocs_algo1, truth_vocs, tau1
    );
    target += reduce_sum(
       inverse_model_lpdf, actual_confusion_algo2, 1,
       n_recs, n_classes, recs_duration,
       vocs_algo2, truth_vocs, tau2
    );
    // ...
}
\end{lstlisting}

Since there can be correlations between the errors made by different algorithms, we modelled these correlations as follows:

\begin{lstlisting}[language=C]
 for (k in 1:n_recs) {
        actual_confusion_rate_algo1[k] = confusion_baseline_algo1[k];
        actual_confusion_rate_algo2[k] = confusion_baseline_algo2[k] .* exp(0.1 * beta_algo1_algo2 .* (actual_confusion_rate_algo1[k]./mus_algo1-1));
 }
\end{lstlisting}

The implementation of this model can be found at \url{https://gin.g-node.org/LAAC-LSCP/speaker-confusion-model/src/main/code/models/dev_combined_simple.stan}


\end{document}

%% file: acronyms.tex
\newacronym{lena}{LENA\texttrademark}{Language ENvironment Analysis}
\newacronym{vtc}{VTC}{Voice Type Classifier}
\newacronym{dpgmm}{DPGMM}{Dirichlet Process Gaussian Mixture Model}

%% file: figures/Fig2d.tex
\definecolor{chi}{HTML}{377eb8}
\definecolor{och}{HTML}{ff7f00}
\definecolor{fem}{HTML}{4daf4a}
\definecolor{mal}{HTML}{f781bf}
\begin{tikzpicture}[baseline={(0,0)}]
    \coordinate (align-top) at (0,0);
    
    \node[state,minimum width=6pt] (chi) at ($(align-top) + (0,0)$) {\textbf{CHI}\\(truth)};
    \node[state,minimum width=6pt] (och) at ($(align-top) + (2.5,0)$) {\textbf{OCH}\\(truth)};
    \node[state,minimum width=6pt] (fem) at ($(align-top) + (5,0)$) {\textbf{FEM}\\(truth)};
    \node[state,minimum width=6pt] (mal) at ($(align-top) + (7.5,0)$) {\textbf{MAL}\\(truth)};
    
    \node[state,minimum width=6pt,fill=chi!20] (chi_algo) at ($(chi) + (0,-3)$) {\textbf{CHI}\\(algo)};
    \node[state,minimum width=6pt,fill=och!20] (och_algo) at ($(och) + (0,-3)$) {\textbf{OCH}\\(algo)};
    \node[state,minimum width=6pt,fill=fem!20] (fem_algo) at ($(fem) + (0,-3)$) {\textbf{FEM}\\(algo)};
    \node[state,minimum width=6pt,fill=mal!20] (mal_algo) at ($(mal) + (0,-3)$) {\textbf{MAL}\\(algo)};

        \draw[->, bend right=10] (chi.south) to (chi_algo);
        \draw[->, bend right=8] (chi.south) to (och_algo);
        \draw[->, bend right=1, line width=1mm,color=red] (chi.south) to (fem_algo);
        \draw[->, bend left=5, line width=1mm,color=red] (chi.south) to (mal_algo);
    
        \draw[->, bend right=10] (och.south) to (chi_algo);
        \draw[->, bend right=8] (och.south) to (och_algo);
        \draw[->, bend right=4, line width=1mm,color=red] (och.south) to (fem_algo);
        \draw[->, bend right=1, line width=1mm,color=red] (och.south) to (mal_algo);
    
        \draw[->, bend right=10] (fem.south) to (chi_algo);
        \draw[->, bend right=8] (fem.south) to (och_algo);
        \draw[->, bend left=1] (fem.south) to (fem_algo);
        \draw[->, bend left=5, line width=1mm,color=red] (fem.south) to (mal_algo);
        \draw[->, bend right=10] (fem.south) to (chi_algo);
        \draw[->, bend left=1, line width=1mm,color=black] (fem.south) to (fem_algo);
    
        \draw[->, bend right=10] (mal.south) to (chi_algo);
        \draw[->, bend right=8] (mal.south) to (och_algo);
        \draw[->, bend left=1, line width=1mm,color=red] (mal.south) to (fem_algo);
        \draw[->, bend left=5, line width=1mm,color=black] (mal.south) to (mal_algo);

        \node[state,minimum width=6pt] (fem_prop) at (6.25,-6) {$\dfrac{\text{\textbf{FEM}}}{\text{\textbf{FEM+MAL}}}$\\(algo)};
        \filldraw[fem!20] (fem_prop.south west) -- (fem_prop.north west) -- (fem_prop.north east) -- cycle;
        \filldraw[mal!20] (fem_prop.south west) -- (fem_prop.south east) -- (fem_prop.north east) -- cycle;
        \node[state,minimum width=6pt] (fem_prop) at (6.25,-6) {$\dfrac{\text{\textbf{FEM}}}{\text{\textbf{FEM+MAL}}}$\\(algo)};
        
        \draw[->, line width=1mm] (fem_algo.south) to [bend right=10] (fem_prop.north);
        \draw[->, line width=1mm] (mal_algo.south) to [bend left=10] (fem_prop.north);

        
        
        
        
    \end{tikzpicture}

%% file: figures/Fig2a_new.tex
\definecolor{chi}{HTML}{377eb8}
\definecolor{och}{HTML}{ff7f00}
\definecolor{fem}{HTML}{4daf4a}
\definecolor{mal}{HTML}{f781bf}

\begin{tikzpicture}[baseline={(0,0)}]
    \coordinate (align-top) at (0,0);
        \node[state] (chi) at (0,0) {\textbf{CHI}\\(truth)};
    
        \node[state] (och) at (2.5,0) {\textbf{OCH}\\(truth)};
    
        \node[state] (fem) at (5,0) {\textbf{FEM}\\(truth)};
    
        \node[state] (mal) at (7.5,0) {\textbf{MAL}\\(truth)};
    
        \node[state,fill=chi!15] (algo_chi) at (0,-3) {\textbf{CHI}\\(algo)};
    
        \node[state,fill=och!15] (algo_och) at (2.5,-3) {\textbf{OCH}\\(algo)};
    
        \node[state,fill=fem!15] (algo_fem) at (5,-3) {\textbf{FEM}\\(algo)};
    
        \node[state,fill=mal!15] (algo_mal) at (7.5,-3) {\textbf{MAL}\\(algo)};
    
        \draw[->, bend right=10, line width=1mm,color=black] (chi.south) to (algo_chi);
        \draw[->, bend right=8] (chi.south) to (algo_och);
        \draw[->, bend right=1, line width=1mm,color=red] (chi.south) to (algo_fem);
        \draw[->, bend left=5] (chi.south) to (algo_mal);
    
        \draw[->, bend right=10] (och.south) to (algo_chi);
        \draw[->, bend right=8, line width=1mm,color=black] (och.south) to (algo_och);
        \draw[->, bend right=4] (och.south) to (algo_fem);
        \draw[->, bend right=1] (och.south) to (algo_mal);
    
        \draw[->, bend right=10] (fem.south) to (algo_chi);
        \draw[->, bend right=8] (fem.south) to (algo_och);
        \draw[->, bend left=1] (fem.south) to (algo_fem);
        \draw[->, bend left=5] (fem.south) to (algo_mal);

        \draw[->, bend right=10, line width=1mm,color=red] (fem.south) to (algo_chi);
        \draw[->, bend left=1, line width=1mm,color=black] (fem.south) to (algo_fem);
    
        \draw[->, bend right=10] (mal.south) to (algo_chi);
        \draw[->, bend right=8] (mal.south) to (algo_och);
        \draw[->, bend left=1] (mal.south) to (algo_fem);
        \draw[->, bend left=5, line width=1mm,color=black] (mal.south) to (algo_mal);

        \draw[<->, bend left=45, line width=1mm,color=black,dashed,color=gray] (chi.north) to node[midway,above,sloped,font=\large\bfseries,color=gray] {?} (fem.north);
        
    \end{tikzpicture}

%% file: figures/Fig2c_new.tex
\definecolor{chi}{HTML}{377eb8}
\definecolor{och}{HTML}{ff7f00}
\definecolor{fem}{HTML}{4daf4a}
\definecolor{mal}{HTML}{f781bf}
\begin{tikzpicture}[baseline={(0,0)}]
    \coordinate (align-top) at (0,0);
    
    \node[state,minimum width=6pt] (sibs) at (3.5, 2) {\textbf{Siblings}};
    
    \node[state,minimum width=6pt] (chi) at (0,0) {\textbf{CHI}\\(truth)};
    \node[state,minimum width=6pt] (och) at (2.5,0) {\textbf{OCH}\\(truth)};
    \node[state,minimum width=6pt] (fem) at (5,0) {\textbf{FEM}\\(truth)};
    \node[state,minimum width=6pt] (mal) at (7.5,0) {\textbf{MAL}\\(truth)};
    
    \node[state,minimum width=6pt,fill=chi!20] (chi_algo) at (0,-3) {\textbf{CHI}\\(algo)};
    \node[state,minimum width=6pt,fill=och!20] (och_algo) at (2.5,-3) {\textbf{OCH}\\(algo)};
    \node[state,minimum width=6pt,fill=fem!20] (fem_algo) at (5,-3) {\textbf{FEM}\\(algo)};
    \node[state,minimum width=6pt,fill=mal!20] (mal_algo) at (7.5,-3) {\textbf{MAL}\\(algo)};

    
    \draw[->,line width=1mm,color=gray] (sibs) -- (och);
    \draw[->,line width=1mm,dashed,color=gray] (sibs) -- node[midway,above,sloped,font=\large\bfseries,color=gray] {?} (fem);
    \draw[->,line width=1mm,dashed,color=gray] (sibs) -- node[midway,above,sloped,font=\large\bfseries,color=gray] {?} (mal);

    \draw[->, bend right=10] (chi.south) to (chi_algo);
    \draw[->, bend right=8] (chi.south) to (och_algo);
    \draw[->, bend right=1] (chi.south) to (fem_algo);
    \draw[->, bend left=5] (chi.south) to (mal_algo);

    \draw[->, bend right=10] (och.south) to (chi_algo);
    \draw[->, bend right=8] (och.south) to (och_algo);
    \draw[->, bend right=4, line width=1mm,color=red] (och.south) to (fem_algo);
    \draw[->, bend right=1, line width=1mm,color=red] (och.south) to (mal_algo);

    \draw[->, bend right=10] (fem.south) to (chi_algo);
    \draw[->, bend right=8] (fem.south) to (och_algo);
    \draw[->, bend left=1] (fem.south) to (fem_algo);
    \draw[->, bend left=5] (fem.south) to (mal_algo);

    \draw[->, bend right=10] (fem.south) to (chi_algo);
    \draw[->, bend left=1, line width=1mm,color=black] (fem.south) to (fem_algo);
    
    \draw[->, bend right=10] (mal.south) to (chi_algo);
    \draw[->, bend right=8] (mal.south) to (och_algo);
    \draw[->, bend left=1] (mal.south) to (fem_algo);
    \draw[->, bend left=5, line width=1mm,color=black] (mal.south) to (mal_algo);
    
    
\end{tikzpicture}

%% file: figures/behavior_model.tex
\definecolor{chi}{HTML}{377eb8}
\definecolor{och}{HTML}{ff7f00}
\definecolor{fem}{HTML}{4daf4a}
\definecolor{mal}{HTML}{f781bf}

\tikzset{node distance = 2.2cm,round/.style={rectangle, rounded corners}}
\begin{tikzpicture}[auto]

    \node (vkirecs_CHI) [draw, fill=chi!20, round] {$v_{k,\text{CHI}}^\text{recs}$};
    \node (vkirecs_fem) [right of=vkirecs_CHI, draw, fill=fem!20, round] {$v_{k,\text{FEM}}^\text{recs}$};
    \node (vkirecs_mal) [right of=vkirecs_fem, draw, fill=mal!20, round] {$v_{k,\text{MAL}}^\text{recs}$};
    \node (vkirecs_och) [right of=vkirecs_mal, draw, fill=och!20, round] {$v_{k,\text{OCH}}^\text{recs}$};
    
    \node (mu_CHI_recs) [above of=vkirecs_CHI, draw, fill=white,yshift=-0.5cm, round] {$\mu_{k,\text{CHI}}^{\text{rec}}$};
    \node (mu_fem) [right of=mu_CHI_recs, draw, fill=white, yshift=+0.5cm, round] {$\mu_{c,\text{FEM}}^{\text{child}}$};
    \node (mu_mal) [right of=mu_fem, draw, fill=white, round] {$\mu_{c,\text{MAL}}^{\text{child}}$};
    \node (mu_och) [right of=mu_mal, draw, fill=white, round] {$\mu_{c,\text{OCH}}^{\text{child}}$};

    \node (n_sibs) [right of=mu_och, draw, fill=blue!20, round] {$n_c^{\text{sibs}}$};

    \node (mu_och_pop) [above of=mu_och, draw, fill=white,yshift=-0.5cm, round] {$\mu_{\text{OCH}}^{\text{pop}}$};
    \node (mu_fem_pop) [above of=mu_fem, draw, fill=white,yshift=-0.5cm, round] {$\mu_{\text{FEM}}^{\text{pop}}$};
    \node (mu_mal_pop) [above of=mu_mal, draw, fill=white,yshift=-0.5cm, round] {$\mu_{\text{MAL}}^{\text{pop}}$};
    \node (mu_chi_pop) [left of=mu_fem_pop, draw, fill=white, round] {$\mu_{\text{CHI}}^{\text{pop}}$};
    
    \node (age_k) [left of=mu_CHI_recs, draw, fill=blue!20, round] {age$_k$};
    
    \draw[->,black] (age_k) -- (mu_CHI_recs);
    \draw[->,bend right=35,thick,postaction={decorate,decoration={text along path, reverse path,
    text={|\sffamily|{$\beta_{dev}$}{}}, text align=center,raise=+1ex,text color=black!50!green}},color=black!50!green] (mu_fem.west) to (mu_CHI_recs.north east);
    \draw[->,bend right=45,thick,postaction={decorate,decoration={text along path, reverse path,
    text={|\sffamily|{$\beta_{dev}$}{}}, text align=center,raise=+1ex, text color=black!50!green}},color=black!50!green] (mu_mal.north) to (mu_CHI_recs.north east);
    
    \draw[->,black] (mu_CHI_recs) -- (vkirecs_CHI);
    \draw[->,black] (mu_fem) -- (vkirecs_fem);
    \draw[->,black] (mu_mal) -- (vkirecs_mal);
    \draw[->,black] (mu_och) -- (vkirecs_och);

    \draw[->,black] (mu_chi_pop) -- (mu_CHI_recs);
    \draw[->,black] (mu_fem_pop) -- (mu_fem);
    \draw[->,black] (mu_mal_pop) -- (mu_mal);
    \draw[->,black] (mu_och_pop) -- (mu_och);

    \draw[->,color=black!50!green] (age_k.east) -- node[above,align=center,color=black!50!green] {$\alpha_c^{\text{dev}}$}(mu_CHI_recs.west);
    
    \draw[->,color=black!50!green,thick] (n_sibs.west) to (mu_och.east);
    \draw[->,bend left=30, thick,color=black!50!green] (n_sibs.south west) to (mu_mal.south east);
    \draw[->,bend left=35, thick,color=black!50!green] (n_sibs.south west) to (mu_fem.south east);

    \draw[->,bend right=30,thick,postaction={decorate,decoration={text along path, reverse path,
    text={|\sffamily|{$\beta_{direct}$}{}}, text align=center,raise=+1ex,text color=black!50!green}},color=black!50!green] (vkirecs_fem.north) to (vkirecs_CHI.north east);
    \draw[->,bend right=25,thick,postaction={decorate,decoration={text along path, reverse path,
    text={|\sffamily|{$\beta_{direct}$}{}}, text align=center,raise=+1ex, text color=black!50!green}},color=black!50!green] (vkirecs_mal.north) to (vkirecs_CHI.north east);
    \draw[->,black] (mu_CHI_recs) -- (vkirecs_CHI);
    
\end{tikzpicture}

%% file: figures/full_model.tex
\definecolor{chi}{HTML}{377eb8}
\definecolor{och}{HTML}{ff7f00}
\definecolor{fem}{HTML}{4daf4a}
\definecolor{mal}{HTML}{f781bf}
\tikzset{node distance = 2.2cm,round/.style={rectangle, rounded corners}}

\begin{tikzpicture}[auto]

    \node (vkirecs_CHI) [draw, align=center, fill=chi!5, round] {$v_{k,\text{CHI}}^\text{recs}$};
    \node (vkirecs_fem) [right of=vkirecs_CHI, draw, fill=fem!5, round] {$v_{k,\text{FEM}}^\text{recs}$};
    \node (vkirecs_mal) [right of=vkirecs_fem, draw, fill=mal!5, round] {$v_{k,\text{MAL}}^\text{recs}$};
    \node (vkirecs_och) [right of=vkirecs_mal, draw, fill=och!5, round] {$v_{k,\text{OCH}}^\text{recs}$};
    
    \node (mu_CHI_recs) [draw, above of=vkirecs_CHI, draw, fill=white,yshift=-0.5cm, round] {$\mu_{k,\text{CHI}}^{\text{rec}}$};
    \node (mu_fem) [right of=mu_CHI_recs, draw, fill=white, yshift=+0.5cm, round] {$\mu_{c,\text{FEM}}^{\text{child}}$};
    \node (mu_mal) [right of=mu_fem, draw, fill=white, round] {$\mu_{c,\text{MAL}}^{\text{child}}$};
    \node (mu_och) [right of=mu_mal, draw, fill=white, round] {$\mu_{c,\text{OCH}}^{\text{child}}$};

    \node (n_sibs) [right of=mu_och, draw, fill=blue!20, round] {$n_c^{\text{sibs}}$};
    
    \node (age_k) [left of=mu_CHI_recs, draw, fill=blue!20, round] {age$_k$};
    
    \node (mu_och_pop) [above of=mu_och, draw, fill=white,yshift=-0.5cm, round] {$\mu_{\text{OCH}}^{\text{pop}}$};
    \node (mu_fem_pop) [above of=mu_fem, draw, fill=white,yshift=-0.5cm, round] {$\mu_{\text{FEM}}^{\text{pop}}$};
    \node (mu_mal_pop) [above of=mu_mal, draw, fill=white,yshift=-0.5cm, round] {$\mu_{\text{MAL}}^{\text{pop}}$};
    \node (mu_chi_pop) [left of=mu_fem_pop, draw, fill=white, round] {$\mu_{\text{CHI}}^{\text{pop}}$};

    \draw[->,black, round] (age_k) -- (mu_CHI_recs);
    \draw[->,bend right=35,thick,postaction={decorate,decoration={text along path, reverse path,
    text={|\sffamily|{$\beta_{dev}$}{}}, text align=center,raise=+1ex, text color=black!50!green}},color=black!50!green] (mu_fem.west) to (mu_CHI_recs.north east);
    \draw[->,red,bend right=45,thick,postaction={decorate,decoration={text along path, reverse path,
    text={|\sffamily|{$\beta_{dev}$}{}}, text align=center,raise=+1ex,text color=black!50!green}}, color=black!50!green] (mu_mal.north) to (mu_CHI_recs.north east);
    \draw[->,black] (mu_CHI_recs) -- (vkirecs_CHI);
    \draw[->,black] (mu_fem) -- (vkirecs_fem);
    \draw[->,black] (mu_mal) -- (vkirecs_mal);
    \draw[->,black] (mu_och) -- (vkirecs_och);

    \draw[->,black] (mu_chi_pop) -- (mu_CHI_recs);
    \draw[->,black] (mu_fem_pop) -- (mu_fem);
    \draw[->,black] (mu_mal_pop) -- (mu_mal);
    \draw[->,black] (mu_och_pop) -- (mu_och);

    \draw[->,color=black!50!green] (age_k.east) -- node[above,align=center,color=black!50!green] {$\alpha_c^{\text{dev}}$}(mu_CHI_recs.west) ;

    \draw[->,color=black!50!green,thick] (n_sibs.west) to (mu_och.east);
    \draw[->,bend left=30, thick,color=black!50!green] (n_sibs.south west) to (mu_mal.south east);
    \draw[->,bend left=35, thick,color=black!50!green] (n_sibs.south west) to (mu_fem.south east);

    \def\myshift#1{\raisebox{1ex}}
    \draw[->,bend right=30,thick,postaction={decorate,decoration={text along path, reverse path,
    text={|\sffamily|{$\beta_{direct}$}{}}, text align=center,raise=+1ex,text color=black!50!green}},color=black!50!green] (vkirecs_fem.north) to (vkirecs_CHI.north east);
    \draw[->,bend right=25,thick,postaction={decorate,decoration={text along path, reverse path,
    text={|\sffamily|{$\beta_{direct}$}{}}, text align=center,raise=+1ex, text color=black!50!green}},color=black!50!green] (vkirecs_mal.north) to (vkirecs_CHI.north east);
    \draw[->,black] (mu_CHI_recs) -- (vkirecs_CHI);
    
    \node (nkirecs_CHI) [below of=vkirecs_CHI, draw, fill=chi!50, yshift=0cm, round] {$n_{k,\text{CHI}}^\text{recs}$};
    \node (nkirecs_fem) [below of=vkirecs_fem, draw, fill=fem!50, yshift=0cm, round] {$n_{k,\text{FEM}}^\text{recs}$};
    \node (nkirecs_mal) [below of=vkirecs_mal, draw, fill=mal!50, yshift=0cm, round] {$n_{k,\text{MAL}}^\text{recs}$};
    \node (nkirecs_och) [below of=vkirecs_och,  draw, fill=och!50, yshift=0cm, round] {$n_{k,\text{OCH}}^\text{recs}$};

    \node (nkiclips_CHI) [below of=nkirecs_CHI, draw, fill=chi!50, yshift=+0.25cm, round] {$n_{k',\text{CHI}}^\text{clips}$};
    \node (nkiclips_fem) [below of=nkirecs_fem, draw, fill=fem!50, yshift=+0.25cm, round] {$n_{k',\text{FEM}}^\text{clips}$};
    \node (nkiclips_mal) [below of=nkirecs_mal, draw, fill=mal!50, yshift=+0.25cm, round] {$n_{k',\text{MAL}}^\text{clips}$};
    \node (nkiclips_och) [below of=nkirecs_och,  draw, fill=och!50, yshift=+0.25cm, round] {$n_{k',\text{OCH}}^\text{clips}$};

    \node (vkiclips_CHI) [below of=nkiclips_CHI, draw, fill=chi!50,yshift=-0.25cm, round] {$v_{k',\text{CHI}}^\text{clips}$};
    \node (vkiclips_fem) [below of=nkiclips_fem, draw, fill=fem!50,yshift=-0.25cm, round] {$v_{k',\text{FEM}}^\text{clips}$};
    \node (vkiclips_mal) [below of=nkiclips_mal, draw, fill=mal!50,yshift=-0.25cm, round] {$v_{k',\text{MAL}}^\text{clips}$};
    \node (vkiclips_och) [below of=nkiclips_och,  draw, fill=och!50,yshift=-0.25cm, round] {$v_{k',\text{OCH}}^\text{clips}$};

    \node [fit=(vkirecs_CHI) (nkirecs_och), inner sep=0mm] (block) {};
    
    \foreach \i in {0,...,3} {
        \pgfmathsetmacro{\xshift}{0.25+0.5*\i}
        \node [rectangle, draw, minimum width=0.5cm, minimum height=0.25cm,fill=chi!20] (sub\i) at ($(block.west) + (\xshift,0)$) {};
    }

    \foreach \i in {4,...,7} {
        \pgfmathsetmacro{\xshift}{0.25+0.5*\i}
        \node [rectangle, draw, minimum width=0.5cm, minimum height=0.25cm,fill=fem!20] (sub\i) at ($(block.west) + (\xshift,0)$) {};
    }

    \foreach \i in {8,...,11} {
        \pgfmathsetmacro{\xshift}{0.25+0.5*\i}
        \node [rectangle, draw, minimum width=0.5cm, minimum height=0.25cm,fill=mal!20] (sub\i) at ($(block.west) + (\xshift,0)$) {};
    }

    \foreach \i in {12,...,15} {
        \pgfmathsetmacro{\xshift}{0.25+0.5*\i}
        \node [rectangle, draw, minimum width=0.5cm, minimum height=0.25cm,fill=och!20] (sub\i) at ($(block.west) + (\xshift,0)$) {};
    }

    \node (lambda_kij) [draw, left of=sub0,xshift=0.25cm, round] {$\lambda_{kij}$};

    \node [fit=(vkiclips_CHI) (nkiclips_och), inner sep=0mm] (block_clips) {};
    
    \foreach \i in {0,...,3} {
        \pgfmathsetmacro{\xshift}{0.25+0.5*\i}
        \node [rectangle, draw, minimum width=0.5cm, minimum height=0.25cm,fill=chi!20] (subc\i) at ($(block_clips.west) + (\xshift,0)$) {};
    }

    \foreach \i in {4,...,7} {
        \pgfmathsetmacro{\xshift}{0.25+0.5*\i}
        \node [rectangle, draw, minimum width=0.5cm, minimum height=0.25cm,fill=fem!20] (subc\i) at ($(block_clips.west) + (\xshift,0)$) {};
    }

    \foreach \i in {8,...,11} {
        \pgfmathsetmacro{\xshift}{0.25+0.5*\i}
        \node [rectangle, draw, minimum width=0.5cm, minimum height=0.25cm,fill=mal!20] (subc\i) at ($(block_clips.west) + (\xshift,0)$) {};
    }

    \foreach \i in {12,...,15} {
        \pgfmathsetmacro{\xshift}{0.25+0.5*\i}
        \node [rectangle, draw, minimum width=0.5cm, minimum height=0.25cm,fill=och!20] (subc\i) at ($(block_clips.west) + (\xshift,0)$) {};
    }

    \node (lambda_clips_kij) [draw, left of=subc0,xshift=0.25cm, round] {$\lambda_{k'ij}$};

    \node (Lambda_kij) [draw, left of=nkirecs_CHI,xshift=-2cm,yshift=-1cm,align=center, round] {$\mu_{ij},\alpha_{ij}$\\($\bm{\nu}$)};
    
    \draw[->,chi] (vkirecs_CHI) to  [out=-90,in=90] (sub0.north);
    \draw[->,chi] (vkirecs_CHI) to  [out=-90,in=90] (sub1.north);
    \draw[->,chi] (vkirecs_CHI) to  [out=-90,in=90] (sub2.north);
    \draw[->,chi] (vkirecs_CHI) to  [out=-90,in=90] (sub3.north);

    \draw[->,fem] (vkirecs_fem) to  [out=-90,in=90] (sub4.north);
    \draw[->,fem] (vkirecs_fem) to  [out=-90,in=90] (sub5.north);
    \draw[->,fem] (vkirecs_fem) to  [out=-90,in=90] (sub6.north);
    \draw[->,fem] (vkirecs_fem) to  [out=-90,in=90] (sub7.north);

    \draw[->,mal] (vkirecs_mal) to  [out=-90,in=90] (sub8.north);
    \draw[->,mal] (vkirecs_mal) to  [out=-90,in=90] (sub9.north);
    \draw[->,mal] (vkirecs_mal) to  [out=-90,in=90] (sub10.north);
    \draw[->,mal] (vkirecs_mal) to  [out=-90,in=90] (sub11.north);

    \draw[->,och] (vkirecs_och) to  [out=-90,in=90] (sub12.north);
    \draw[->,och] (vkirecs_och) to  [out=-90,in=90] (sub13.north);
    \draw[->,och] (vkirecs_och) to  [out=-90,in=90] (sub14.north);
    \draw[->,och] (vkirecs_och) to  [out=-90,in=90] (sub15.north);

    \draw[->,chi] (sub0.south) to  [out=-90,in=90,looseness=0.5] ($(nkirecs_CHI.north)+(-0.1,0)$) ;
    \draw[->,fem] (sub4.south) to  [out=-90,in=90,looseness=0.5] ($(nkirecs_CHI.north)+(0,0)$);
    \draw[->,mal] (sub8.south) to  [out=-90,in=90,looseness=0.5] ($(nkirecs_CHI.north)+(0.1,0)$);
    \draw[->,och] (sub12.south) to  [out=-90,in=90,looseness=0.5] ($(nkirecs_CHI.north)+(0.2,0)$);

    \draw[->,chi]  (sub1.south) to  [out=-90,in=90,looseness=0.5] ($(nkirecs_fem.north)+(-0.1,0)$);
    \draw[->,fem] (sub5.south) to  [out=-90,in=90,looseness=0.5] ($(nkirecs_fem.north)+(0,0)$);
    \draw[->,mal] (sub9.south) to  [out=-90,in=90,looseness=0.5] ($(nkirecs_fem.north)+(+0.1,0)$);
    \draw[->,och] (sub13.south) to  [out=-90,in=90,looseness=0.5] ($(nkirecs_fem.north)+(+0.2,0)$);

    \draw[->,chi]  (sub2.south) to  [out=-90,in=90,looseness=0.5] ($(nkirecs_mal.north)+(-0.1,0)$);
    \draw[->,fem] (sub6.south) to  [out=-90,in=90,looseness=0.5] ($(nkirecs_mal.north)+(0,0)$);
    \draw[->,mal] (sub10.south) to  [out=-90,in=90,looseness=0.5] ($(nkirecs_mal.north)+(+0.1,0)$);
    \draw[->,och] (sub14.south) to  [out=-90,in=90,looseness=0.5] ($(nkirecs_mal.north)+(+0.2,0)$);

    \draw[->,chi]  (sub3.south) to  [out=-90,in=90,looseness=0.5] ($(nkirecs_och.north)+(-0.1,0)$);
    \draw[->,fem] (sub7.south) to  [out=-90,in=90,looseness=0.5] ($(nkirecs_och.north)+(0,0)$);
    \draw[->,mal] (sub11.south) to  [out=-90,in=90,looseness=0.5] ($(nkirecs_och.north)+(+0.1,0)$);
    \draw[->,och] (sub15.south) to  [out=-90,in=90,looseness=0.5]($(nkirecs_och.north)+(+0.2,0)$);

    \draw[->,chi] (vkiclips_CHI) to  [out=90,in=-90] (subc0.south);
    \draw[->,chi] (vkiclips_CHI) to  [out=90,in=-90] (subc1.south);
    \draw[->,chi] (vkiclips_CHI) to  [out=90,in=-90] (subc2.south);
    \draw[->,chi] (vkiclips_CHI) to  [out=90,in=-90] (subc3.south);

    \draw[->,fem] (vkiclips_fem) to  [out=90,in=-90] (subc4.south);
    \draw[->,fem] (vkiclips_fem) to  [out=90,in=-90] (subc5.south);
    \draw[->,fem] (vkiclips_fem) to  [out=90,in=-90] (subc6.south);
    \draw[->,fem] (vkiclips_fem) to  [out=90,in=-90] (subc7.south);

    \draw[->,mal] (vkiclips_mal) to  [out=90,in=-90] (subc8.south);
    \draw[->,mal] (vkiclips_mal) to  [out=90,in=-90] (subc9.south);
    \draw[->,mal] (vkiclips_mal) to  [out=90,in=-90] (subc10.south);
    \draw[->,mal] (vkiclips_mal) to  [out=90,in=-90] (subc11.south);

    \draw[->,och] (vkiclips_och) to  [out=90,in=-90] (subc12.south);
    \draw[->,och] (vkiclips_och) to  [out=90,in=-90] (subc13.south);
    \draw[->,och] (vkiclips_och) to  [out=90,in=-90] (subc14.south);
    \draw[->,och] (vkiclips_och) to  [out=90,in=-90] (subc15.south);

    \draw[->,chi] (subc0.north) to  [out=90,in=-90,looseness=0.5] ($(nkiclips_CHI.south)+(-0.1,0)$) ;
    \draw[->,fem] (subc4.north) to  [out=90,in=-90,looseness=0.5] ($(nkiclips_CHI.south)+(0,0)$);
    \draw[->,mal] (subc8.north) to  [out=90,in=-90,looseness=0.5] ($(nkiclips_CHI.south)+(0.1,0)$);
    \draw[->,och] (subc12.north) to  [out=90,in=-90,looseness=0.5] ($(nkiclips_CHI.south)+(0.2,0)$);

    \draw[->,chi]  (subc1.north) to  [out=90,in=-90,looseness=0.5] ($(nkiclips_fem.south)+(-0.1,0)$);
    \draw[->,fem] (subc5.north) to  [out=90,in=-90,looseness=0.5] ($(nkiclips_fem.south)+(0,0)$);
    \draw[->,mal] (subc9.north) to  [out=90,in=-90,looseness=0.5] ($(nkiclips_fem.south)+(+0.1,0)$);
    \draw[->,och] (subc13.north) to  [out=90,in=-90,looseness=0.5] ($(nkiclips_fem.south)+(+0.2,0)$);

    \draw[->,chi]  (subc2.north) to  [out=90,in=-90,looseness=0.5] ($(nkiclips_mal.south)+(-0.1,0)$);
    \draw[->,fem] (subc6.north) to  [out=90,in=-90,looseness=0.5] ($(nkiclips_mal.south)+(0,0)$);
    \draw[->,mal] (subc10.north) to  [out=90,in=-90,looseness=0.5] ($(nkiclips_mal.south)+(+0.1,0)$);
    \draw[->,och] (subc14.north) to  [out=90,in=-90,looseness=0.5] ($(nkiclips_mal.south)+(+0.2,0)$);

    \draw[->,chi]  (subc3.north) to  [out=90,in=-90,looseness=0.5] ($(nkiclips_och.south)+(-0.1,0)$);
    \draw[->,fem] (subc7.north) to  [out=90,in=-90,looseness=0.5] ($(nkiclips_och.south)+(0,0)$);
    \draw[->,mal] (subc11.north) to  [out=90,in=-90,looseness=0.5] ($(nkiclips_och.south)+(+0.1,0)$);
    \draw[->,och] (subc15.north) to  [out=90,in=-90,looseness=0.5] ($(nkiclips_och.south)+(+0.2,0)$);

    \draw[->,black] (lambda_kij) to (sub0);

    \draw[->,black] (lambda_clips_kij) to (subc0);

    \draw[->,black, bend left=30] (Lambda_kij.north) to (lambda_kij.west);

    \draw[->,black, bend right=30] (Lambda_kij.south) to (lambda_clips_kij.west);

    \node [fit=(nkiclips_och) (vkiclips_och)] (calibration) {};
    \path let \p1=(calibration.north west), \p2 = (calibration.south west) in
       node [right of=calibration, xshift=-0.7cm] {%
       \pgfmathsetmacro\heightoffit{.6*(\y1-\y2)}%
       \resizebox{!}{\heightoffit pt}{\}}%
     };

     \node [right of=calibration, xshift=0.5cm, rotate=90, align=center] (calib) {Calibration\\data};

         \node [fit=(nkirecs_CHI) (nkiclips_CHI)] (algo) {};
    \path let \p1=(algo.north west), \p2 = (algo.south west) in
       node [left of=algo, xshift=0.9cm,rotate=180] {%
       \pgfmathsetmacro\heightoffit{.6*(\y1-\y2)}%
       \resizebox{!}{\heightoffit pt}{\}}%
     };

      \node [left of=algo, xshift=-0.05cm, rotate=90, align=center] (calib) {Algorithm\\output};

      \node [align=center,yshift=-0.4cm] (manual) at ($(vkiclips_fem.south)!0.5!(vkiclips_mal.south)$) {``True'' counts from manual annotations};

      \node [fit=(nkirecs_och) (vkirecs_och)] (fullrecs) {};
    \path let \p1=(fullrecs.north west), \p2 = (fullrecs.south west) in
       node [right of=fullrecs, xshift=-0.7cm] {%
       \pgfmathsetmacro\heightoffit{.6*(\y1-\y2)}%
       \resizebox{!}{\heightoffit pt}{\}}%
     };

     \node [right of=fullrecs, xshift=0.5cm, rotate=90, align=center] (full) {Full\\recordings};
\end{tikzpicture}

%% file: tables/data.tex
\begin{tabular}{lccccccccc}
\toprule
 & \multicolumn{4}{c}{Full recordings} & \multicolumn{5}{c}{Human annotations} \\
\cmidrule(lr){2-5} \cmidrule(lr){6-10}
Corpus & \# Child. & \# Recs & \shortstack{Time \\(h)} & \shortstack{Age-range \\(mo)} & \# Child. & \# Recs & \shortstack{Time \\(h)} & \shortstack{Age-range \\(mo)} & \shortstack{Sampling \\ strategy} \\
\midrule
bergelson & 44 & 450 & 3600 & 6-17 & 10 & 13 & 5.0 & 7-17 & Random \\
cougar & 27 & 143 & 1144 & 2-67 & 26 & 27 & 6.5 & 25-37 & High vol {\footnotesize(CHI-adult)} \\
kidd & 96 & 554 & 4432 & 9-26 & 0 & 0 & - & - & - \\
lucid & 35 & 224 & 1792 & 11-32 & 10 & 10 & 5.0 & 11-31 & Random \\
warlaumont & 9 & 17 & 136 & 3-18 & 10 & 14 & 5.0 & 3-9 & Random \\
winnipeg & 6 & 13 & 104 & 2-19 & 9 & 10 & 5.0 & 2-32 & Random \\
fausey-trio & 0 & 0 & - & - & 23 & 23 & 1.1 & 6-12 & High vol {\footnotesize (MAL, OCH)} \\
\bottomrule
\end{tabular}

%% file: tables/dof.tex
\resizebox{\textwidth}{!}{
\begin{tabular}{@{}p{6em}llllll@{}}
\toprule
\textbf{Level} & \multicolumn{2}{c}{\textbf{Observations}} & \multicolumn{3}{c}{\textbf{Parameters}} \\
\cmidrule(lr){2-3} \cmidrule(lr){4-6}
 & \textbf{Variable} & \textbf{Dimensions} & \textbf{Variable} & \textbf{Dimensions} & \textbf{Prior} \\
\midrule
\parbox[t]{6em}{\textbf{Recordings}\strut} & \begin{tabular}[t]{@{}l@{}}$n^{\text{recs}}$ \\ age\end{tabular} & \begin{tabular}[t]{@{}l@{}}1401 $\times$ 4 \\ 1401\end{tabular} & \begin{tabular}[t]{@{}l@{}}$v^{\text{recs}}_{k,\text{CHI}}$ \\ $v^{\text{recs}}_{k,s}$ ($s{\neq}\text{CHI}$) \\ $\lambda^{k}_{ij}$\end{tabular} & \begin{tabular}[t]{@{}l@{}}1401 \\ 1401 $\times$ 3 \\ 1401 $\times$ 4 $\times$ 4\end{tabular} & \begin{tabular}[t]{@{}l@{}}$\mathrm{Gamma}(\alpha^{\mathrm{child}}, \alpha^{\mathrm{child}}/\mu_k^{\mathrm{rec}})$ \\ $\mathrm{Gamma}(\alpha^{\mathrm{child}}, \alpha^{\mathrm{child}}/\mu_c)$ \\ $\mathrm{Gamma}(\alpha_{ij}, \alpha_{ij}/\mu_{ij})$\end{tabular} \\
 & \textbf{Total} & \textbf{9807} & & \textbf{28020} & \\
\midrule
\parbox[t]{6em}{\textbf{Children}\strut} & \begin{tabular}[t]{@{}l@{}}$S_c$\end{tabular} & \begin{tabular}[t]{@{}l@{}}217\end{tabular} & \begin{tabular}[t]{@{}l@{}}$\mu_c$ \\ $\alpha_c^{\text{dev}}$\end{tabular} & \begin{tabular}[t]{@{}l@{}}217 $\times$ 3 \\ 217\end{tabular} & \begin{tabular}[t]{@{}l@{}}$\mathrm{Gamma}(\alpha^{\mathrm{pop}}_d, \alpha^{\mathrm{pop}}_{S_c}/(\mu^{\mathrm{pop}} \exp(S_c \beta)))$ \\ $\mathrm{Normal}(\alpha^{\mathrm{dev}}, \sigma^{\mathrm{dev}})$\end{tabular} \\
 & \textbf{Total} & \textbf{217} & & \textbf{868} & \\
\midrule
\parbox[t]{6em}{\textbf{Calibration}\strut} & \begin{tabular}[t]{@{}l@{}}Clip recording \\ $v^{\text{clips}}$ \\ $n^{\text{clips}}$\end{tabular} & \begin{tabular}[t]{@{}l@{}}5226 \\ 5226 $\times$ 4 \\ 5226 $\times$ 4\end{tabular} & \begin{tabular}[t]{@{}l@{}}$(\lambda_{ij})$\end{tabular} & \begin{tabular}[t]{@{}l@{}}97 $\times$ 4 $\times$ 4\end{tabular} & \begin{tabular}[t]{@{}l@{}}$\mathrm{Gamma}(\alpha_{ij}, \alpha_{ij}/\mu_{ij})$\end{tabular} \\
 & \textbf{Total} & \textbf{78390} & & \textbf{1552} & \\
\midrule
\parbox[t]{6em}{\textbf{Population}\strut} &  &  & \begin{tabular}[t]{@{}l@{}}$(\alpha_{ij})$ \\ $(\mu_{ij})$ \\ $\tau$ \\ $\alpha^{\text{child}}$ \\ $\alpha^{\text{pop}}$ \\ $\mu$ \\ $\beta^{\text{sib}}_{\text{OCH}}$ \\ $\beta^{\text{sib}}_{\text{ADU}}$ \\ $p_{\text{sib}}$ \\ $\alpha_{\text{dev}}$ \\ $\sigma_{\text{dev}}$ \\ $\beta^{\text{dev}}$ \\ $\beta^{\text{direct}}$\end{tabular} & \begin{tabular}[t]{@{}l@{}}4 $\times$ 4 \\ 4 $\times$ 4 \\ 1 \\ 4 \\ 2 $\times$ 3 \\ 4 \\ 1 \\ 1 \\ 1 \\ 1 \\ 1 \\ 1 \\ 1\end{tabular} & \begin{tabular}[t]{@{}l@{}}$\mathrm{Pareto}(1, 1.5)$ \\ $\mu_{ii}: \mathrm{Exp}(1)$, $\mu_{i{\neq}j}: \mathrm{Exp}(10)$ \\ $\mathrm{Exponential}(1)$ \\ $\mathrm{Gamma}(4, 1)$ \\ $\mathrm{Gamma}(8, 1)$ \\ $\mathrm{Gamma}(2, 8)$ \\ $\mathrm{Normal}(0, 1)$ \\ $\mathrm{Normal}(0, 1)$ \\ $\mathrm{Uniform}(0, 1)$ \\ $\mathrm{Normal}(0, 1)$ \\ $\mathrm{Exponential}(1)$ \\ $\mathrm{Normal}(0, 1)$ \\ $\mathrm{Normal}(0, 1)$\end{tabular} \\
 & \textbf{Total} & \textbf{0} & & \textbf{54} & \\
\bottomrule
\end{tabular}}

%% file: tables/correlation_vtc_lena.tex
\begin{table}[H]
\centering
\begin{tabular}{l|cccc}
\hline
 & CHI & OCH & FEM & MAL \\
\hline
\multicolumn{5}{l}{\textbf{$R^2$}} \\
Before calibration & \cellcolor[rgb]{0.505,0.699,0.000}\begin{tabular}{@{}c@{}} 0.747 \\[-1ex] {\scriptsize [0.72, 0.77]} \end{tabular} & \cellcolor[rgb]{0.726,0.833,0.000}\begin{tabular}{@{}c@{}} 0.637 \\[-1ex] {\scriptsize [0.61, 0.67]} \end{tabular} & \cellcolor[rgb]{0.756,0.852,0.000}\begin{tabular}{@{}c@{}} 0.622 \\[-1ex] {\scriptsize [0.59, 0.65]} \end{tabular} & \cellcolor[rgb]{0.625,0.772,0.000}\begin{tabular}{@{}c@{}} 0.688 \\[-1ex] {\scriptsize [0.66, 0.71]} \end{tabular} \\
After calibration & \cellcolor[rgb]{0.514,0.705,0.000}\begin{tabular}{@{}c@{}} 0.743 \\[-1ex] {\scriptsize [0.72, 0.77]} \end{tabular} & \cellcolor[rgb]{0.575,0.742,0.000}\begin{tabular}{@{}c@{}} \textbf{0.713}$^{\ast}$ \\[-1ex] {\scriptsize [0.69, 0.74]} \end{tabular} & \cellcolor[rgb]{0.534,0.716,0.000}\begin{tabular}{@{}c@{}} \textbf{0.733}$^{\ast}$ \\[-1ex] {\scriptsize [0.71, 0.76]} \end{tabular} & \cellcolor[rgb]{0.504,0.699,0.000}\begin{tabular}{@{}c@{}} \textbf{0.748}$^{\ast}$ \\[-1ex] {\scriptsize [0.72, 0.77]} \end{tabular} \\
\hline
\end{tabular}
\caption{\label{table:correlation_vtc_lena}Correlation ($R^2$) between LENA and VTC vocalization counts measured for each speaker, with and without calibration. Brackets indicate 95\% confidence intervals. Best values are shown in bold, when before/after differences are significant ($\ast$).}
\label{tab:correlations}
\end{table}

%% file: tables/calibration_vtc_data.tex
\resizebox{\linewidth}{!}{%
\begin{tabular}{l|cccc}
\hline
 & CHI & OCH & FEM & MAL \\
 $N$ & ($N=28$) & ($N=28$) & ($N=28$) & ($N=28$) \\
\hline
\multicolumn{5}{l}{\textbf{$R^2$}} \\
Before (algo) & \cellcolor[rgb]{0.939,0.865,0.000}\begin{tabular}{@{}c@{}} 0.432 \\[-1ex] {\scriptsize [0.14, 0.68]} \end{tabular} & \cellcolor[rgb]{0.955,0.972,0.000}\begin{tabular}{@{}c@{}} 0.523 \\[-1ex] {\scriptsize [0.23, 0.75]} \end{tabular} & \cellcolor[rgb]{0.667,0.797,0.000}\begin{tabular}{@{}c@{}} 0.667 \\[-1ex] {\scriptsize [0.41, 0.83]} \end{tabular} & \cellcolor[rgb]{0.919,0.822,0.000}\begin{tabular}{@{}c@{}} 0.411 \\[-1ex] {\scriptsize [0.12, 0.67]} \end{tabular} \\
After (calibration) & \cellcolor[rgb]{0.981,0.988,0.000}\begin{tabular}{@{}c@{}} 0.510 \\[-1ex] {\scriptsize [0.22, 0.74]} \end{tabular} & \cellcolor[rgb]{0.583,0.746,0.000}\begin{tabular}{@{}c@{}} 0.708 \\[-1ex] {\scriptsize [0.47, 0.85]} \end{tabular} & \cellcolor[rgb]{0.283,0.564,0.000}\begin{tabular}{@{}c@{}} \textbf{0.859}$^{\ast}$ \\[-1ex] {\scriptsize [0.72, 0.93]} \end{tabular} & \cellcolor[rgb]{0.605,0.760,0.000}\begin{tabular}{@{}c@{}} \textbf{0.698}$^{\ast}$ \\[-1ex] {\scriptsize [0.45, 0.85]} \end{tabular} \\
\hline
\multicolumn{5}{l}{\textbf{ICC}} \\
Before (algo) & \cellcolor[rgb]{0.738,0.425,0.000}\begin{tabular}{@{}c@{}} 0.213 \\[-1ex] {\scriptsize [0.00, 0.54]} \end{tabular} & \cellcolor[rgb]{0.826,0.894,0.000}\begin{tabular}{@{}c@{}} 0.587 \\[-1ex] {\scriptsize [0.28, 0.78]} \end{tabular} & \cellcolor[rgb]{0.683,0.304,0.000}\begin{tabular}{@{}c@{}} 0.152 \\[-1ex] {\scriptsize [0.00, 0.49]} \end{tabular} & \cellcolor[rgb]{0.982,0.961,0.000}\begin{tabular}{@{}c@{}} 0.481 \\[-1ex] {\scriptsize [0.14, 0.72]} \end{tabular} \\
After (calibration) & \cellcolor[rgb]{0.867,0.919,0.000}\begin{tabular}{@{}c@{}} \textbf{0.567}$^{\ast}$ \\[-1ex] {\scriptsize [0.26, 0.77]} \end{tabular} & \cellcolor[rgb]{0.535,0.718,0.000}\begin{tabular}{@{}c@{}} 0.732 \\[-1ex] {\scriptsize [0.50, 0.87]} \end{tabular} & \cellcolor[rgb]{0.301,0.575,0.000}\begin{tabular}{@{}c@{}} \textbf{0.849}$^{\ast}$ \\[-1ex] {\scriptsize [0.70, 0.93]} \end{tabular} & \cellcolor[rgb]{0.410,0.641,0.000}\begin{tabular}{@{}c@{}} \textbf{0.795}$^{\ast}$ \\[-1ex] {\scriptsize [0.61, 0.90]} \end{tabular} \\
\hline
\end{tabular}
}

%% file: tables/calibration_lena_data.tex
\resizebox{\linewidth}{!}{%
\begin{tabular}{l|cccc}
\hline
 & CHI & OCH & FEM & MAL \\
 $N$ & ($N=28$) & ($N=28$) & ($N=28$) & ($N=28$) \\
\hline
\multicolumn{5}{l}{\textbf{$R^2$}} \\
Before (algo) & \cellcolor[rgb]{0.962,0.917,0.000}\begin{tabular}{@{}c@{}} 0.458 \\[-1ex] {\scriptsize [0.17, 0.70]} \end{tabular} & \cellcolor[rgb]{0.924,0.834,0.000}\begin{tabular}{@{}c@{}} 0.417 \\[-1ex] {\scriptsize [0.13, 0.67]} \end{tabular} & \cellcolor[rgb]{0.742,0.843,0.000}\begin{tabular}{@{}c@{}} 0.629 \\[-1ex] {\scriptsize [0.36, 0.81]} \end{tabular} & \cellcolor[rgb]{0.901,0.782,0.000}\begin{tabular}{@{}c@{}} 0.391 \\[-1ex] {\scriptsize [0.11, 0.66]} \end{tabular} \\
After (calibration) & \cellcolor[rgb]{0.793,0.874,0.000}\begin{tabular}{@{}c@{}} 0.604 \\[-1ex] {\scriptsize [0.32, 0.79]} \end{tabular} & \cellcolor[rgb]{0.807,0.575,0.000}\begin{tabular}{@{}c@{}} 0.288 \\[-1ex] {\scriptsize [0.04, 0.57]} \end{tabular} & \cellcolor[rgb]{0.665,0.796,0.000}\begin{tabular}{@{}c@{}} 0.667 \\[-1ex] {\scriptsize [0.41, 0.83]} \end{tabular} & \cellcolor[rgb]{0.945,0.879,0.000}\begin{tabular}{@{}c@{}} 0.440 \\[-1ex] {\scriptsize [0.15, 0.69]} \end{tabular} \\
\hline
\multicolumn{5}{l}{\textbf{ICC}} \\
Before (algo) & \cellcolor[rgb]{0.775,0.505,0.000}\begin{tabular}{@{}c@{}} 0.252 \\[-1ex] {\scriptsize [0.00, 0.57]} \end{tabular} & \cellcolor[rgb]{0.865,0.918,0.000}\begin{tabular}{@{}c@{}} 0.568 \\[-1ex] {\scriptsize [0.26, 0.77]} \end{tabular} & \cellcolor[rgb]{0.566,0.047,0.000}\begin{tabular}{@{}c@{}} 0.023 \\[-1ex] {\scriptsize [0.00, 0.39]} \end{tabular} & \cellcolor[rgb]{0.757,0.466,0.000}\begin{tabular}{@{}c@{}} 0.233 \\[-1ex] {\scriptsize [0.00, 0.55]} \end{tabular} \\
After (calibration) & \cellcolor[rgb]{0.442,0.661,0.000}\begin{tabular}{@{}c@{}} \textbf{0.779}$^{\ast}$ \\[-1ex] {\scriptsize [0.58, 0.89]} \end{tabular} & \cellcolor[rgb]{0.974,0.943,0.000}\begin{tabular}{@{}c@{}} 0.472 \\[-1ex] {\scriptsize [0.13, 0.71]} \end{tabular} & \cellcolor[rgb]{0.377,0.621,0.000}\begin{tabular}{@{}c@{}} \textbf{0.811}$^{\ast}$ \\[-1ex] {\scriptsize [0.64, 0.91]} \end{tabular} & \cellcolor[rgb]{0.657,0.791,0.000}\begin{tabular}{@{}c@{}} \textbf{0.672}$^{\ast}$ \\[-1ex] {\scriptsize [0.41, 0.83]} \end{tabular} \\
\hline
\end{tabular}
}

%% file: tables/calibration_vtc_synthetic_table.tex
\resizebox{\linewidth}{!}{%
\begin{tabular}{l|cccc}
\hline
 & CHI & OCH & FEM & MAL \\
 $N$ & ($N=1000$) & ($N=1000$) & ($N=1000$) & ($N=1000$) \\
\hline
\multicolumn{5}{l}{\textbf{$R^2$}} \\
Before (algo) & \cellcolor[rgb]{0.784,0.869,0.000}\begin{tabular}{@{}c@{}} 0.608 \\[-1ex] {\scriptsize [0.57, 0.64]} \end{tabular} & \cellcolor[rgb]{0.702,0.344,0.000}\begin{tabular}{@{}c@{}} 0.172 \\[-1ex] {\scriptsize [0.13, 0.22]} \end{tabular} & \cellcolor[rgb]{0.924,0.834,0.000}\begin{tabular}{@{}c@{}} 0.417 \\[-1ex] {\scriptsize [0.37, 0.46]} \end{tabular} & \cellcolor[rgb]{0.600,0.120,0.000}\begin{tabular}{@{}c@{}} 0.060 \\[-1ex] {\scriptsize [0.03, 0.09]} \end{tabular} \\
After (calibration) & \cellcolor[rgb]{0.554,0.729,0.000}\begin{tabular}{@{}c@{}} \textbf{0.723}$^{\ast}$ \\[-1ex] {\scriptsize [0.69, 0.75]} \end{tabular} & \cellcolor[rgb]{0.870,0.921,0.000}\begin{tabular}{@{}c@{}} \textbf{0.565}$^{\ast}$ \\[-1ex] {\scriptsize [0.52, 0.60]} \end{tabular} & \cellcolor[rgb]{0.844,0.905,0.000}\begin{tabular}{@{}c@{}} \textbf{0.578}$^{\ast}$ \\[-1ex] {\scriptsize [0.54, 0.62]} \end{tabular} & \cellcolor[rgb]{0.691,0.322,0.000}\begin{tabular}{@{}c@{}} \textbf{0.161}$^{\ast}$ \\[-1ex] {\scriptsize [0.12, 0.20]} \end{tabular} \\
\hline
\multicolumn{5}{l}{\textbf{Relative error}} \\
Before (algo) & \cellcolor[rgb]{0.917,0.950,0.000}\begin{tabular}{@{}c@{}} 0.847 \\[-1ex] {\scriptsize [0.81, 0.89]} \end{tabular} & \cellcolor[rgb]{0.975,0.944,0.000}\begin{tabular}{@{}c@{}} 1.119 \\[-1ex] {\scriptsize [1.05, 1.19]} \end{tabular} & \cellcolor[rgb]{0.941,0.871,0.000}\begin{tabular}{@{}c@{}} 1.297 \\[-1ex] {\scriptsize [1.24, 1.35]} \end{tabular} & \cellcolor[rgb]{0.838,0.644,0.000}\begin{tabular}{@{}c@{}} 2.108 \\[-1ex] {\scriptsize [1.93, 2.29]} \end{tabular} \\
After (calibration) & \cellcolor[rgb]{0.701,0.818,0.000}\begin{tabular}{@{}c@{}} \textbf{0.540}$^{\ast}$ \\[-1ex] {\scriptsize [0.51, 0.57]} \end{tabular} & \cellcolor[rgb]{0.813,0.886,0.000}\begin{tabular}{@{}c@{}} \textbf{0.685}$^{\ast}$ \\[-1ex] {\scriptsize [0.65, 0.72]} \end{tabular} & \cellcolor[rgb]{0.790,0.873,0.000}\begin{tabular}{@{}c@{}} \textbf{0.653}$^{\ast}$ \\[-1ex] {\scriptsize [0.62, 0.69]} \end{tabular} & \cellcolor[rgb]{0.989,0.993,0.000}\begin{tabular}{@{}c@{}} \textbf{0.979}$^{\ast}$ \\[-1ex] {\scriptsize [0.94, 1.02]} \end{tabular} \\
\hline
\end{tabular}
}

%% file: tables/calibration_lena_synthetic_table.tex
\resizebox{\linewidth}{!}{%
\begin{tabular}{l|cccc}
\hline
 & CHI & OCH & FEM & MAL \\
 $N$ & ($N=1000$) & ($N=1000$) & ($N=1000$) & ($N=1000$) \\
\hline
\multicolumn{5}{l}{\textbf{$R^2$}} \\
Before (algo) & \cellcolor[rgb]{0.936,0.859,0.000}\begin{tabular}{@{}c@{}} 0.429 \\[-1ex] {\scriptsize [0.38, 0.48]} \end{tabular} & \cellcolor[rgb]{0.704,0.350,0.000}\begin{tabular}{@{}c@{}} 0.175 \\[-1ex] {\scriptsize [0.13, 0.22]} \end{tabular} & \cellcolor[rgb]{0.816,0.597,0.000}\begin{tabular}{@{}c@{}} 0.298 \\[-1ex] {\scriptsize [0.25, 0.35]} \end{tabular} & \cellcolor[rgb]{0.650,0.232,0.000}\begin{tabular}{@{}c@{}} 0.116 \\[-1ex] {\scriptsize [0.08, 0.16]} \end{tabular} \\
After (calibration) & \cellcolor[rgb]{0.798,0.877,0.000}\begin{tabular}{@{}c@{}} \textbf{0.601}$^{\ast}$ \\[-1ex] {\scriptsize [0.56, 0.64]} \end{tabular} & \cellcolor[rgb]{0.874,0.924,0.000}\begin{tabular}{@{}c@{}} \textbf{0.563}$^{\ast}$ \\[-1ex] {\scriptsize [0.52, 0.60]} \end{tabular} & \cellcolor[rgb]{0.952,0.894,0.000}\begin{tabular}{@{}c@{}} \textbf{0.447}$^{\ast}$ \\[-1ex] {\scriptsize [0.40, 0.49]} \end{tabular} & \cellcolor[rgb]{0.713,0.370,0.000}\begin{tabular}{@{}c@{}} \textbf{0.185}$^{\ast}$ \\[-1ex] {\scriptsize [0.14, 0.23]} \end{tabular} \\
\hline
\multicolumn{5}{l}{\textbf{Relative error}} \\
Before (algo) & \cellcolor[rgb]{0.982,0.961,0.000}\begin{tabular}{@{}c@{}} 1.081 \\[-1ex] {\scriptsize [1.03, 1.14]} \end{tabular} & \cellcolor[rgb]{0.974,0.943,0.000}\begin{tabular}{@{}c@{}} 1.120 \\[-1ex] {\scriptsize [1.07, 1.17]} \end{tabular} & \cellcolor[rgb]{0.881,0.738,0.000}\begin{tabular}{@{}c@{}} 1.708 \\[-1ex] {\scriptsize [1.64, 1.78]} \end{tabular} & \cellcolor[rgb]{0.871,0.716,0.000}\begin{tabular}{@{}c@{}} 1.793 \\[-1ex] {\scriptsize [1.72, 1.87]} \end{tabular} \\
After (calibration) & \cellcolor[rgb]{0.780,0.866,0.000}\begin{tabular}{@{}c@{}} \textbf{0.640}$^{\ast}$ \\[-1ex] {\scriptsize [0.60, 0.68]} \end{tabular} & \cellcolor[rgb]{0.803,0.880,0.000}\begin{tabular}{@{}c@{}} \textbf{0.670}$^{\ast}$ \\[-1ex] {\scriptsize [0.64, 0.71]} \end{tabular} & \cellcolor[rgb]{0.864,0.917,0.000}\begin{tabular}{@{}c@{}} \textbf{0.760}$^{\ast}$ \\[-1ex] {\scriptsize [0.73, 0.80]} \end{tabular} & \cellcolor[rgb]{0.981,0.988,0.000}\begin{tabular}{@{}c@{}} \textbf{0.962}$^{\ast}$ \\[-1ex] {\scriptsize [0.92, 1.00]} \end{tabular} \\
\hline
\end{tabular}
}

%% file: tables/regression_table.tex
\begin{table}[H]
\resizebox{\textwidth}{!}{
\begin{tabular}{ll|c|cccc|}
\cline{3-7}
 & & \multirow{3}{*}{\textbf{\begin{tabular}[c]{@{}c@{}}Manual\\ annotations\end{tabular}}} & \multicolumn{4}{c|}{\textbf{Automated annotations}} \\ \cline{4-7} 
 & \multicolumn{1}{c|}{} & & \multicolumn{2}{c|}{\textbf{Prior to calibration}} & \multicolumn{2}{c|}{\textbf{After calibration}} \\ \cline{4-7} 
 & \multicolumn{1}{c|}{} & & \multicolumn{1}{c|}{\textbf{VTC}} & \multicolumn{1}{c|}{\textbf{LENA}} & \multicolumn{1}{c|}{\textbf{VTC}} & \textbf{LENA} \\ \hline
\multicolumn{1}{|l|}{\textbf{Speech quantity}} & \begin{tabular}[c]{@{}l@{}}Female adult\\ proportion\end{tabular} & \begin{tabular}{c} $0.76$ \\ $[0.64, 0.84]$ \end{tabular} & \multicolumn{1}{c|}{\begin{tabular}{c} $0.64$ \\ $[0.63, 0.65]$ \end{tabular}} & \multicolumn{1}{c|}{\begin{tabular}{c} $0.73$ \\ $[0.71, 0.75]$ \end{tabular}} & \multicolumn{1}{c|}{\begin{tabular}{c} $0.74$ \\ $[0.70, 0.77]$ \end{tabular}} & \begin{tabular}{c} $0.78$ \\ $[0.74, 0.82]$ \end{tabular} \\ \hline
\multicolumn{1}{|l|}{\multirow{3}{*}{\textbf{\begin{tabular}[c]{@{}l@{}}Effect\\of independent variables\\ on speech quantities\end{tabular}}}} & \begin{tabular}[c]{@{}l@{}}Age $\to$ output\end{tabular} & \begin{tabular}{c} $0.18$ \\ $[-0.19, 0.53]$ \end{tabular} & \multicolumn{1}{c|}{\begin{tabular}{c} $\mathbf{0.30}$ \\ $\mathbf{[0.26, 0.35]}$ \end{tabular}} & \multicolumn{1}{c|}{\begin{tabular}{c} $\mathbf{0.42}$ \\ $\mathbf{[0.36, 0.48]}$ \end{tabular}} & \multicolumn{1}{c|}{\begin{tabular}{c} $\mathbf{0.43}$ \\ $\mathbf{[0.37, 0.49]}$ \end{tabular}} & \begin{tabular}{c} $\mathbf{0.51}$ \\ $\mathbf{[0.43, 0.58]}$ \end{tabular} \\ \cline{2-7} 
\multicolumn{1}{|l|}{} & \begin{tabular}[c]{@{}l@{}}Siblings $\to$\\input from children\end{tabular} & \begin{tabular}{c} $\mathbf{1.34}$ \\ $\mathbf{[0.28, 2.29]}$ \end{tabular} & \multicolumn{1}{c|}{\begin{tabular}{c} $\mathbf{0.58}$ \\ $\mathbf{[0.48, 0.68]}$ \end{tabular}} & \multicolumn{1}{c|}{\begin{tabular}{c} $\mathbf{0.64}$ \\ $\mathbf{[0.54, 0.75]}$ \end{tabular}} & \multicolumn{1}{c|}{\begin{tabular}{c} $\mathbf{1.14}$ \\ $\mathbf{[0.93, 1.37]}$ \end{tabular}} & \begin{tabular}{c} $\mathbf{1.06}$ \\ $\mathbf{[0.86, 1.24]}$ \end{tabular} \\ \cline{2-7} 
\multicolumn{1}{|l|}{} & \begin{tabular}[c]{@{}l@{}}Siblings $\to$\\adult input\end{tabular} & \begin{tabular}{c} $-0.78$ \\ $[-2.61, 1.04]$ \end{tabular} & \multicolumn{1}{c|}{\begin{tabular}{c} $-0.48$ \\ $[-1.06, 0.11]$ \end{tabular}} & \multicolumn{1}{c|}{\begin{tabular}{c} $\mathbf{-1.39}$ \\ $\mathbf{[-2.12, -0.62]}$ \end{tabular}} & \multicolumn{1}{c|}{\begin{tabular}{c} $\mathbf{-1.84}$ \\ $\mathbf{[-2.73, -0.96]}$ \end{tabular}} & \begin{tabular}{c} $\mathbf{-1.90}$ \\ $\mathbf{[-2.91, -0.85]}$ \end{tabular} \\ \hline
\multicolumn{1}{|l|}{\multirow{2}{*}{\textbf{\begin{tabular}[c]{@{}l@{}}Associations between\\ speech quantities\end{tabular}}}} & \begin{tabular}[c]{@{}l@{}}Input $\to$ output\\(direct)\end{tabular} & \begin{tabular}{c} $-0.18$ \\ $[-1.71, 1.36]$ \end{tabular} & \multicolumn{1}{c|}{\begin{tabular}{c} $\mathbf{0.73}$ \\ $\mathbf{[0.57, 0.89]}$ \end{tabular}} & \multicolumn{1}{c|}{\begin{tabular}{c} $\mathbf{1.16}$ \\ $\mathbf{[0.91, 1.41]}$ \end{tabular}} & \multicolumn{1}{c|}{\begin{tabular}{c} $\mathbf{0.71}$ \\ $\mathbf{[0.41, 1.02]}$ \end{tabular}} & \begin{tabular}{c} $\mathbf{1.19}$ \\ $\mathbf{[0.80, 1.55]}$ \end{tabular} \\ \cline{2-7} 
\multicolumn{1}{|l|}{} & \begin{tabular}[c]{@{}l@{}}Input $\to$ output\\(long-term)\end{tabular} & \begin{tabular}{c} $-0.07$ \\ $[-1.78, 1.62]$ \end{tabular} & \multicolumn{1}{c|}{\begin{tabular}{c} $\mathbf{0.50}$ \\ $\mathbf{[0.26, 0.75]}$ \end{tabular}} & \multicolumn{1}{c|}{\begin{tabular}{c} $0.24$ \\ $[-0.07, 0.54]$ \end{tabular}} & \multicolumn{1}{c|}{\begin{tabular}{c} $\mathbf{0.54}$ \\ $\mathbf{[0.20, 0.93]}$ \end{tabular}} & \begin{tabular}{c} $0.27$ \\ $[-0.13, 0.70]$ \end{tabular} \\ \hline
\end{tabular}
}
\caption{Comparison of regression estimates from manual and automated annotations, including their posterior 95\% credible intervals. Significant effects appear in bold characters. }
\label{table:regression_comparison}
\end{table}

%% file: tables/icc_table.tex
\begin{table}[H]
\centering
\begin{tabular}{lcccc}
\toprule
Speaker & Human--Human & Human--VTC & Human--LENA & LENA--VTC \\
Clips & ($N=61$) & ($N=61$) & ($N=61$) & ($N=61$)\\
\midrule
Child & \cellcolor[rgb]{0.227,0.530,0.000}\begin{tabular}{@{}c@{}} $0.89$ \\[-0.6ex] {\scriptsize [0.82, 0.93]} \end{tabular} & \cellcolor[rgb]{0.903,0.786,0.000}\begin{tabular}{@{}c@{}} $0.39$ \\[-0.6ex] {\scriptsize [0.16, 0.58]} \end{tabular} & \cellcolor[rgb]{0.735,0.839,0.000}\begin{tabular}{@{}c@{}} $0.63$ \\[-0.6ex] {\scriptsize [0.46, 0.76]} \end{tabular} & \cellcolor[rgb]{0.906,0.793,0.000}\begin{tabular}{@{}c@{}} $0.40$ \\[-0.6ex] {\scriptsize [0.16, 0.59]} \end{tabular} \\
Other Child & \cellcolor[rgb]{0.566,0.736,0.000}\begin{tabular}{@{}c@{}} $0.72$ \\[-0.6ex] {\scriptsize [0.57, 0.82]} \end{tabular} & \cellcolor[rgb]{0.994,0.987,0.000}\begin{tabular}{@{}c@{}} $0.49$ \\[-0.6ex] {\scriptsize [0.28, 0.66]} \end{tabular} & \cellcolor[rgb]{0.955,0.901,0.000}\begin{tabular}{@{}c@{}} $0.45$ \\[-0.6ex] {\scriptsize [0.23, 0.63]} \end{tabular} & \cellcolor[rgb]{0.725,0.833,0.000}\begin{tabular}{@{}c@{}} $0.64$ \\[-0.6ex] {\scriptsize [0.46, 0.77]} \end{tabular} \\
Female Adult & \cellcolor[rgb]{0.392,0.630,0.000}\begin{tabular}{@{}c@{}} $0.80$ \\[-0.6ex] {\scriptsize [0.69, 0.88]} \end{tabular} & \cellcolor[rgb]{0.933,0.853,0.000}\begin{tabular}{@{}c@{}} $0.43$ \\[-0.6ex] {\scriptsize [0.20, 0.61]} \end{tabular} & \cellcolor[rgb]{0.684,0.304,0.000}\begin{tabular}{@{}c@{}} $0.15$ \\[-0.6ex] {\scriptsize [0.00, 0.38]} \end{tabular} & \cellcolor[rgb]{0.997,0.998,0.000}\begin{tabular}{@{}c@{}} $0.50$ \\[-0.6ex] {\scriptsize [0.29, 0.67]} \end{tabular} \\
Male Adult & \cellcolor[rgb]{0.359,0.611,0.000}\begin{tabular}{@{}c@{}} $0.82$ \\[-0.6ex] {\scriptsize [0.72, 0.89]} \end{tabular} & \cellcolor[rgb]{0.919,0.951,0.000}\begin{tabular}{@{}c@{}} $0.54$ \\[-0.6ex] {\scriptsize [0.34, 0.70]} \end{tabular} & \cellcolor[rgb]{0.912,0.806,0.000}\begin{tabular}{@{}c@{}} $0.40$ \\[-0.6ex] {\scriptsize [0.17, 0.59]} \end{tabular} & \cellcolor[rgb]{0.919,0.821,0.000}\begin{tabular}{@{}c@{}} $0.41$ \\[-0.6ex] {\scriptsize [0.18, 0.60]} \end{tabular} \\
\bottomrule
\end{tabular}
\caption{Intraclass Correlation Coefficient (ICC) measures of inter-rater agreement, based on vocalization counts in $N=61\times$ 15s clips annotated by two human annotators.}
\label{tab:icc_agreement}
\end{table}

%% file: tables/dof_multivariate.tex
\resizebox{\textwidth}{!}{
\begin{tabular}{@{}p{6em}llllll@{}}
\toprule
\textbf{Level} & \multicolumn{2}{c}{\textbf{Observations}} & \multicolumn{3}{c}{\textbf{Parameters}} \\
\cmidrule(lr){2-3} \cmidrule(lr){4-6}
 & \textbf{Variable} & \textbf{Dimensions} & \textbf{Variable} & \textbf{Dimensions} & \textbf{Prior} \\
\midrule
\parbox[t]{6em}{\textbf{Recordings}\strut} & \begin{tabular}[t]{@{}l@{}}$n^{\text{recs}}$ \\ age\end{tabular} & \begin{tabular}[t]{@{}l@{}}1401 $\times$ 4 \\ 1401\end{tabular} &  &  &  \\
 & \textbf{Total} & \textbf{9807} & & \textbf{0} & \\
\midrule
\parbox[t]{6em}{\textbf{Children}\strut} & \begin{tabular}[t]{@{}l@{}}$S_c$\end{tabular} & \begin{tabular}[t]{@{}l@{}}217\end{tabular} & \begin{tabular}[t]{@{}l@{}}$\mu_c$ \\ $\alpha_c^{\text{dev}}$\end{tabular} & \begin{tabular}[t]{@{}l@{}}217 $\times$ 3 \\ 217\end{tabular} & \begin{tabular}[t]{@{}l@{}}$\mathrm{Gamma}(\alpha^{\mathrm{pop}}_d, \alpha^{\mathrm{pop}}_{S_c}/(\mu^{\mathrm{pop}} \exp(S_c \beta)))$ \\ $\mathrm{Normal}(\alpha^{\mathrm{dev}}, \sigma^{\mathrm{dev}})$\end{tabular} \\
 & \textbf{Total} & \textbf{217} & & \textbf{868} & \\
\midrule
\parbox[t]{6em}{\textbf{Population}\strut} &  &  & \begin{tabular}[t]{@{}l@{}}$\alpha^{\text{child}}$ \\ $\alpha^{\text{pop}}$ \\ $\mu$ \\ $L_\Omega$ \\ $L_\sigma$ \\ $\beta^{\text{sib}}_{\text{OCH}}$ \\ $\beta^{\text{sib}}_{\text{ADU}}$ \\ $p_{\text{sib}}$ \\ $\alpha_{\text{dev}}$ \\ $\sigma_{\text{dev}}$ \\ $\beta^{\text{dev}}$ \\ $\beta^{\text{direct}}$\end{tabular} & \begin{tabular}[t]{@{}l@{}}4 \\ 2 $\times$ 3 \\ 4 \\ 4 $\times$ 4 \\ 4 \\ 1 \\ 1 \\ 1 \\ 1 \\ 1 \\ 1 \\ 1\end{tabular} & \begin{tabular}[t]{@{}l@{}}$\mathrm{Gamma}(4, 1)$ \\ $\mathrm{Gamma}(8, 1)$ \\ $\mathrm{Gamma}(2, 8)$ \\ $\mathrm{LKJCholesky}(1)$ \\ $\mathrm{Gamma}(2, 4)$ \\ $\mathrm{Normal}(0, 1)$ \\ $\mathrm{Normal}(0, 1)$ \\ $\mathrm{Uniform}(0, 1)$ \\ $\mathrm{Normal}(0, 1)$ \\ $\mathrm{Exponential}(1)$ \\ $\mathrm{Normal}(0, 1)$ \\ $\mathrm{Normal}(0, 1)$\end{tabular} \\
 & \textbf{Total} & \textbf{0} & & \textbf{41} & \\
\bottomrule
\end{tabular}}